\pdfoutput=1
\documentclass{article}

\usepackage{xcolor}
\usepackage{float}
\usepackage[preprint]{corl_2026}
\hypersetup{pdftitle={Mulligan: Performance-Guided Data Collection for Efficient On-Robot Learning}, pdfauthor={Lars Ankile, Perry Dong, Rohan Bhowmik, Aneesh Muppidi, David D. Yuan, Shuran Song, Chelsea Finn}, pdfsubject={}}
\usepackage{etoolbox}
\makeatletter
\patchcmd{\@maketitle}{{\LARGE\bf \@title\par}}{{\LARGE\fontfamily{ppl}\bfseries \@title\par}}{}{\PatchFailed}
\makeatother

\usepackage{algorithm}
\usepackage{algpseudocode}
\usepackage{amsmath}
\usepackage{amssymb}
\usepackage{bbm}
\usepackage{array}
\usepackage{graphicx}
\usepackage{wrapfig}
\usepackage{capt-of}
\usepackage{afterpage}
\usepackage{xspace}
\usepackage{etoc}
\usepackage{placeins}
\makeatletter
\renewcommand{\paragraph}{\@startsection{paragraph}{4}{\z@}{0.1pt}{-1em}{\normalsize\bf}}
\renewcommand{\section}{\@startsection{section}{1}{\z@}{-1.5ex \@plus -0.3ex \@minus -0.2ex}{1.0ex \@plus 0.2ex}{\large\bf\raggedright}}
\renewcommand{\subsection}{\@startsection{subsection}{2}{\z@}{-1.2ex \@plus -0.3ex \@minus -0.2ex}{0.5ex \@plus 0.2ex}{\normalsize\bf\raggedright}}
\let\corlmakecaption\@makecaption
\renewcommand{\@makecaption}[2]{\small\corlmakecaption{#1}{#2}}
\makeatother
\usepackage{tikz}
\newcommand{\golfflag}{\tikz[baseline=-0.05ex, x=1ex, y=1ex]{
  \fill[green!55!black] (0,0.3) ellipse (0.95 and 0.3);
  \fill[black!80] (0.05,0.3) ellipse (0.28 and 0.09);
  \draw[line width=0.13ex, black!75, line cap=round] (0,0.3) -- (0,2.1);
  \fill[red!85!black] (0,2.1) -- (1.05,1.65) -- (0,1.2) -- cycle;
}}
\newcommand{\globe}{\tikz[baseline=-0.05ex, x=1ex, y=1ex]{
  \fill[blue!12!white] (0.85,0.9) circle (0.85);
  \draw[line width=0.1ex, blue!45!black] (0.85,0.9) circle (0.85);
  \draw[line width=0.08ex, blue!45!black] (0.85,0.9) ellipse (0.36 and 0.85);
  \draw[line width=0.08ex, blue!45!black] (0,0.9) -- (1.7,0.9);
  \draw[line width=0.08ex, blue!45!black] (0.15,1.35) -- (1.55,1.35);
  \draw[line width=0.08ex, blue!45!black] (0.15,0.45) -- (1.55,0.45);
}}

\usepackage[nameinlink]{cleveref}
\crefname{section}{Sec.}{Secs.}
\Crefname{section}{Section}{Sections}
\crefname{figure}{Fig.}{Figs.}
\Crefname{figure}{Figure}{Figures}
\crefname{table}{Tab.}{Tabs.}
\Crefname{table}{Table}{Tables}
\crefname{algorithm}{Alg.}{Algs.}
\Crefname{algorithm}{Algorithm}{Algorithms}
\crefname{appendix}{App.}{Apps.}
\Crefname{appendix}{Appendix}{Appendices}

\definecolor{collectfig}{HTML}{17BECF}
\definecolor{learnfig}{HTML}{9467BD}
\colorlet{collectcolor}{collectfig!70!black}
\colorlet{learncolor}{learnfig!85!black}
\newcommand{\collectaxis}[1]{\textcolor{collectcolor}{#1}}
\newcommand{\learnaxis}[1]{\textcolor{learncolor}{#1}}

\definecolor{oursink}{HTML}{0F5A5A}
\newcommand{\ours}{\textcolor{oursink}{\textsf{\bfseries Mulligan}}\xspace}

\newcommand{\taskname}[1]{\textsc{#1}\xspace}
\newcommand{\marker}{\taskname{Insert Marker}}
\newcommand{\nut}{\taskname{Thread Nut}}
\newcommand{\cable}{\taskname{Route Cable}}
\newcommand{\sqnarrow}{\taskname{Square-Narrow}}
\newcommand{\sqbroad}{\taskname{Square-Broad}}

\makeatletter
\newcommand{\backlink}[1]{%
  \par\nobreak\vskip-3\p@
  \noindent{\footnotesize\itshape\color{black!55}Referenced from #1.\par}%
  \nobreak\vskip1\p@
  \@afterheading
}
\makeatother

\title{\ours{}: Performance-Guided Data Collection\\for Efficient On-Robot Learning\vspace{-4pt}}

\newlength{\authorlinewidth}
\newcommand{\authorsep}{\hspace{1.3em}}
\author{
  \makebox[\authorlinewidth][c]{\textbf{Lars Ankile\textsuperscript{\golfflag}\authorsep Perry Dong\authorsep Rohan Bhowmik\authorsep Aneesh Muppidi\authorsep David D. Yuan}}\vspace{.35em}\\
  \makebox[\authorlinewidth][c]{\textbf{Shuran Song\authorsep Chelsea Finn}}\vspace{.75em}\\
  Stanford University\vspace{.75em}\\
  \globe~\url{https://mulligan.page/}\enspace\textbullet\enspace\golfflag~\texttt{ankile@stanford.edu}
}

\begin{document}
\etocdepthtag.toc{main}
\settowidth{\authorlinewidth}{\textbf{Lars Ankile\textsuperscript{\golfflag}\authorsep Perry Dong\authorsep Rohan Bhowmik\authorsep Aneesh Muppidi\authorsep David D. Yuan}}
\raggedbottom
\maketitle
\vspace{-18pt}

\begin{center}
  \includegraphics[width=\linewidth]{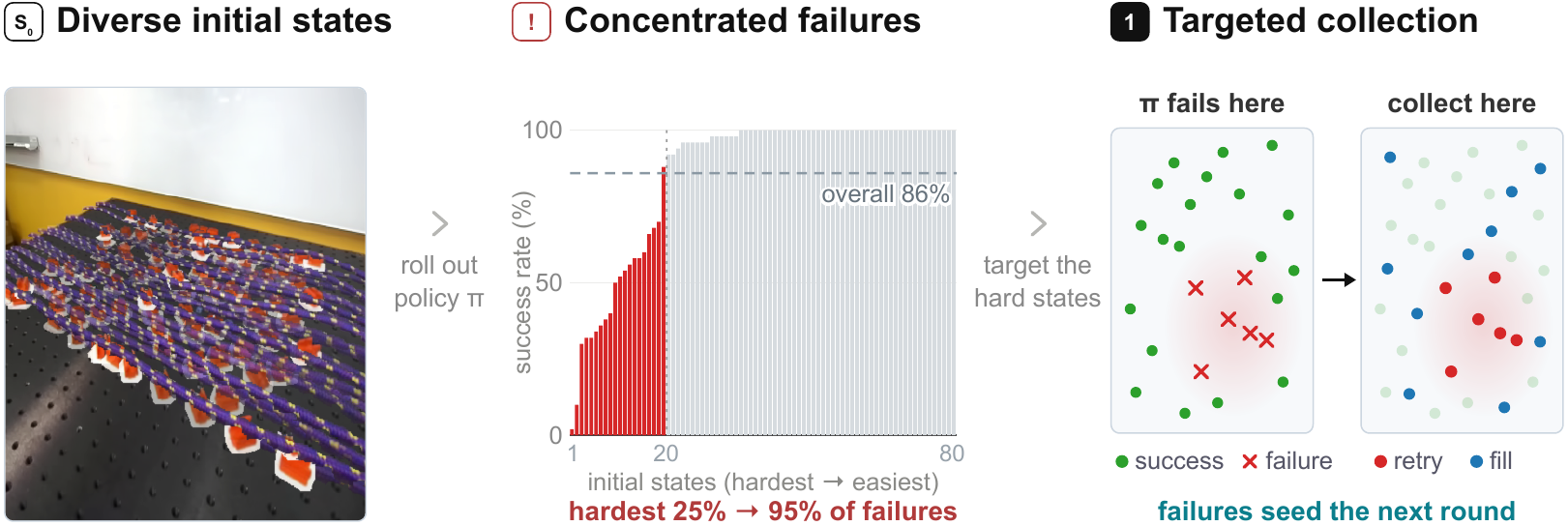}
  \captionof{figure}{%
    \emph{Diverse initial states:} We study learning in deployment with wide initial-state distributions.
    \emph{Concentrated failures:} Failures are not spread evenly across the initial-state space.
    \collectaxis{\emph{Targeted collection:}} Retry observed failures (red) and fill the rest with the least-explored states (blue), to spend operator time where needed.
  }\label{fig:80}
\end{center}
\vspace{-2pt}

\begin{abstract}
  Learning from human demonstrations is a reliable way to teach robots new tasks, but the gains from each additional demonstration shrink as the policy improves. Continued improvement can instead come from supervised deployment, where an operator places the objects and intervenes when the policy fails.
  We ask how to maximize improvement from a fixed budget of supervised episodes on high-precision manipulation tasks with wide ranges of object placements.
  We observe that failures can concentrate in a small subset of initial states, so uniform collection spends much of the operator's time on states the policy already handles.
  \ours{} makes the initial-state distribution a decision, starting each round's episodes at observed failures and untried states. To further improve data efficiency, we augment interactive imitation learning with a value function trained on all data, including failures that imitation discards.
  Across three real-world tasks evaluated on 2,550 held-out, blinded episodes and two simulated tasks, \ours{} outperforms uniform initial-state sampling at matched collection budgets, and combined with value-based action selection, HiL-IDQL+\ours{}, improves final real-task success by 10--34 percentage points. With operator interventions, the human-robot team completes 98\% of collection episodes, remaining productive while the policy learns. Videos, code, and data are available at \url{https://mulligan.page/}.
\end{abstract}

\section{Introduction}
\label{sec:107}

Human demonstrations are a popular starting point for teaching a robot a new manipulation task~\cite{chi2023diffusionpolicy,zhao2023aloha,physicalintelligence2024pi0,kim_openvla_2024,khazatsky2024droid}, but the returns on additional demonstrations often diminish as success rate improves, especially for tasks requiring high precision~\cite{mandlekar2021robomimic,zhao2024alohaunleashed,ankile2024resip,ankile2024juicer,mandlekar2023mimicgen}.
A common next step is to collect corrections with a human in the loop (HiL), where an operator places the objects for each episode, takes control when the policy goes off track, and the corrections join the training set~\cite{ross2011dagger,kelly2019hgdagger,wu2025robocopilot,hu2025rac}.
Recent systems show this reliably achieves high success rates on physical hardware~\cite{hu2025rac,physicalintelligence2025pi06,wang2026divl}, but every episode still costs operator time.
The cost is easier to bear in supervised deployment, where an operator already oversees the robot on real tasks and its rollouts arrive as a byproduct.
We study the batch-online form of this loop~\cite{dong2025batchonline}, which retrains the policy only between rounds, and ask how to get the most improvement out of each supervised episode.

\begin{wrapfigure}{r}{0.40\textwidth}
    \centering
    \vspace{-10pt}
    \includegraphics[width=\linewidth]{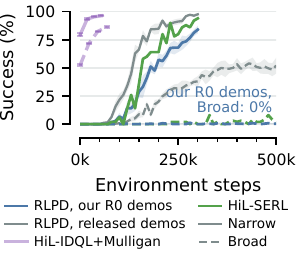}
    \caption{RLPD~\cite{ball2023rlpd,dong2025expo} reaches 84\% on \sqnarrow but 0\% on \sqbroad; HiL-SERL~\cite{luo2024hilserl}, with interventions, ends at 94\% and 2\%.}
    \label{fig:79}
    \vspace{-6pt}
\end{wrapfigure}
Several lines of work address parts of this question.
Interactive imitation learning trains any policy class by behavior cloning (BC) on the corrections but cannot learn from the failed rollouts that deployment yields as a byproduct~\cite{wu2025robocopilot,hu2025rac}.
\phantomsection\label{sec:106}%
Real-world RL learns from every rollout, failures included, through a value function, but gathers those rollouts by exploring online on the hardware~\cite{luo2024hilserl,luo2024serl,chen2025conrft,luo2024rlif,dong2026expoft,lei2025rl100,ankile2025resfit,xiao2025pld} and has shown its best results under narrow initial-state distributions~\cite{luo2024hilserl,luo2024serl,chen2025conrft} (ranges in \cref{app:45}). In simulation, its learner~\cite{ball2023rlpd} fails when the range widens, and human interventions~\cite{luo2024hilserl} do not recover it (\cref{fig:79}).
Batch-online systems~\cite{dong2025batchonline} combine corrections with a value function, training it on deployed rollouts as well, with a policy updated only between rounds and on the wide initial-state distributions of real deployment~\cite{physicalintelligence2025pi06,wang2026divl}.
The lines differ in policy class and learning update.
Recent work scales corrections and RL to vision-language-action (VLA) policies rather than asking how much improvement each supervised episode can provide.

Our key observation is that failures are often unevenly distributed across the initial-state space. E.g., in \cref{fig:80} (middle), 95\% of a policy's failures fall in 25\% of the initial states even though its overall success rate is 86\%.
Raising the success rate means correcting those failures where they happen.
Under uniform collection, the policy makes fewer mistakes as it improves, and the operator becomes increasingly passive.
Yet the operator who decides when to intervene also places the objects, so the initial-state distribution is a decision rather than a fixed part of the task, and can be made deliberately.
The choice is independent of the policy class and learning update, so it combines with any of the methods above.
The question is how to find those states without exhaustive search.

\ours{} finds them with the outcomes each round already produces, deciding which initial states the next round visits and what the operator records there (\cref{fig:80}, right, and \cref{sec:111}). The first round covers the initial-state space evenly and yields the first evidence of where the policy fails. Each later round balances retrying the initial states where the previous one failed with exploring the rest of the space for new failures. At a retried initial state, a correction begins where the policy went off track, so when the mistake is big, it teaches only recovery. Instead, the operator can record a full demonstration from the same initial state, a \emph{counterfactual demonstration} that places the missing behavior in the policy's support. \ours{} only decides what each round collects, so we evaluate it with two learners, imitation alone and imitation with a value function trained on all rollouts, failures included, that selects among the policy's actions~(\cref{sec:109}).

Our main contribution is \ours{}, a data-collection strategy for batch-online imitation and reinforcement learning that selects each round's initial states from the policy's observed performance and yields more improvement from the same supervised budget.
Across three real-world tasks~(\cref{fig:102}), evaluated on 2,550 held-out, blinded episodes, and two simulated tasks, \ours{} outperforms uniform initial-state sampling at matched supervised collection budgets.
Combined with value-based action selection, HiL-IDQL+\ours{} improves final-round success on the real tasks by 10--34 percentage points over HG-DAgger~\cite{kelly2019hgdagger} and completes more tasks per robot minute~(\cref{sec:105}).
With an operator intervening, the human-robot team completes 98\% of collection episodes with no round below 90\%, so learning does not come at the cost of task completion.

\section{Related Work}
\label{sec:112}

\paragraph{Human-in-the-loop robot learning.}
We build on interactive imitation learning, which collects human corrections under the policy's own state distribution and provides recovery supervision where the policy drifts~\citep{ross2011dagger,kelly2019hgdagger}. This line has studied when the human should intervene, by expert judgment~\citep{kelly2019hgdagger} or estimated uncertainty~\citep{hoque2021thriftydagger,hoque2022fleetdagger}, and has scaled collection to long-horizon bimanual manipulation with expressive policies~\citep{zhao2024alohaunleashed,wu2025robocopilot,hu2025rac}. Interventions also carry implicit reward signal~\citep{spencer2020interventions}, and a second family combines human-in-the-loop collection with RL: RLIF~\citep{luo2024rlif} derives rewards from intervention signals, HiL-SERL~\citep{luo2024hilserl} learns from corrections via RLPD~\citep{ball2023rlpd}, and ConRFT~\citep{chen2025conrft} and EXPO-FT~\citep{dong2025expo,dong2026expoft} extend this to finetuning VLAs. We similarly learn from the robot's own rollouts as well as the corrections, but update the policy between rounds rather than online. Closest to our setting are RECAP~\citep{physicalintelligence2025pi06} and LWD~\citep{wang2026divl}, which improve VLAs over rounds of supervised deployment at scale. These recipes focus on the learning update, e.g., advantage-conditioned training on accumulated experience in RECAP. \ours{} instead decides where each round's episodes begin and leaves the learning update unchanged. We evaluate it with imitation alone and with a value function trained on all rollouts, and it could be combined with either recipe.

\phantomsection\label{sec:113}%
\paragraph{Real-world RL and autonomous self-improvement.}
Reinforcement learning has trained policies on physical robots since early policy search over motor skills~\citep{peters2008motorskills,kober2009motorprimitives} and end-to-end visuomotor learning from pixels~\citep{levine2016visuomotor,kalashnikov2018qtopt}, and recent systems make on-robot training sample efficient~\citep{luo2024serl,dong2026expoft,lei2025rl100,seo2024cqn,hu2023ibrl}. These systems report their best results under centimeter-scale initial-state randomization~\citep{luo2024hilserl,luo2024serl,chen2025conrft} (ranges in \cref{app:45}), where a few episodes reach every initial state. We study ranges wide enough that a round's budget covers only part of the space, so the choice of where to start decides which states the operator supervises. A related line improves robots from their own experience in batches, without a human in the loop~\citep{lampe2024stacking,yang2023robofume,ghasemipour2025selfimproving}, and \citet{dong2025batchonline} show that a Q-function with best-of-$N$ extraction improves policies in autonomous batch-online interaction. We adopt their value-learning and extraction design, since best-of-$N$ improves a generative policy without backpropagating through its sampling chain (\cref{app:47}), and add the operator's corrections and counterfactual demonstrations to the value function's training data. \cref{app:43} extends this discussion to real-world RL systems and to RL finetuning of expressive policies.

\paragraph{Choosing where episodes begin.}
Where episodes begin has been a design choice in RL since restart distributions were used to direct policy improvement~\citep{kakade2002cpi}. Curricula move initial states toward intermediate difficulty, expanding outward from the goal~\citep{florensa2017reverse}, generating goals of suitable difficulty~\citep{florensa2018goalgen}, resetting to states along demonstrations~\citep{salimans2018montezuma,resnick2018backplay,dai2021aced,tao2024rfcl,bauza2024demostart} or a guide policy's rollouts~\citep{uchendu2023jsrl}, or returning to stored states and replaying the levels with the highest learning potential alongside fresh ones~\citep{ecoffet2021goexplore,jiang2021plr}. \citet{narvekar2020curriculum} and \citet{portelas2020acl} survey the area. Hard-example mining concentrates training on what a model gets wrong within a fixed dataset~\citep{shrivastava2016ohem}. For real robots, recent work chooses what to collect by which environmental factors to vary~\citep{gao2024datacollection}, which failure conditions a model predicts~\citep{sagar2024robomd}, where the policy is uncertain~\citep{karli2026recall}, which failure states its executions reveal~\citep{zhao2026flare,hao2026far}, which perturbations to apply during a demonstration~\citep{huang2025adc}, or in what order a human demonstrates~\citep{hou2025activecurriculum}. \ours{} moves the initial-state choice to a supervised deployment, where the operator places each initial state, and mixes retried and fresh states as prioritized level replay does. Because the operator can complete the task from any state, collection is not limited to a band of intermediate difficulty and can go to the hardest states found so far, where a counterfactual demonstration records the missing behavior within the same round.

\section{\texorpdfstring{\ours{}}{Mulligan}: Performance-Guided Collection for Supervised Deployment}
\label{sec:110}

\ours{} is a \collectaxis{collection strategy} that decides where each round's episodes start and what the operator records there (\cref{sec:111}). We pair it with two learners. HG-DAgger~\cite{kelly2019hgdagger} behavior-clones the human data, and HiL-IDQL, our human-in-the-loop variant of IDQL~\cite{hansenestruch2023idql}, keeps the same actor and adds a critic trained on all data, failures included. Actor and critic learning are \learnaxis{decoupled}, and at deployment the critic reranks the actor's samples (\cref{sec:109}). \cref{alg:1} gives the loop both learners run.
\begin{algorithm}[t]
    \caption{The supervised improvement loop, with \ours{} choosing the initial states in the Analyze step. HG-DAgger skips the critic and deploys $\pi_\theta$ alone.}
    \label{alg:1}
    \small
    \begin{algorithmic}[1]
        \State $\mathcal{D}_h, \mathcal{D}_\text{all} \gets \mathcal{D}_0$, base demonstrations at scrambled Sobol initial states in $\mathcal{S}_0$
        \For{round $r = 1, \dots, R$}
        \State \textbf{Train:} actor $\pi_\theta$ by BC on $\mathcal{D}_h$; critic $(Q_\psi, V_\xi)$ on $\mathcal{D}_\text{all}$ (\cref{sec:109})
        \State \textbf{Evaluate:} deploy $\pi_\theta$ with $Q_\psi$-reranking (\cref{eq:65}) for $K$ episodes; add data to $\mathcal{D}_\text{all}$
        \State \textbf{Analyze:} choose the next $M$ initial states with \textsc{SelectInitialStates} (\cref{alg:2})
        \State \textbf{Collect:} $M$ supervised episodes, each counterfactual demonstration replacing the last coverage-fill initial state; add successful episodes' human frames to $\mathcal{D}_h$ and all frames to $\mathcal{D}_\text{all}$
        \EndFor
    \end{algorithmic}
\end{algorithm}

\paragraph{Problem setup.}
We model each task as a Markov decision process $(\mathcal{S}, \mathcal{A}, P, r, \gamma, \rho_0)$ with states $s \in \mathcal{S}$, actions $a \in \mathcal{A}$, transition dynamics $P$, a sparse reward $r$ that is $1$ when the task is completed and $0$ otherwise, discount $\gamma$, and an initial-state distribution $\rho_0$, uniform over a bounded range of object poses $\mathcal{S}_0$.
The goal is a policy $\pi_\theta$ that maximizes autonomous success rate on initial states drawn from $\rho_0$.
We study a \emph{human-in-the-loop} (HiL) \emph{batch-online RL} setting that proceeds in rounds under the supervision of an \emph{operator}. The operator places the objects at the start of each episode, watches the policy act, and may take over to correct it. Before the first round, the operator records a base set $\mathcal{D}_0$ of demonstrations. Each round then deploys the policy for $K$ episodes without intervention to measure performance, and for a budget of $M$ supervised episodes, in which the operator intervenes when the policy fails. Every transition from both is stored. Between rounds, we retrain the models on all accumulated data and redeploy.
The policy's parameters stay fixed within a round, since physical deployment favors a stable and performant policy over one that changes continually through online exploration. Fixed parameters also mean every episode in a round comes from the same policy, so the round's outcomes describe that policy. We analyze the accumulated data at the round boundary to plan the next round's collection. Because the operator places the objects, the supervised episodes can start from any initial state in $\mathcal{S}_0$ rather than from states sampled from $\rho_0$, and \ours{} chooses which ones.

\paragraph{Data and notation.}
Supervised deployment produces four kinds of data. The operator provides \emph{demonstrations}, teleoperated completions of the whole task, which include $\mathcal{D}_0$ and the counterfactual demonstrations of \cref{sec:108}, and \emph{corrections}, in which the operator takes over mid-episode. The robot contributes its own \emph{autonomous successes} and \emph{autonomous failures}. Besides the reward, each frame carries only a tag saying whether the human or the policy controlled it. We write $\mathcal{D}_h$ for the human-controlled frames of successful episodes (demonstrations, counterfactuals, and the human segments of corrections) and $\mathcal{D}_\text{all}$ for every collected frame, failures included.

\paragraph{Learning algorithms.}
We train an actor and a critic separately and combine them only at deployment.
The actor $\pi_\theta(\mathbf{a} \mid s)$ is a Diffusion Policy~\cite{chi2023diffusionpolicy}, a generative model over \emph{action chunks} $\mathbf{a} = (a_t, \dots, a_{t+H-1})$, of which the first $h \le H$ actions, the \emph{executed prefix} $\mathbf{a}_{1:h}$, are executed before the policy is queried again. BC fits $\pi_\theta$ to a dataset of state and chunk pairs by maximum likelihood, which a diffusion model optimizes through its denoising loss.
The critic is trained by implicit $Q$-learning (IQL)~\cite{kostrikov2022iql}, which learns $Q_\psi$ and a value function $V_\xi$ from a dataset without sampling actions from the policy. The critic scores only the executed prefix of a chunk, so $\mathbf{a}$ inside $Q_\psi$ denotes $\mathbf{a}_{1:h}$ throughout. $V_\xi$ is fit to $Q_\psi$ by expectile regression, minimizing $\mathbb{E}\big[\,|\tau - \mathbbm{1}[u < 0]|\,u^2\,\big]$ with $u = Q_\psi(s, \mathbf{a}) - V_\xi(s)$. For $\tau > 0.5$, the value tracks the better actions. $Q_\psi$ is then regressed onto the target $r + \gamma V_\xi(s')$ for each transition $(s, \mathbf{a}, r, s')$.
At deployment, implicit diffusion $Q$-learning (IDQL)~\cite{hansenestruch2023idql} uses the critic to select among the actor's samples, improving the actor without gradients. At each state it draws $N$ chunks $\{\mathbf{a}_1, \dots, \mathbf{a}_N\} \sim \pi_\theta(\cdot \mid s)$ and executes
\begin{equation}
    \label{eq:65}
    \mathbf{a}_{i^\star}, \qquad i^\star = \arg\max_{i} Q_\psi(s, \mathbf{a}_i),
\end{equation}
which we refer to as best-of-$N$ (BoN) sampling and which later work has adopted as a stable improvement operator for generative policies~\cite{dong2025batchonline,li2025qchunking,dong2025expo}.

\subsection{\texorpdfstring{\collectaxis{Performance-guided collection}}{Performance-guided collection}}
\label{sec:111}

As \cref{fig:80} (middle) shows, not all of $\mathcal{S}_0$ is equally difficult. The policy is near-perfect over most of $\mathcal{S}_0$, and a small region accounts for almost all of its failures, even though the policy was trained on episodes with evenly-spread initial states (per-region success rates in \cref{app:10}). When failures cluster this way, collecting uniformly across the initial states spends most of the operator's time on states where the policy is already proficient. The states worth visiting are those where the policy can learn the most.
Autonomous improvement with RL is often most effective from states with \emph{intermediate difficulty}, where the policy sometimes succeeds and sometimes fails, and curricula are designed to stay in that regime~\cite{florensa2018goalgen,narvekar2020curriculum,bailey2026sgs}. A human-in-the-loop obviates the need for a curriculum, as the operator can complete the task from any state and directly supply the missing behavior in the hardest states, where it stands to help the most.
The question becomes how to find these hard states, since hardness is observed only where the policy has already been tried. Our collection strategy balances returning to past failures with improving the coverage of $\mathcal{S}_0$, so that we also discover new failures as known ones are fixed.
\cref{alg:2} selects each round's initial states. We describe its two steps, \emph{failure promotion} and \emph{coverage fill}, and then \emph{counterfactual demonstrations}.

\begin{algorithm}[t]
    \caption{\textsc{SelectInitialStates}: performance-guided initial-state selection}
    \label{alg:2}
    \small
    \begin{algorithmic}[1]
        \Require budget $M$; collected initial states $\mathcal{P}$; initial states $\mathcal{F}$ of the round's evaluation failures; promotion cap $m_f$; nearest-neighbor distance $d(s, \mathcal{Y}) = \min_{y \in \mathcal{Y}} \lVert s - y \rVert$
        \State $\mathcal{G} \gets$ up to $m_f$ states from $\mathcal{F}$, earliest failure stage first \Comment{failure promotion}
        \While{$|\mathcal{G}| < M$}
        \State $\mathcal{G} \gets \mathcal{G} \cup \{\arg\max_{s \in \mathcal{S}_0} d(s,\, \mathcal{P} \cup \mathcal{G})\}$ \Comment{coverage fill}
        \EndWhile
        \State \Return $\mathcal{G}$
    \end{algorithmic}
\end{algorithm}

\paragraph{Failure promotion.} An observed failure is the best available evidence of where the policy tends to fail. Each round initializes part of its episodes at the exact initial states where the deployed policy failed in the evaluation just completed. Failures are taken in order of the task stage at which the policy failed, earliest and most severe first (stage labels in \cref{app:57}), up to $m_f$ promoted failures, which reserves the rest of the budget for the coverage fill (guardrails in \cref{app:11}).

\paragraph{Coverage fill.} Coverage begins before any failure has been observed. With a budget of tens of episodes, i.i.d.\ uniform samples leave gaps and clusters by chance, so the round-$0$ base demonstrations are recorded at initial states drawn from a scrambled Sobol sequence~\cite{sobol1967,owen1998scrambling,joe2008sobol}, a \emph{low-discrepancy} sequence whose points cover $\mathcal{S}_0$ more evenly (comparison in \cref{app:54}).
In later rounds, the episodes left after promotion keep supervision reaching unvisited parts of $\mathcal{S}_0$. These depend on everything collected so far, so the coverage fill uses farthest-point selection~\cite{gonzalez1985clustering,eldar1997fps}. Each coverage-fill state is the point of $\mathcal{S}_0$ farthest from every collected initial state, including this round's promoted failures.
The distance $d$ in \cref{alg:2} is normalized Euclidean, computed after scaling each pose coordinate to the unit interval and treating yaw as periodic, with per-task axis weights in \cref{app:34}.

\paragraph{Counterfactual demonstrations.}
\phantomsection\label{sec:108}
A correction begins from a state the policy reached on its own, so it demonstrates recovery from a mistake rather than what the policy should have done. For example, when the policy drops an object in transit, the correction picks it back up, which is useful, but the more important lesson is to carry the object without dropping it. In such cases, the operator completes the correction, resets to the same initial state, and demonstrates the whole task, a \emph{counterfactual demonstration} (CF) that places the missing behavior in the actor's support. When corrections only need to nudge a stuck or slightly misaligned robot to demonstrate the intended behavior, no CF is recorded. We count each CF toward the $M$-episode budget so rounds with and without CFs cost the same number of episodes.

\paragraph{Instantiation.} Round $0$ records $100$ demonstrations ($200$ on \sqbroad), and each later round collects $M = 50$ episodes ($100$ on \cable and in simulation). The $K = 50$ evaluation episodes start from a separate block of initial states, drawn fresh each round and shared by all compared methods, so success is measured on states unseen in training. In the real world, the operator places objects from the reset schematic in \cref{fig:92}. Per-task selection settings, including an optional coverage-fill bias toward harder states, are in \cref{app:34,app:27}.

\subsection{\texorpdfstring{\learnaxis{Policy learning: decoupled actor and critic}}{Policy learning: decoupled actor and critic}}
\label{sec:109}

Collection produces two pools of data: human data from successful episodes only and autonomous rollouts that include outright failures. The actor $\pi_\theta$ is trained by BC on $\mathcal{D}_h$ alone, whereas prior work folds autonomous successes into the actor's mixture~\cite{dong2025batchonline,hu2025rac}, which in our ablation lowers success (\cref{subsec:116}). The critic $(Q_\psi, V_\xi)$ is trained on all of $\mathcal{D}_\text{all}$, so failures that the actor's data excludes still inform action selection. No gradients pass between the two, so the actor can be any policy class, and each component can be evaluated on its own. Beyond the choice of data, the critic departs from IQL in four ways: its transition length, its target, its value head, and the scaling of its action input.
First, a transition spans the $h$ executed steps of a chunk, so the critic bootstraps $h$ steps ahead~\cite{li2025qchunking},
\begin{equation}
    \label{eq:68}
    y = \sum_{i=0}^{h-1}\gamma^{i} r_{t+i} + \gamma^{h} (1-\texttt{done})\,V_\xi(s_{t+h}),
\end{equation}
where $\texttt{done}$ marks the end of an episode.
Second, over long horizons this one-chunk target propagates reward slowly, so the critic instead regresses onto the $\lambda$-return~\cite{sutton2018rl}, an exponentially weighted average of chunked $n$-step targets in the spirit of GAE~\cite{schulman2016gae},
\begin{equation}
    \label{eq:67}
    y^{\lambda} = (1-\lambda)\sum_{n \ge 1} \lambda^{\,n-1} y^{(n)}, \qquad y^{(n)} = \sum_{i=0}^{nh-1}\gamma^{i} r_{t+i} + \gamma^{nh}\,V_\xi(s_{t+nh}),
\end{equation}
truncated at the end of the episode, with $\lambda$ annealed from near $1$ to $0$ over early training so that the target reduces to \cref{eq:68} and its bias vanishes (\cref{app:32}). Third, its value head reimplements distributional implicit value learning (DIVL)~\cite{wang2026divl}, replacing the scalar value with a categorical return distribution (\cref{app:17}). Finally, since these tasks alternate between large transport motions and small insertion corrections, the critic rescales its action input by a state-conditioned affine transform, $\tilde{\mathbf{a}} = (\mathbf{a} - \mu(s)) \odot \exp(-\ell(s))$, with offsets $\mu(s)$ and log-scales $\ell(s)$ produced by a small network, so that action differences are represented at a scale that depends on the current state (\cref{app:26}). At deployment, the critic selects among $N$ chunks drawn in one batched forward pass (\cref{eq:65}).

\section{Experiments}
\label{sec:105}

\noindent\begin{minipage}{\linewidth}
    Our experiments address three questions:
    \begin{list}{}{%
            \leftmargin=1.7em
            \labelwidth=1.5em
            \labelsep=0.2em
            \itemindent=0pt
            \itemsep=0.15em
            \parsep=0pt
            \topsep=0.35em
            \partopsep=0pt
        }
        \item[\textbf{Q1.}] \textit{Comparison.}
              How does \ours{} compare to existing methods?~(\cref{subsec:115})
        \item[\textbf{Q2.}] \textit{Ablations.}
              What does each part of \ours{} contribute?~(\cref{subsec:116})
        \item[\textbf{Q3.}] \textit{Productivity.} Does learning come at the cost of completing tasks during collection?~(\cref{subsec:117})
    \end{list}
\end{minipage}

\setcounter{subsection}{-1}
\subsection{Experiment Setup}
\label{subsec:114}

\begin{figure}[t]
    \centering
    \newcommand{\taskrow}[3]{%
        \begin{tikzpicture}
            \node[inner sep=0pt, anchor=north west] (img) at (0,0) {\includegraphics[width=\linewidth, trim=0 65 0 0, clip]{#1}};
            \node[anchor=north west, inner sep=1.5pt, fill=black, fill opacity=0.6, text opacity=1, text=white, font=\scriptsize, align=left] at (img.north west) {#2};
            \foreach \stage [count=\k from 0] in {#3}
            \node[anchor=south, inner sep=1pt, fill=black, fill opacity=0.5, text opacity=1, text=white, font=\scriptsize] at ([xshift={(\k+0.5)*\linewidth/6}, yshift=1pt] img.south west) {\strut\stage};
        \end{tikzpicture}}
    \taskrow{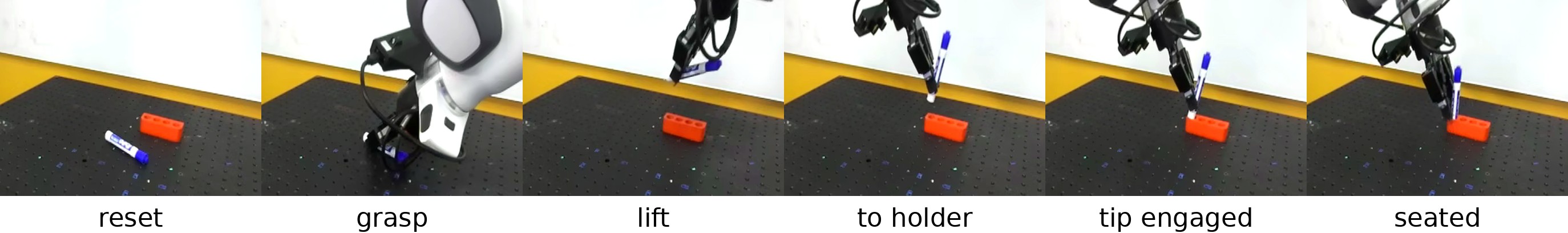}{(a) \marker:\\$198$ steps ($13.2$\,s)}{reset, grasp, lift, to holder, tip engaged, seated}\\
    \taskrow{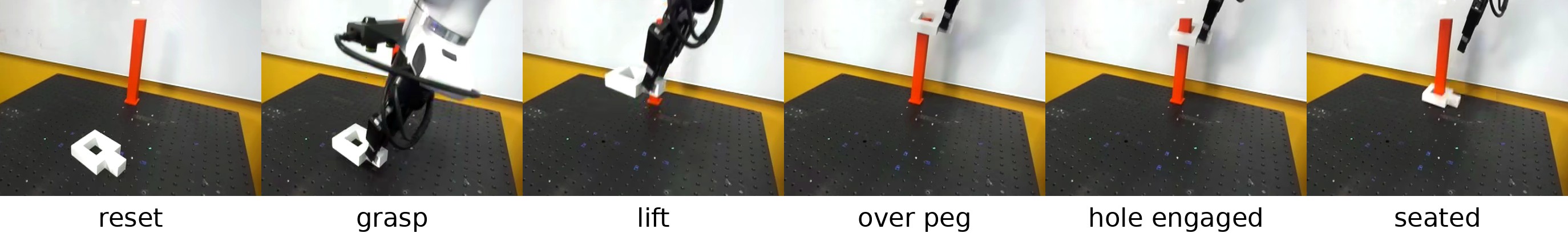}{(b) \nut:\\$144$ steps ($9.6$\,s)}{reset, grasp, lift, over peg, hole engaged, seated}\\
    \taskrow{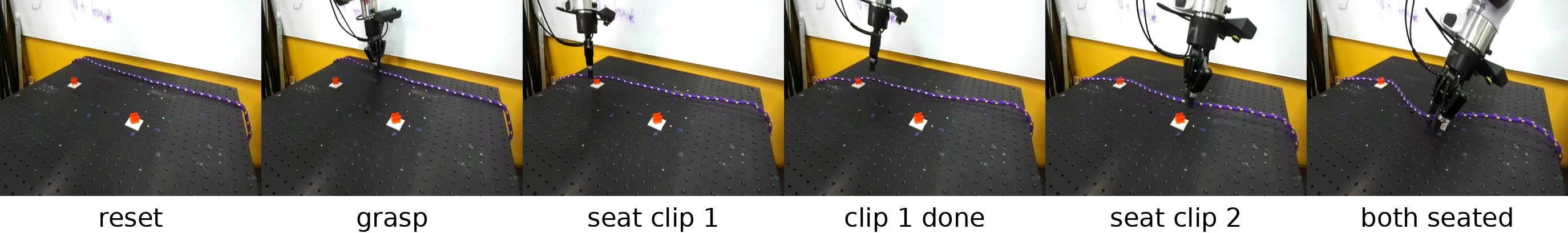}{(c) \cable:\\$335$ steps ($22.3$\,s)}{reset, grasp, seat clip 1, clip 1 done, seat clip 2, both seated}
    \caption{The three real-world tasks. Each row is the median-length successful held-out episode of the final-round HiL-IDQL+\ours{} policy from the side camera, with frames at the rungs of each task's stage ladder (\cref{app:57}), not at even intervals. The median round-0 demonstration takes $12.7$, $10.1$, and $18.6$\,s. In (b), the policy carries the nut above the camera's field of view between frames.}
    \label{fig:102}
\end{figure}
\begin{figure}[t]
    \centering
    \newlength{\resetpanelheight}
    \setlength{\resetpanelheight}{0.2735\linewidth}
    \includegraphics[height=\resetpanelheight]{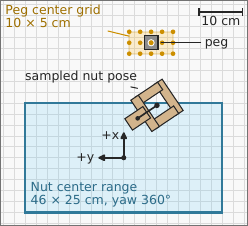}\hfill
    \includegraphics[height=\resetpanelheight]{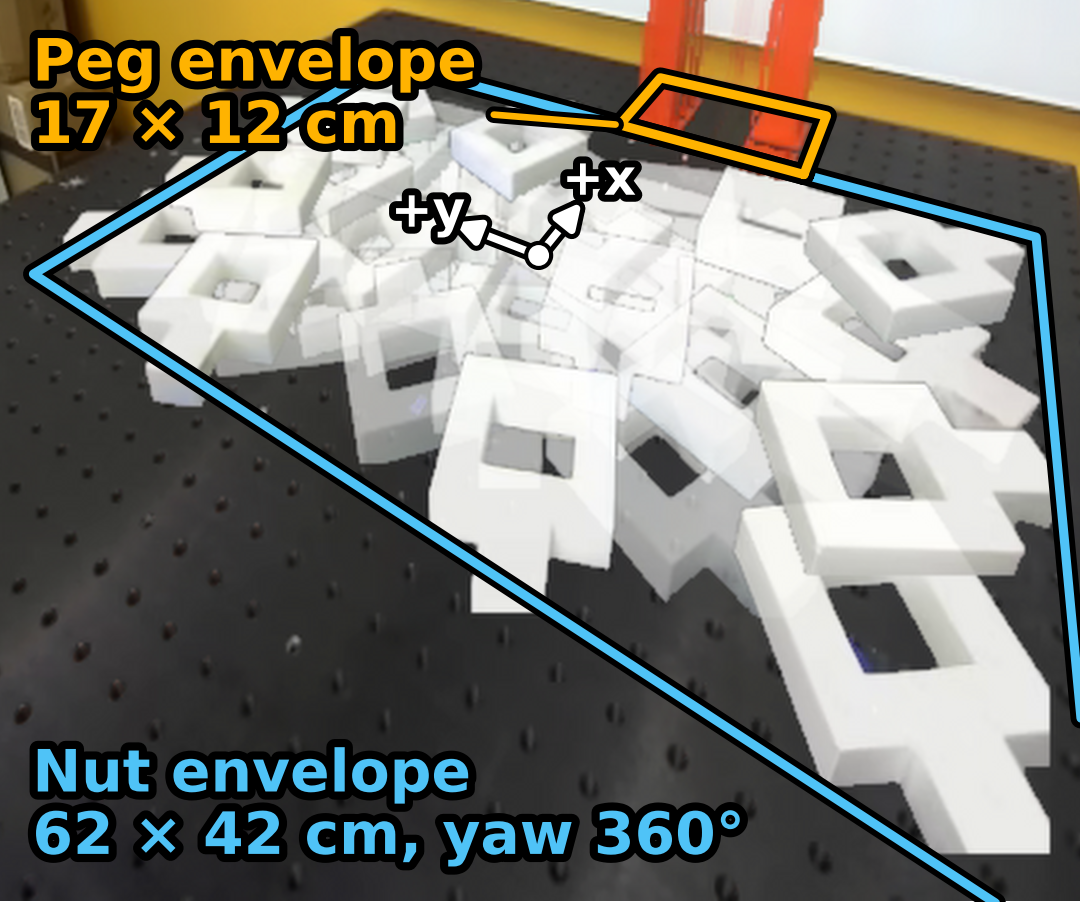}\hfill
    \includegraphics[height=\resetpanelheight]{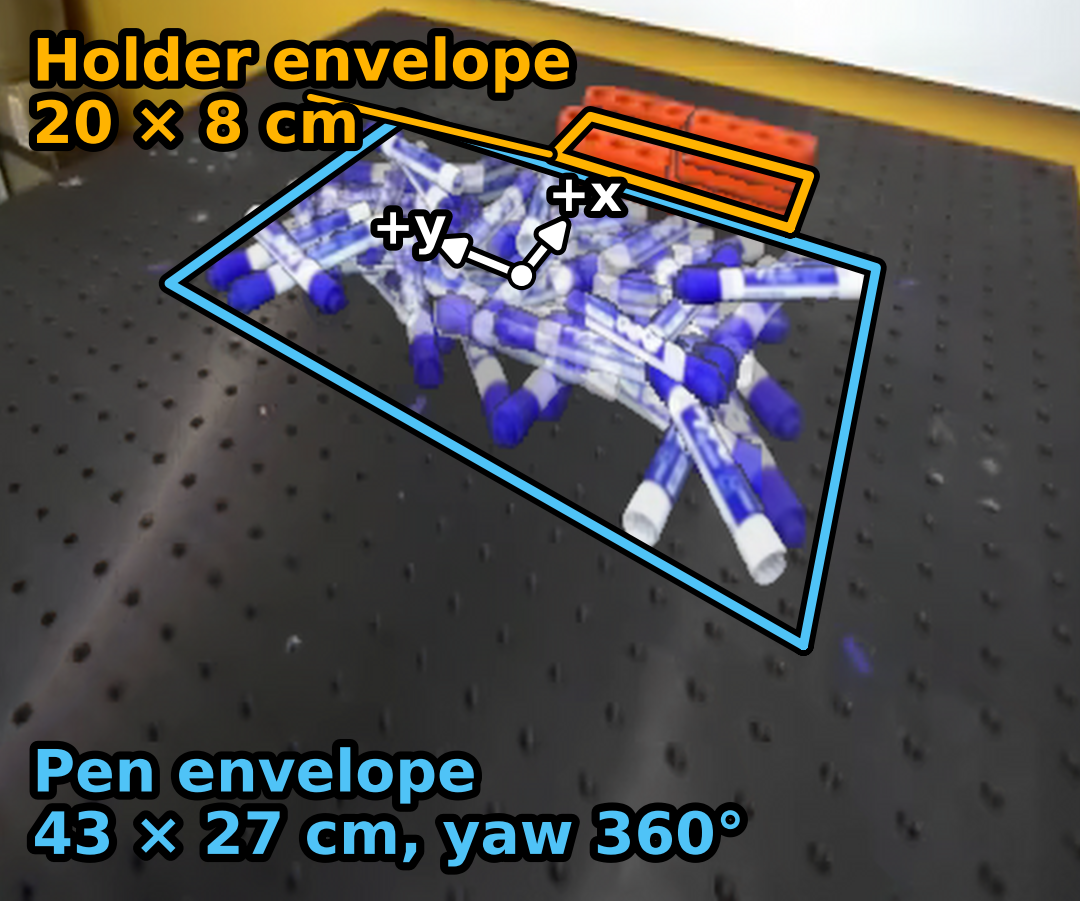}
    \caption{Left: the \nut reset-helper card from which the operator places each episode's objects. The nut center is sampled uniformly over the blue range at any yaw, and the peg is snapped to the orange grid (all three tasks in \cref{app:62}). Middle and right: side-camera overlays of boundary and interior reset poses for \nut and \marker, with each object's envelope, its center range grown by its reach, outlined and labeled. The \cable overlay is in \cref{fig:80}, left.}
    \label{fig:89}
    \label{fig:92}
\end{figure}
\paragraph{Task suite.}
For real-world experiments, we use the DROID~\cite{khazatsky2024droid} Franka Panda platform and evaluate the three tasks in \cref{fig:102}: \marker (grasp a pen and insert it into a holder), \nut (grasp a nut and place it onto a peg), and \cable (seat a rope into two clips in succession). All tasks end in a tight-clearance insertion (e.g., $\sim$1 mm for \marker) or seating stage where small pose errors cause failure, and objects are randomized over wide initial-state ranges (e.g., 15--46 cm). \cref{fig:89} and \cref{fig:80} (left) show the ranges, and \cref{app:45} lists them.

In simulation, we use two Robosuite~\cite{zhu2020robosuite} versions of the same square-nut insertion task. \sqnarrow is the Robomimic~\cite{mandlekar2021robomimic} Square task, which randomizes the nut pose with the peg fixed. \sqbroad is the MimicGen~\cite{mandlekar2023mimicgen} variant that expands the nut-position range and also randomizes the peg position (implementation names and versions in \cref{app:61}).

\begin{figure}[t]
    \centering
    \includegraphics[width=\linewidth]{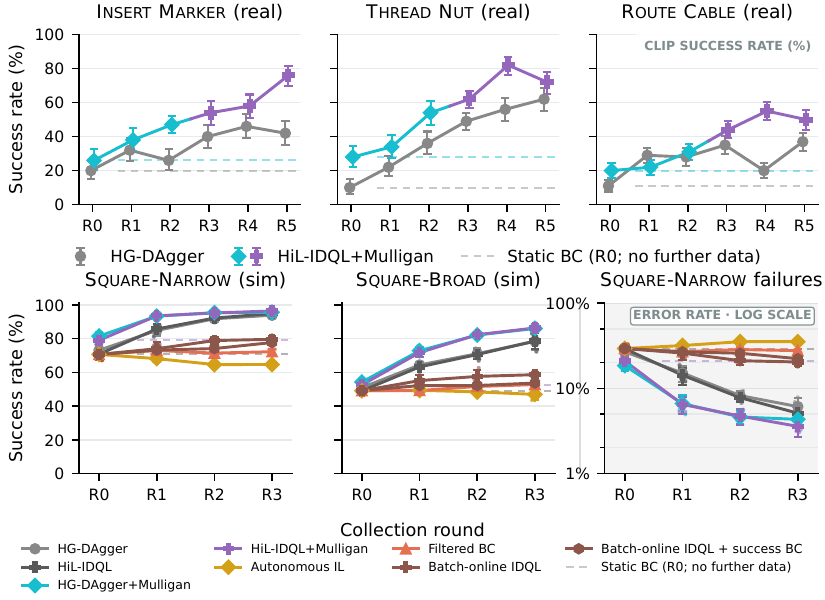}
    \caption{\textbf{Held-out performance over DAgger rounds, real world (top) and simulation (bottom).} Real-world panels show full-task success on \marker and \nut and clip success rate on \cable, the fraction of two clips seated (per-method full-task success in \cref{app:29}). Real-world whiskers are $\pm 1$ SE over $50$ episodes, or $100$ for the pooled \marker R2 HiL-IDQL+\ours{} point and both \nut R3 points (\cref{app:25}). \sqnarrow failures ($100 - \text{SR}$) use a log scale. Dashed lines mark each campaign's static round-0 policy. For real-world HiL-IDQL+\ours{}, teal points use the actor alone and purple points add critic reranking.}
    \label{fig:87}
    \label{fig:97}
    \label{fig:98}
\end{figure}

\paragraph{Compared methods.}
The human-in-the-loop methods vary two design choices. Initial states are either sampled uniformly or chosen by \ours{}, which selects them from the policy's observed performance and adds counterfactual demonstrations (\cref{sec:111}, \cref{app:21}). The learner is either HG-DAgger~\cite{kelly2019hgdagger}, which behavior-clones the human data, or HiL-IDQL, which trains the same actor with an IQL critic on all data and reranks $N$ sampled action chunks at deployment (BoN, \cref{sec:109}; $N$ per task in \cref{app:31}).
We name each method by the learning method and append ``+\ours{}'' when collection is performance-guided, giving HG-DAgger, HG-DAgger+\ours{}, HiL-IDQL, and HiL-IDQL+\ours{}.
Each comparison changes one choice and holds the rest fixed. All four share a Diffusion Policy U-Net actor~\cite{chi2023diffusionpolicy}, the critic formulation, and the per-round collection budget (architectures and hyperparameters in \cref{app:36}). We run all 4 combinations in simulation. In the real world, we run one collection campaign per strategy, so we deploy HG-DAgger against the full method, HiL-IDQL+\ours{}. Reranking helps only once on-policy data has accumulated (\cref{subsec:116}), so we deploy the actor alone in rounds 0--2 and the reranked actor from round 3 on (teal and purple points in \cref{fig:87}). Each round's data are collected by a policy trained on all earlier data. That campaign's actor evaluated without reranking, HG-DAgger+\ours{}, appears only in \cref{app:29}.
In simulation, we also compare against autonomous self-improvement methods, branching from the shared round-0 checkpoints: self-imitation retrains on its own rollouts~\cite{pomerleau1988alvinn,chi2023diffusionpolicy}, filtered BC on only the successful ones~\cite{mirchandani2024scaling}, and batch-online IDQL trains a critic on all rollouts for BoN reranking~\cite{hansenestruch2023idql,dong2025batchonline}. RLPD~\cite{ball2023rlpd,dong2025expo} and HiL-SERL~\cite{luo2024hilserl} start from the same demonstrations.

\paragraph{Evaluation metrics.}
We report autonomous success rate (SR) on held-out initial states and, on \sqnarrow, its complement, failure rate ($100 - \text{SR}$).
On \cable, we use clip success rate in \cref{fig:87}, the mean $0$--$2$ clip score divided by two, and we report the full-task success (both clips seated) in the text and appendix. We also measure execution speed (time-to-completion of successful episodes), throughput (completed task units per robot minute, including time spent by failures), and the rate of human intervention during collection.

\paragraph{Evaluation protocol.}
We keep evaluation separate from collection and run it without interventions. Each round scores every method on one fresh block of $50$ Sobol-sampled initial states that no policy has trained from (\cref{sec:111}), with the methods interleaved on a shuffled list under operator-blind labels, and collection is blinded and interleaved the same way. After scoring, the block's rollouts enter $\mathcal{D}_\text{all}$, so success rates are comparable within a round but, with a new block each round and possible day-to-day drift, not across rounds. In \marker R2 and \nut R3, the reranked policy was scored on additional blocks in later sessions, and each point pools every block on which its policy was scored in that round (\cref{app:25}). In simulation, every policy trains for a preset number of gradient steps and is scored on one fixed set of initial states over five seeds. Real-world whiskers are Wilson one-standard-error intervals~\cite{wilson1927} and simulation intervals are seed-level two-sided Student-$t$ 95\% intervals. Blinding, paired per-state tests, and interval constructions are in \cref{app:25}.

\subsection{Q1: How does \ours{} compare to existing methods?}
\label{subsec:115}

\paragraph{Held-out performance.}

On all three real tasks in \cref{fig:87}, \ours{} ends above uniform sampling, and pooled over all rounds the improvement is significant on each task (paired tests over initial states, $p < 0.005$ on \marker and \nut, and $p = 0.04$ on \cable).
At the final checkpoint, HiL-IDQL+\ours{} finishes $+34$, $+10$, and $+16$ percentage points (pp) above HG-DAgger in full-task success on \marker, \nut, and \cable (counts and tests in \cref{app:55}). The plotted \cable clip success rate is $37\%$ for HG-DAgger and $50\%$ for HiL-IDQL+\ours{}.

In simulation, the HiL methods improve far more than any autonomous method. Without the human in the loop, only the BoN-reranked methods improve, so a value-based objective appears necessary. On \sqnarrow, HiL-IDQL+\ours{} continues to drive failures down through the final round (\cref{fig:98}, log scale). There, HiL-SERL needs over ten times the frames of HiL-IDQL+\ours{} to pass $90\%$, and RLPD does not pass it (\cref{fig:79}, \cref{app:8}).

\begin{figure}[t]
    \centering
    \includegraphics[width=\linewidth]{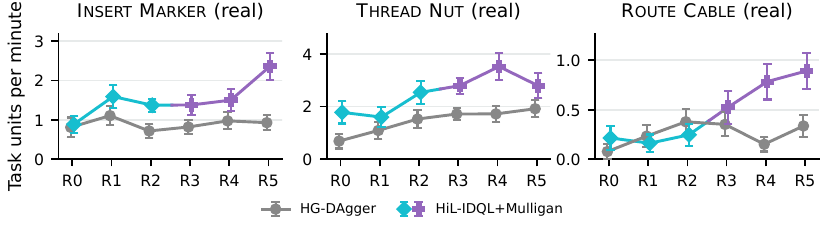}
    \caption{Completed tasks per robot minute, counting time on failed episodes. A completed \cable task requires both clips seated. Colors follow \cref{fig:87}, with $\pm 1$ bootstrap SE whiskers.}
    \label{fig:95}
\end{figure}

\paragraph{Execution speed and throughput.}
These gains do not come at the expense of execution speed. HG-DAgger has been reported to trade speed for success~\cite{hu2025rac}, whereas \cref{fig:95} shows HiL-IDQL+\ours{} completing more tasks per robot minute than HG-DAgger on all three real tasks (time-to-completion and simulation throughput in \cref{app:40,app:52}).

\begin{figure}[t]
    \centering
    \includegraphics[width=\linewidth]{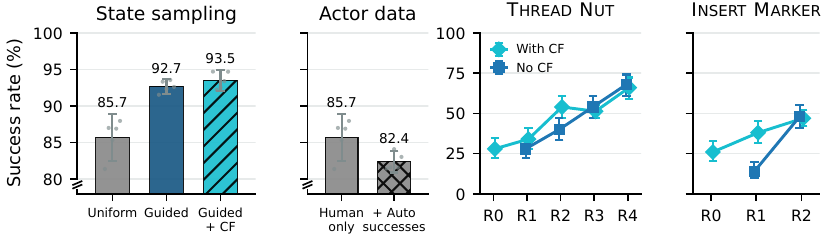}
    \caption{\textbf{Component ablations in simulation (left two) and the real world (right two).} \textbf{State sampling} and \textbf{actor data} on \sqnarrow round~1, with five-seed means, seed dots, and $95\%$ $t$-intervals (\cref{app:51}). \textbf{Counterfactual demos} on \nut and \marker through the last measured no-CF round, with Wilson $\pm 1$ SE whiskers. R0 is the shared source actor.}
    \label{fig:72}
\end{figure}

\subsection{Q2: What does each component contribute?}
\label{subsec:116}

\paragraph{\collectaxis{Data-collection strategies.}}

\textit{Initial-state sampling strategy:} Holding learning fixed, we compare collection strategies. Sobol sampling of the round-0 initial states yields equal or better demo-only point estimates on all five tasks, largest on \nut and \sqnarrow, ranging in simulation from $+8.2$ pp on \sqnarrow to $+3.5$ pp on \sqbroad. \cref{fig:72} (first panel) compares held-out success on \sqnarrow round~1. Mean success rises from uniform to performance-guided collection, with a smaller further increase with CF demos. The full comparison, with Sobol sampling, is in \cref{app:51}, and sampler definitions and per-round counts in \cref{app:50}.

\textit{Counterfactual demonstrations:} \cref{fig:72} (right two panels) compares performance-guided collection with and without CF demos on \nut and \marker. On \marker, round-1 success is $38\%$ with CF versus $14\%$ without; the arms reach $47\%$ and $48\%$ at round~2. The early gap is smaller on \nut: $34\%$ versus $28\%$ at round~1 and $54\%$ versus $40\%$ at round~2, with similar performance by rounds~3--4. These results suggest that CF demos are most useful early, when corrections from difficult initial states may show recovery but not how to complete the task. Larger panels for both tasks and evaluation details are in \cref{app:13}.

\paragraph{\learnaxis{Decoupled actor- and critic-learning.}}

\textit{Value-augmented sampling:} Holding collection fixed, we compare the actor alone with the critic-reranked actor (all methods and rounds in \cref{app:29}). Reranking provides little benefit early and raises success only in later rounds, once enough on-policy data is collected. We also see that performance-guided sampling and critic-reranking contribute at different parts of the task. With \marker and \nut decomposed into grasp and insertion stages (stage-wise success in \cref{app:56}), performance-guided collection outperforms uniform sampling at the grasp stage in early rounds and reaches parity once both have collected enough data, while the critic's gains concentrate at the precise insertion stage.

\textit{Actor-critic data curation:} In \cref{fig:72} (second panel), adding successful autonomous segments to the actor's human data lowers mean success by $3.4$ percentage points, with a $95\%$ Welch interval of $[-6.5,-0.2]$. Full estimates and critic/data-mixture ablations are in \cref{app:51,app:63}.

\begin{figure}[t]
    \begin{minipage}[t]{0.30\linewidth}
        \vspace{0pt}
        \centering
        \includegraphics[width=\linewidth]{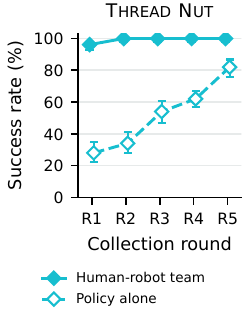}
        \caption{Fresh collection and held-out collector success on \nut (Wilson $z{=}1$). All three tasks and protocol details are in \cref{app:15}.}
        \label{fig:73}
    \end{minipage}\hfill
    \begin{minipage}[t]{0.68\linewidth}
        \vspace{0pt}
        \centering
        \includegraphics[width=\linewidth]{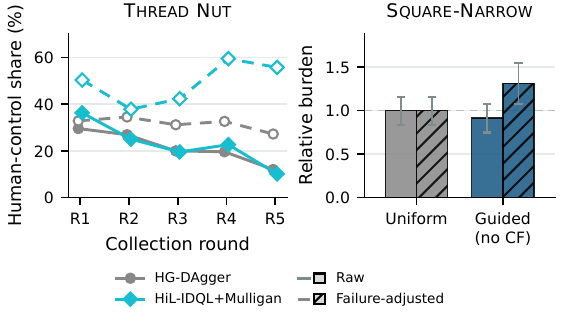}
        \caption{Human burden on \nut (left, human-control frame share) and \sqnarrow (right, intervention-episode share relative to uniform), raw and divided by the collector's failure rate. Both panels exclude CF demos. Wilson $95\%$ bounds treat the normalization as fixed (all tasks and definitions in \cref{app:12}).}
        \label{fig:78}
    \end{minipage}
\end{figure}

\subsection{Q3: Does learning impede productivity?}
\label{subsec:117}

\paragraph{Deployment-time collection is productive.}
An operator supervises every collection episode and can take over when the policy fails. Despite low initial success rates and performance-guided sampling, the human-robot team completes $98\%$ of the 1,009 fresh collection episodes across the three real tasks, with no round below $90\%$, even though a mistake the operator cannot recover from, such as an object knocked off the table, ends the episode as a failure. \cref{fig:73} shows this for \nut, and \cref{app:15} for all three tasks. Each round is thus a batch of tasks completed at near-perfect reliability that also improves the policy. The intervention burden shrinks as learning progresses.

\paragraph{Burden relative to collector performance.}
In \cref{fig:78}, the raw share of human-controlled frames (solid) is similar for both methods and generally declines across rounds, despite their different held-out success rates. Dividing this share by the collector's held-out failure rate $(1-\text{SR})$ gives a higher ratio for \ours{} on \nut (dashed). In simulation, guided sampling likewise shows lower raw but higher failure-adjusted intervention incidence. This is consistent with directing supervision toward difficult states, but the ratio also depends on intervention duration and on the failure-rate divisor, so it is not a direct measure of the sampled states' failure probability.

\section{Discussion}
\label{sec:104}

We studied supervised deployment, where an operator places each episode's objects and intervenes when the policy fails, and asked how to maximize improvement per supervised episode. Choosing each round's initial states from observed failures, and recording counterfactual demonstrations where a correction would only teach recovery, improves the policy faster than uniform collection at equal budget on three real and two simulated tasks. A critic trained on rollouts that arrive anyway, failures included, adds additional later-round gains by reranking actor samples.

We identify three limitations. First, \ours{} assumes initial states can be parameterized and placed at collection time, which supervised collection provides but a fully autonomous deployment may not. In that case, the robot could only select among naturally arriving initial states or drive itself to informative ones. Second, we evaluated \ours{} with one imitation learner and one value-based learner under one formulation of the human-in-the-loop MDP, e.g., bootstrapping value targets across intervention boundaries (\cref{tab:121}). How the same collection strategy performs with other base RL methods and other base policies, including generalist policies updated between rounds, is open. Third, the critic's gains stay outside the actor, since retraining on autonomous rollouts degrades it (\cref{subsec:116}). Distilling the reranked behavior back into the actor could compound gains across rounds. More broadly, our results suggest that how fast a deployed policy improves depends on which experience it collects.

\clearpage

\acknowledgments{We thank Andy Tang for help with running evaluations, Ajay Sridhar for help setting up the physical robot, and Colin Li for helpful discussions early in the project. We also thank Marcel Torne, Anthony Simeonov, and Luke Bailey for feedback on the paper draft. We thank John Yao for help with the tasks, including props and 3D printing, Zhanyi Sun for useful input in early project discussions, and the members of the REAL Lab for input during presentations. Luke Bailey also provided the title font. This work was supported by an NSF CAREER award.

Compute for this work was provided by Yotta Labs Inc.\ through its Research Credit Program. This work used Anvil at Purdue University and Delta at the National Center for Supercomputing Applications through allocation CIS251374 from the Advanced Cyberinfrastructure Coordination Ecosystem: Services \& Support (ACCESS) program, which is supported by U.S. National Science Foundation grants \#2138259, \#2138286, \#2138307, \#2137603, and \#2138296. Delta is supported by the National Science Foundation (award OAC 2005572) and the State of Illinois. Some of the computing for this project was performed on the Sherlock cluster. We would like to thank Stanford University and the Stanford Research Computing Center for providing computational resources and support that contributed to these research results.}

\bibliography{references}

\clearpage

\appendix

\etocdepthtag.toc{appendix}
\makeatletter
\begingroup
  \etocsettagdepth{main}{none}
  \etocsettagdepth{appendix}{subsection}
  \etocsettocstyle{\section*{Appendix contents}}{}
  \let\appendix@tocsection\l@section
  \renewcommand{\l@section}[2]{\appendix@tocsection{#1}{#2}\nopagebreak[4]}
  \small
  \tableofcontents
\endgroup
\makeatother
\clearpage

\section{Author Contributions}
\label{app:4}

LA conceived the project, analyzed how the success rate varies across initial states, designed \ours{} and HiL-IDQL, and built the simulation, teleoperation, real-robot, training, and evaluation infrastructure.
LA collected most of the demonstrations and corrections, ran real-world evaluations, implemented, ran, and analyzed the experiments and baselines, and wrote the paper.
PD provided input on the algorithm, experimental design, and early paper structure and framing.
RB helped collect real-world data, ran real-robot evaluations, helped conceive the state-conditioned action scaling in the critic, contributed to the abstract and introduction, provided writing feedback, and suggested the name \ours{}.
AM took part in technical discussions, helped run evaluations, advised on the HiL-SERL baseline, contributed to the related-work framing, and provided writing feedback.
DY built the stage-labeling and human-review tools behind the stage-wise analysis, helped run real-robot evaluations, and provided writing feedback.
SS advised the project's early development, suggested batch rather than online data collection, proposed training the actor on successful human data and the critic on all data, and provided writing feedback.
CF supervised the project and helped develop the idea of focusing data collection on the initial states where the policy fails.
CF also suggested IDQL for value learning and guided the choice of tasks, baselines, and experiments.
CF helped set the paper's framing, including by drafting an early version of the abstract, and provided extensive writing feedback.

\clearpage\suppressfloats[t]

\section{Extended Results}
\label{app:46}

We expand the main results on held-out performance and throughput, component contributions, and collection success and human burden, in Q1--Q3 order.
Binary real-world curves use Wilson one-standard-error intervals ($z{=}1$).
Paired comparisons report $95\%$ intervals, and other metrics specify their uncertainty in the captions.
Task definitions and evaluation details are in \cref{app:59,app:25}.

\subsection{Full per-arm success rates across rounds}
\label{app:29}
\backlink{\cref{subsec:115}}

\begin{figure}[H]
  \centering
  \includegraphics[width=\linewidth]{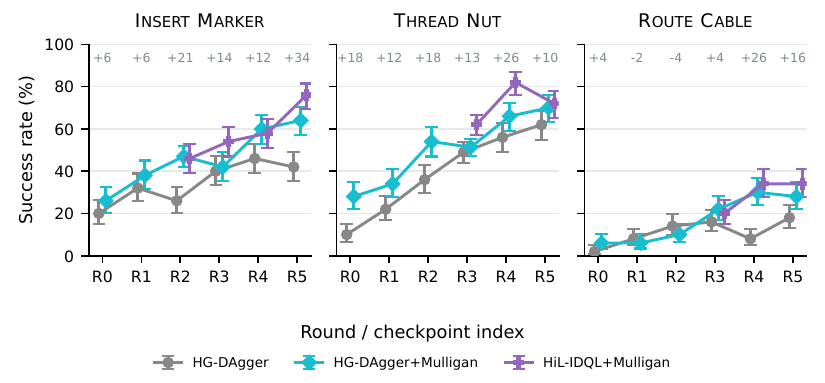}
  \caption{Per-arm full-success results for the policies in \cref{fig:87}, with HG-DAgger (grey), HG-DAgger+\ours{} (teal), and HiL-IDQL+\ours{} with critic reranking (purple) as separate series.
    The top row gives the per-round delta of the policy deployed in the \ours{} campaign over the uniform-sampling HG-DAgger baseline in percentage points.
    Intervals are Wilson-score one-standard-error ($z{=}1$).
    The \cable panel shows full-episode success for rounds R0--R5, scored on one shared block of $50$ initial states.}
  \label{fig:88}
\end{figure}

\cref{fig:88} splits the main headline (\cref{fig:87}) into its three arms.
Here we use full-episode success for \cable, which requires both clips, while the main headline reports the mean clip score divided by two.

\FloatBarrier

\subsection{Headline comparisons and statistical evidence}
\label{app:55}
\backlink{\cref{subsec:115}}

The real-world tests behind \cref{subsec:115} pair arms by initial state.
In each block, we run that round's own actors on the same held-out Sobol states, blind and interleaved in one session (\cref{app:25}).

\paragraph{Final-round comparisons.}
\cref{tab:127} gives the per-task final-round tests behind the headline claim of 10--34 percentage points above HG-DAgger.
The final-checkpoint HG-DAgger+\ours{} to HiL-IDQL+\ours{} increment is positive on all three tasks but individually significant on none.
Across the \cable rounds, the full-success reranking increment changes sign.

\begin{table}[!htbp]
  \centering
  \footnotesize
  \setlength{\tabcolsep}{3pt}
  \caption{Final-round paired comparisons.
    We evaluated \marker R5 in one contemporaneous interleaved session over a single Sobol-$50$ block, and \nut R5 in the final three-arm session on 50 fresh Sobol initial states ($n{=}50$ per arm) with the selected FiLM critic.
    On \cable, R5 is the $600$-episode round, and we score both-clip success on all $50$ initial states of the shared block.
    We report paired-bootstrap intervals and exact McNemar $p$-values.
    ``vs.\ base'' contrasts an arm with that task's uniform HG-DAgger arm, and ``vs.\ actor'' with HG-DAgger+\ours{}.
    Dashes mark contrasts we do not report for that round.}
  \label{tab:127}
  \begin{tabular}{llccc}
\hline
Task & Arm & Success & Contrast & $\Delta$ [CI], $p$ \\
\hline
\marker R5 & HG-DAgger & $21/50$ ($42\%$) & --- & --- \\
 & \shortstack[l]{HG-DAgger+\\Mulligan} & $32/50$ ($64\%$) & vs.\ base & $+22$ pp $[+4, +40]$, $\mathbf{p = 0.0347}$ \\
 & \shortstack[l]{HiL-IDQL+\\Mulligan} & $38/50$ ($76\%$) & vs.\ base & $+34$ pp $[+16, +52]$, $\mathbf{p = 0.0015}$ \\
 & & & vs.\ actor & $+12$ pp $[0, +26]$, $p = 0.146$ \\
\hline
\nut R5 & HG-DAgger & $31/50$ ($62\%$) & --- & --- \\
 & \shortstack[l]{HG-DAgger+\\Mulligan} & $35/50$ ($70\%$) & vs.\ base & $+8$ pp $[-8, +24]$, $p = 0.454$ \\
 & \shortstack[l]{HiL-IDQL+\\Mulligan} & $36/50$ ($72\%$) & vs.\ base & $+10$ pp $[-8, +26]$, $p = 0.359$ \\
 & & & vs.\ actor & $+2$ pp $[-12, +14]$, $p = 1.000$ \\
\hline
\cable R5 & HG-DAgger & $9/50$ ($18\%$) & --- & --- \\
 & \shortstack[l]{HG-DAgger+\\Mulligan} & $14/50$ ($28\%$) & vs.\ base & $+10$ pp $[-4, +24]$, $p = 0.302$ \\
 & \shortstack[l]{HiL-IDQL+\\Mulligan} & $17/50$ ($34\%$) & vs.\ base & $+16$ pp $[+2, +30]$, $p = 0.0768$ \\
 & & & vs.\ actor & $+6$ pp $[-10, +22]$, $p = 0.629$ \\
\hline
\end{tabular}

\end{table}

\paragraph{Round-pooled treatment effect.}
In \cref{tab:118}, we pool the matched blocks of all rounds for the uniform HG-DAgger baseline and the actor-only HG-DAgger+\ours{} arm, sum within-block discordant pairs, and bootstrap the paired delta stratified within block.
Each block compares contemporaneous actors, which makes the result a treatment effect pooled over rounds and not a single-checkpoint comparison.
Performance-guided collection adds $+11.7$ percentage points on \marker, $+10.9$ on \nut, and $+6.0$ on \cable.
Rescues outnumber regressions on every task ($78$ to $43$ on \marker, $80$ to $42$ on \nut, $42$ to $24$ on \cable).
On \nut, we count the final three-arm R5 session once per initial state. Superseded R5 evaluations enter neither the headline nor the pooled comparison.

\begin{table}[!htbp]
  \centering
  \footnotesize
  \setlength{\tabcolsep}{4pt}
  \caption{Round-pooled paired comparison of uniform HG-DAgger against HG-DAgger+\ours{}.
    We compute exact McNemar $p$-values on the summed discordant pairs and bootstrap the intervals stratified within block.
    All six \cable blocks use the same $50$ initial states.}
  \label{tab:118}
  \begin{tabular}{lccccc}
\hline
Task & Paired $n$ & HG-DAgger & \shortstack[l]{HG-DAgger+\\Mulligan} & Paired $\Delta$ [$95\%$ CI] & Exact McNemar $p$ \\
\hline
\marker & $300$ & $103$ ($34.3\%$) & $138$ ($46.0\%$) & $+11.7$ pp $[+5, +19]$ & $\mathbf{0.0019}$ \\
\nut & $350$ & $142$ ($40.6\%$) & $180$ ($51.4\%$) & $+10.9$ pp $[+5, +17]$ & $\mathbf{0.0007}$ \\
\cable & $300$ & $33$ ($11.0\%$) & $51$ ($17.0\%$) & $+6.0$ pp $[+1, +11]$ & $\mathbf{0.0356}$ \\
\hline
\end{tabular}

\end{table}

\paragraph{\cable round comparisons.}
On \cable, we evaluated $15$ policies on the same $50$ initial states, $750$ rollouts in all (\cref{app:25}).
\cref{tab:144} reports every round, R0--R5 with $100$--$600$ training episodes per arm, in both full-episode success and the unnormalized clip score.

\begin{table}[!htbp]
  \centering
  \footnotesize
  \caption{\cable results per round on the shared block of $50$ initial states.
    A success count requires both clips seated, and progress is the mean $0$--$2$ clip score.
    Triplets list HG-DAgger / HG-DAgger+\ours{} / HiL-IDQL+\ours{}, with a dash where we did not evaluate the reranked arm.}
  \label{tab:144}
  \begin{tabular}{lccc}
\hline
Round & Training episodes / arm & Full successes & Mean clip score \\
\hline
R0 & $100$ & $1/3/\text{---}$ & $0.22/0.40/\text{---}$ \\
R1 & $200$ & $4/3/\text{---}$ & $0.58/0.44/\text{---}$ \\
R2 & $300$ & $7/5/\text{---}$ & $0.56/0.62/\text{---}$ \\
R3 & $400$ & $8/11/10$ & $0.70/0.82/0.88$ \\
R4 & $500$ & $4/15/17$ & $0.40/1.12/1.10$ \\
R5 & $600$ & $9/14/17$ & $0.74/0.98/1.00$ \\
\hline
\end{tabular}

\end{table}

HG-DAgger+\ours{} outperforms HG-DAgger in mean clip score at every round except R1.

\paragraph{Simulation.}
\cref{tab:140} gives the per-round contrasts behind the simulation panels of \cref{fig:87}.
We train five seeds per arm and score each on the same fixed held-out grid ($8{,}000$ episodes on \sqnarrow, $30{,}000$ on \sqbroad).
Treating the seed as the unit of analysis, we report Welch-$t$ differences of seed means with $95\%$ intervals.
Performance-guided collection beats uniform collection at every round on both tasks, with or without critic reranking.
On \sqnarrow the margin shrinks as both arms near the ceiling. On \sqbroad it peaks at R2.
The full recipe beats HG-DAgger at every round except \sqbroad R0.
There the two arms differ only in Sobol versus uniform demonstrations, and reranking has no on-policy data yet.

\begin{table}[!htbp]
  \centering
  \scriptsize
  \setlength{\tabcolsep}{3pt}
  \caption{Simulation per-round contrasts in percentage points of held-out success: Welch-$t$ difference of five-seed means with $95\%$ intervals and two-sided $p$-values.
    Actor-only arms use the headline evaluation of each seed's final checkpoint.
    Critic-reranked arms ($N{=}32$) use the DIVL critics of \cref{app:24}, which are the $N{=}32$ numbers plotted in the main paper.}
  \label{tab:140}
  \begin{tabular}{lcccc}
\hline
Contrast & R0 & R1 & R2 & R3 \\
\hline
\multicolumn{5}{l}{\textit{\sqnarrow}} \\
\shortstack[l]{HG-DAgger+\ours{}\\vs.\ HG-DAgger} & \shortstack{$+8.4$ $[+5.5, +11.3]$\\$\mathbf{p < 0.001}$} & \shortstack{$+8.5$ $[+5.3, +11.8]$\\$\mathbf{p < 0.001}$} & \shortstack{$+3.6$ $[+2.5, +4.7]$\\$\mathbf{p < 0.001}$} & \shortstack{$+1.8$ $[+0.3, +3.3]$\\$\mathbf{p = 0.028}$} \\
\shortstack[l]{HiL-IDQL+\ours{}\\vs.\ HiL-IDQL} & \shortstack{$+8.2$ $[+5.2, +11.2]$\\$\mathbf{p < 0.001}$} & \shortstack{$+7.8$ $[+4.6, +11.0]$\\$\mathbf{p = 0.001}$} & \shortstack{$+3.0$ $[+2.0, +4.1]$\\$\mathbf{p < 0.001}$} & \shortstack{$+1.5$ $[+0.5, +2.5]$\\$\mathbf{p = 0.007}$} \\
\shortstack[l]{HiL-IDQL+\ours{}\\vs.\ HG-DAgger} & \shortstack{$+5.9$ $[+3.1, +8.7]$\\$\mathbf{p = 0.001}$} & \shortstack{$+8.7$ $[+5.4, +11.9]$\\$\mathbf{p < 0.001}$} & \shortstack{$+3.5$ $[+2.3, +4.7]$\\$\mathbf{p < 0.001}$} & \shortstack{$+2.5$ $[+1.0, +4.1]$\\$\mathbf{p = 0.006}$} \\
\hline
\multicolumn{5}{l}{\textit{\sqbroad}} \\
\shortstack[l]{HG-DAgger+\ours{}\\vs.\ HG-DAgger} & \shortstack{$+3.3$ $[+1.0, +5.6]$\\$\mathbf{p = 0.011}$} & \shortstack{$+8.5$ $[+6.4, +10.5]$\\$\mathbf{p < 0.001}$} & \shortstack{$+11.1$ $[+7.9, +14.3]$\\$\mathbf{p < 0.001}$} & \shortstack{$+7.6$ $[+2.7, +12.4]$\\$\mathbf{p = 0.009}$} \\
\shortstack[l]{HiL-IDQL+\ours{}\\vs.\ HiL-IDQL} & \shortstack{$+3.5$ $[+1.1, +6.0]$\\$\mathbf{p = 0.011}$} & \shortstack{$+8.5$ $[+6.1, +10.9]$\\$\mathbf{p < 0.001}$} & \shortstack{$+11.8$ $[+9.0, +14.5]$\\$\mathbf{p < 0.001}$} & \shortstack{$+7.8$ $[+3.1, +12.5]$\\$\mathbf{p = 0.007}$} \\
\shortstack[l]{HiL-IDQL+\ours{}\\vs.\ HG-DAgger} & \shortstack{$+1.7$ $[-0.9, +4.2]$\\$p = 0.17$} & \shortstack{$+7.4$ $[+5.0, +9.8]$\\$\mathbf{p < 0.001}$} & \shortstack{$+11.3$ $[+8.1, +14.5]$\\$\mathbf{p < 0.001}$} & \shortstack{$+7.9$ $[+3.0, +12.7]$\\$\mathbf{p = 0.008}$} \\
\hline
\end{tabular}

\end{table}

\FloatBarrier

\subsection{Real-world execution speed and throughput}
\label{app:40}
\backlink{\cref{subsec:115}}

\begin{figure}[!htbp]
  \centering
  \includegraphics[width=\linewidth]{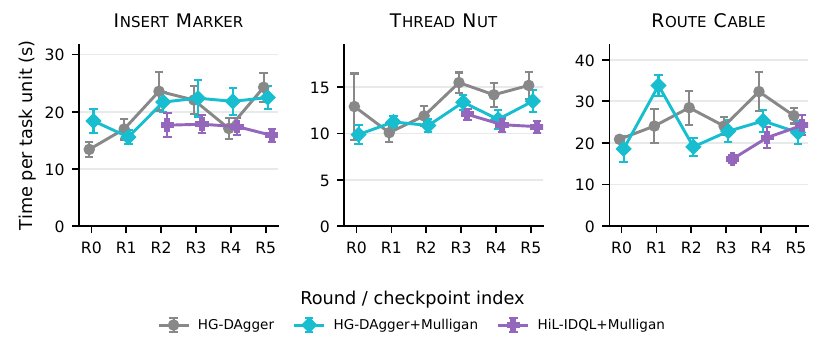}
  \caption{Execution speed on the held-out evaluations: mean robot time per completed task unit (one standard error over completed units), where a unit is a full task completion on every task, requiring both clips on \cable.
    \cable uses all six rounds on its shared block of $50$ initial states.
    Failed episodes are excluded from this duration metric.}
  \label{fig:94}
\end{figure}

\begin{figure}[!htbp]
  \centering
  \includegraphics[width=\linewidth]{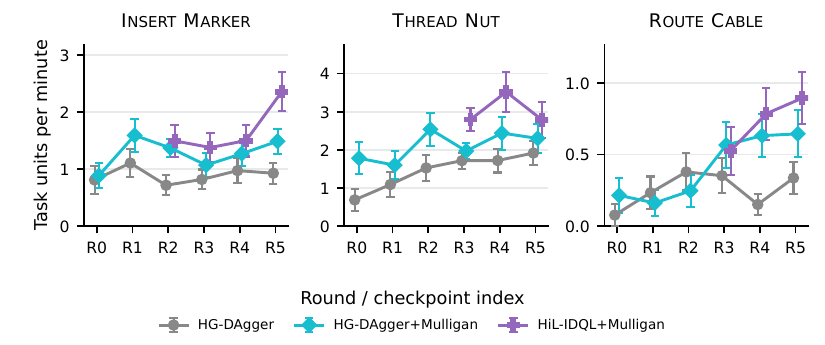}
  \caption{Throughput on the held-out evaluations: completed task units per robot minute over \emph{all} evaluation episode time (one bootstrap standard error over evaluation episodes).
    A task unit is a full episode on all three tasks, and \cable uses all six rounds on its shared block of $50$ initial states.
    Unlike the main-text summary in \cref{fig:95}, this view shows HG-DAgger+\ours{} and HiL-IDQL+\ours{} together wherever both were evaluated.
    Failed and timed-out episodes spend robot time but produce no units.}
  \label{fig:96}
\end{figure}

\cref{fig:94} reports execution speed on the real-world held-out evaluations behind \cref{fig:95}.
\cref{fig:96} shows all three policy arms behind the main-text summary.
For \cable, we use full-episode completion times at $15$ Hz.
On \marker, the reranked arm completes successful episodes faster than the actor-only arm in the final round.
Unlike completion time, throughput also charges time spent on failures.

\FloatBarrier

\subsection{Simulation execution speed and throughput}
\label{app:52}
\backlink{\cref{subsec:115}}

\begin{figure}[!htbp]
  \centering
  \includegraphics[width=\linewidth]{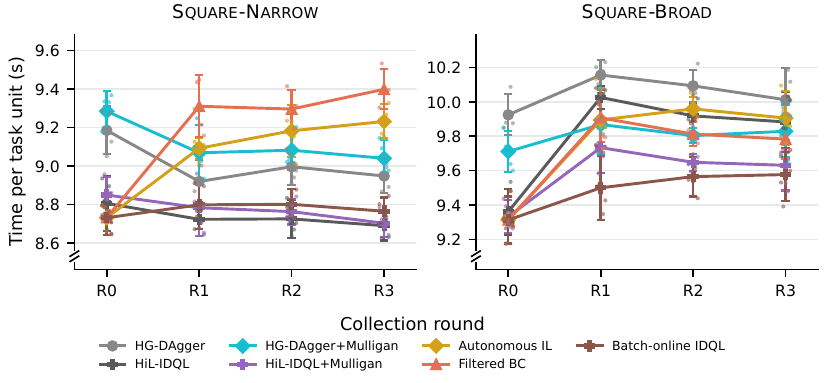}
  \caption{Simulation execution speed on the fixed-grid held-out evaluations for \sqnarrow (left) and \sqbroad (right): mean successful-rollout duration in seconds (env control steps at 20~Hz), averaged over the five training seeds (two-sided Student-t 95\% CIs, dots are per-seed means, y-axes are zoomed at the axis break).
    Series colors and markers follow \cref{fig:97}.}
  \label{fig:99}
\end{figure}

\begin{figure}[!htbp]
  \centering
  \includegraphics[width=\linewidth]{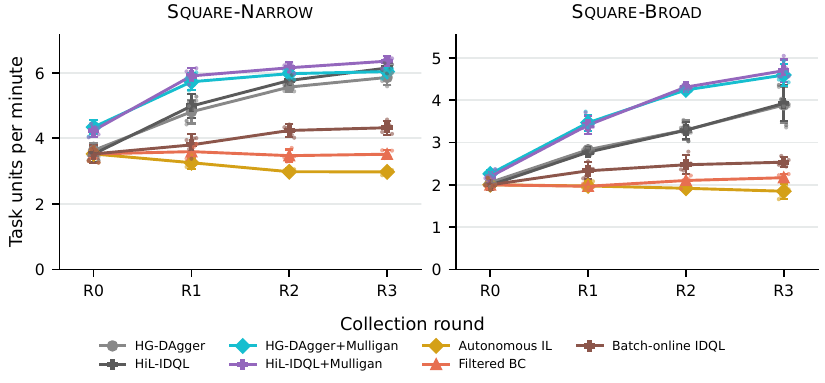}
  \caption{Throughput on the simulation fixed-grid held-out evaluations for \sqnarrow (left) and \sqbroad (right): successful rollouts per robot-minute over \emph{all} evaluation episode time (two-sided Student-t 95\% CIs over five seeds, dots are per-seed values, series and marker conventions as in \cref{fig:99}).
    Failed episodes spend robot time but produce no completions, so this metric combines reliability and speed, as in \cref{fig:95}.}
  \label{fig:100}
\end{figure}

\cref{fig:99,fig:100} extend the real-world execution-speed and throughput analysis (\cref{fig:94} and \cref{fig:95}) to the simulation campaigns behind \cref{fig:97}.
The reranked supervised arms use DIVL critics, and the autonomous arms use scalar-IQL critics (\cref{app:38}).
As \cref{fig:99} shows, successful-rollout durations move by less than a second across rounds while success rates climb by tens of points, so the throughput gains in \cref{fig:100} track reliability.
The two \ours{} collection arms have the highest units per minute at every round on both tasks, except that HiL-IDQL passes HG-DAgger+\ours{} at \sqnarrow R3.
The autonomous-deployment arms improve more slowly, and the best of them (batch-online IDQL) remains below every supervised arm at R3.

\FloatBarrier

\subsection{Sampling-strategy and actor-data ablations}
\label{app:51}
\backlink{\cref{subsec:116}}

\cref{tab:139} gives the numbers behind the two simulation panels of \cref{fig:72}.
All six cells share the round-0 dataset, the round-1 collection budget, and the training recipe, so each row differs from the reference row only in what its label lists.
The human-only rows move the sampler from uniform to Sobol to performance-guided and then add counterfactuals, and the second row ablates the actor's training set.
When we add successful autonomous rollouts to the baseline actor's training mixture, the actor degrades.

\begin{table}[!htbp]
  \centering
  \footnotesize
  \caption{Sampling-strategy and actor-data ablation on \sqnarrow round 1.
    Entries are fixed-grid held-out success rates with $N{=}32$ DIVL reranking, mean $\pm$ standard error over five seeds, and $\Delta$ compares each row with the uniform reference (Welch-t $95\%$ intervals).
    ``HO'' is a human-only actor, and ``$+$auto successes'' adds successful autonomous rollouts.}
  \label{tab:139}
  \begin{tabular}{lcc}
\hline
Collection strategy / actor data & Grid SR (\%) & $\Delta$ [$95\%$ CI] \\
\hline
Uniform, no-CF, HO (reference) & $85.73 \pm 1.16$ & --- \\
Uniform, no-CF, $+$auto successes & $82.37 \pm 0.52$ & $-3.36$ $[-6.53, -0.18]$ \\
Sobol coverage, no-CF, HO & $89.59 \pm 0.45$ & $+3.86$ $[+0.69, +7.03]$ \\
Sobol coverage, with-CF, HO & $91.49 \pm 0.61$ & $+5.76$ $[+2.55, +8.96]$ \\
Performance-guided ($\beta{=}0.5$), no-CF, HO & $92.69 \pm 0.36$ & $+6.96$ $[+3.78, +10.13]$ \\
Performance-guided ($\beta{=}0.5$), with-CF, HO & $93.52 \pm 0.52$ & $+7.80$ $[+4.62, +10.97]$ \\
\hline
\end{tabular}

\end{table}

At round 0, we collect each task's demonstration budget once with uniform and once with Sobol sampling.
Sobol initial states raise DIVL-reranked success on \sqnarrow from $70.80 \pm 0.97$ to $79.03 \pm 0.86$ ($+8.23$ pp $[+5.24, +11.22]$) and on \sqbroad from $49.03 \pm 0.74$ to $52.56 \pm 0.77$ ($+3.53$ pp $[+1.07, +5.98]$).
We evaluate with $N{=}32$ on the grid over five seeds (mean $\pm$ SE, Welch-$t$ 95\% intervals).
On \sqnarrow, the headline HiL-IDQL+\ours{} sequence then runs $79.03 \to 93.52 \to 95.28 \to 96.43$ across R0--R3.

\subsection{Counterfactual-demonstration ablation on \texorpdfstring{\marker}{Insert Marker} and \texorpdfstring{\nut}{Thread Nut}}
\label{app:13}
\backlink{\cref{subsec:116}}

\begin{figure}[!htbp]
  \centering
  \includegraphics[width=\linewidth]{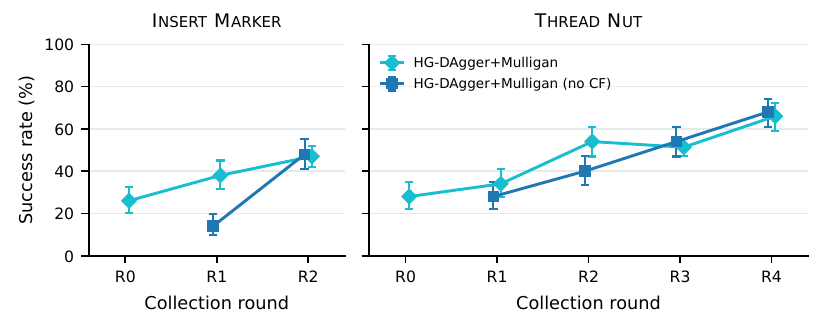}
  \caption{Counterfactual-demonstration ablation on \marker (left) and \nut (right): held-out success rate over DAgger rounds for HG-DAgger+\ours{} with and without CF demos, with Wilson-score one-standard-error intervals ($z{=}1$).
    Each panel ends at the last round with a no-CF evaluation (\marker R2, \nut R4).
    R0 is the shared source actor, shown once.
    Most points use 50 episodes, and the with-CF \marker R2 and \nut R3 points pool 100 and 150 episodes.
    Removing counterfactuals lowers early-round success more on \marker than on \nut, and on both tasks the arms converge as corrections accumulate.}
  \label{fig:84}
\end{figure}

\cref{fig:84} enlarges the \marker and \nut counterfactual-demonstration comparisons shown in \cref{fig:72}.
Most points use 50 held-out episodes.
The with-CF \nut R3 point pools 150 episodes of the same checkpoint, and the with-CF \marker R2 point pools 100 episodes.

\FloatBarrier

\subsection{Per-stage decomposition of the real-world tasks}
\label{app:56}
\backlink{\cref{subsec:116}}

\begin{figure}[!htbp]
  \centering
  \includegraphics[width=\linewidth]{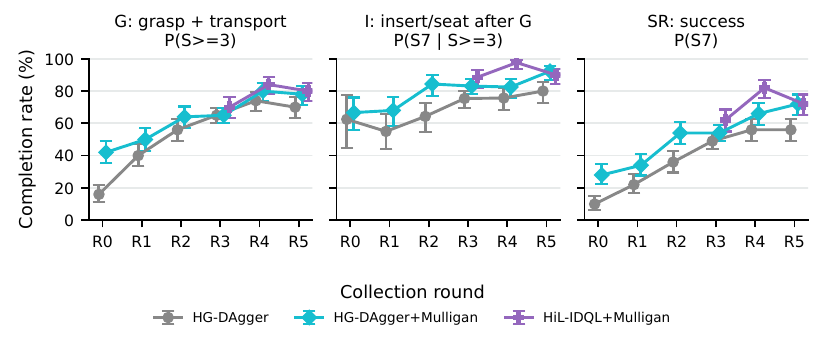}
  \caption{Per-stage decomposition on \nut across rounds: grasp+transport (left), insertion/seat conditioned on grasp (middle), and overall success (right), with Wilson-score one-standard-error intervals ($z{=}1$).
    The grasp-stage gap narrows in later rounds, and the critic improves conditional insertion success at R3--R4.
    At R5, HG-DAgger+\ours{} and HiL-IDQL+\ours{} each have $36/50$ stage successes.
    The R5 HG-DAgger+\ours{} labels come from an earlier three-arm session and the HiL-IDQL+\ours{} labels from a candidate-critic evaluation, as the final R5 evaluation has no stage labels.}
  \label{fig:91}
\end{figure}

\begin{figure}[!htbp]
  \centering
  \includegraphics[width=\linewidth]{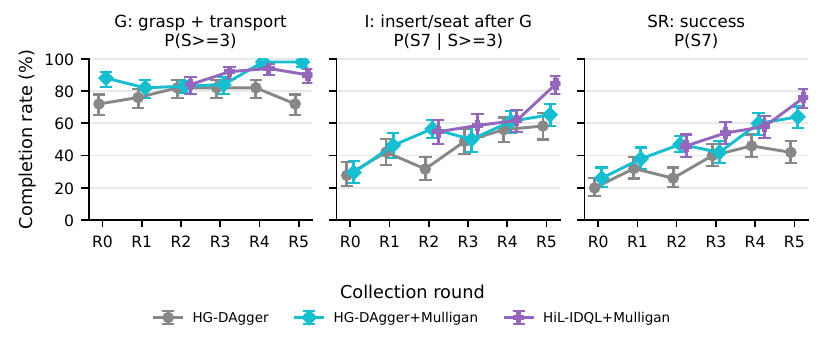}
  \caption{Per-stage decomposition on \marker across rounds: grasp+transport (left), insertion/seat conditioned on grasp (middle), and overall success (right), with Wilson-score one-standard-error intervals ($z{=}1$).
    The collection arm retains a grasp-stage advantage at R5, while reranking improves conditional insertion success.
    R5 uses the stage labels of the headline three-arm session.}
  \label{fig:90}
\end{figure}

We decompose each \marker and \nut episode into grasp, insertion, and overall-success stages (\cref{fig:91,fig:90}).
On \nut, performance-guided collection improves early grasp success and the gap narrows in later rounds.
On \marker, a grasp-stage advantage remains at R5.
The critic improves conditional insertion success at several checkpoints, most clearly on \marker R5 and \nut R3--R4.
The \nut R5 stage-labeled blocks show no reranking gain.
Stage labels come from the labeler described in \cref{app:57}.
We use each arm's available stage-labeled blocks, which can be a subset of the binary-outcome evaluations used in the headline.

\FloatBarrier

\subsection{Collection success: unassisted comparison and all real-world tasks}
\label{app:15}
\backlink{\cref{subsec:117}}

\begin{figure}[!htbp]
  \centering
  \includegraphics[width=\linewidth]{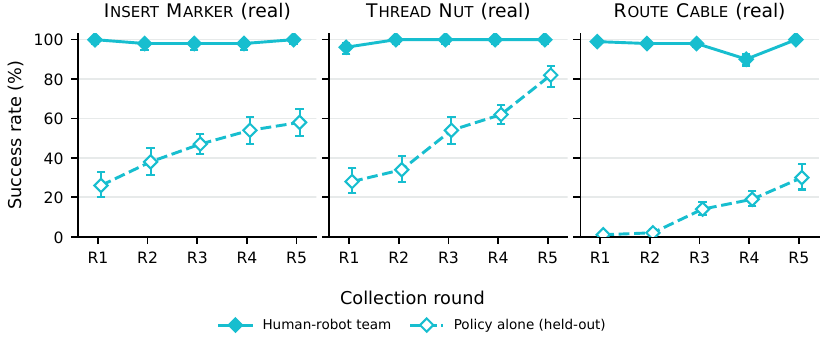}
  \caption{Collection success on \marker, \nut, and \cable.
    The human-robot team completes $98\%$ of fresh collection episodes, with no round below $90\%$ (solid), while the collector policy's preceding-round held-out success is lower (dashed).
    Whiskers are Wilson-score one-standard-error intervals ($z{=}1$).
    CF replay episodes are excluded, and \cable rounds are the $100$-episode increments of \cref{tab:142}.}
  \label{fig:71}
\end{figure}

\cref{fig:71} expands the main-text \nut panel (\cref{fig:73}) to all three real tasks.
The solid series counts successful fresh collection episodes, excluding CF replay.
The remaining failures include episodes in which the operator's own intervention failed.
The dashed series is held-out episode success of the policy that collected that round: collection round $k$ deploys the round $k{-}1$ checkpoint (on \cable R1--R4, the earlier Cartesian-velocity checkpoints), including the reranked collectors when they were used.
Whiskers are Wilson one-standard-error intervals ($z{=}1$).

Collection and held-out evaluation use different initial states, so these curves are an observational comparison.
On \cable the rounds are the paper's $100$-episode increments (\cref{app:20}), and the collector series pools the contemporaneous held-out evaluations of each increment's sessions.
\cref{tab:142} lists the counts behind that panel, where collection success measures full-episode completion rather than the headline clip score.
The R4 dip on \cable comes from $11$ fresh attempts beyond the credited quota that ended in unrecoverable failures.

\begin{figure}[!htbp]
  \centering
  \includegraphics[width=\linewidth]{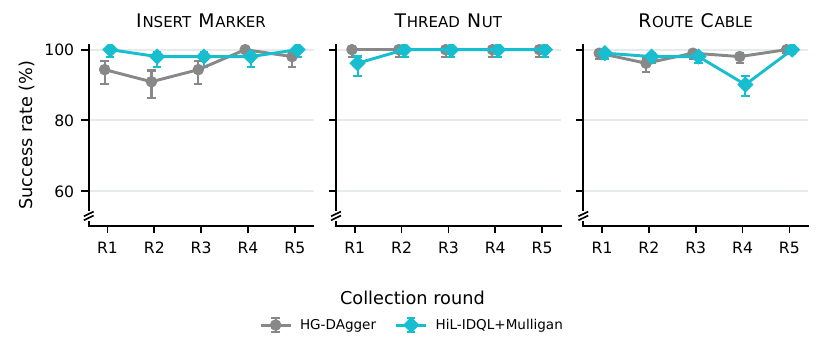}
  \caption{Human-robot team success rate during DAgger collection, per round (Wilson-score one-standard-error intervals with $z{=}1$, y-axis truncated at 50\%), on \marker (left), \nut (middle), and \cable (right, $100$-episode increments), for the uniform-sampling (grey, HG-DAgger) and \ours{} (teal, HiL-IDQL+\ours{}) collection campaigns.
    Over all rounds, the \ours{} campaign completes $98\%$ of collection episodes with no round below $90\%$, and the uniform-sampling campaign completes $98\%$ with no round below $91\%$.}
  \label{fig:85}
\end{figure}

\cref{fig:85} compares the uniform-sampling and \ours{} collection campaigns in the team-only view.

\begin{table}[!htbp]
  \centering
  \footnotesize
  \setlength{\tabcolsep}{4pt}
  \caption{\cable collection on the paper's $100$-episode increments, both HG-DAgger arms.
    ``Fresh / CF'' splits each increment's credited $100$ episodes into episodes that start under policy control and counterfactual replays that the operator demonstrates from the first frame.
    Collection success and human share count only the policy-initiated episodes.
    Success is successful fresh episodes over fresh attempts, and human share is the fraction of their valid frames under human control.
    The comparison is descriptive, since collector policies and sampled initial states change across increments.
    At R5, the HG-DAgger+\ours{} operator stopped fresh collection at the $88$ episodes the increment credits, rather than the $100$ of earlier rounds, because the R5 dataset without counterfactuals would not be used.
    R5 collection success is out of $88$ attempts.}
  \label{tab:142}
  \begin{tabular}{llccc}
    \hline
    Round & Arm               & Fresh / CF  & Collection success & Human share \\
    \hline
    R1    & HG-DAgger         & $100$ / $0$ & $100/101$          & $43.6\%$    \\
          & HG-DAgger+\ours{} & $87$ / $13$ & $100/101$          & $44.2\%$    \\
    R2    & HG-DAgger         & $100$ / $0$ & $100/104$          & $44.0\%$    \\
          & HG-DAgger+\ours{} & $76$ / $24$ & $100/102$          & $41.6\%$    \\
    R3    & HG-DAgger         & $100$ / $0$ & $100/101$          & $38.9\%$    \\
          & HG-DAgger+\ours{} & $89$ / $11$ & $100/102$          & $38.0\%$    \\
    R4    & HG-DAgger         & $100$ / $0$ & $100/102$          & $32.3\%$    \\
          & HG-DAgger+\ours{} & $90$ / $10$ & $100/111$          & $35.7\%$    \\
    R5    & HG-DAgger         & $100$ / $0$ & $100/100$          & $35.1\%$    \\
          & HG-DAgger+\ours{} & $88$ / $12$ & $88/88$            & $39.3\%$    \\
    \hline
  \end{tabular}
\end{table}

\FloatBarrier

\subsection{Collection burden on all real-world tasks}
\label{app:12}
\backlink{\cref{subsec:117}}

\begin{figure}[!htbp]
  \centering
  \includegraphics[width=\linewidth]{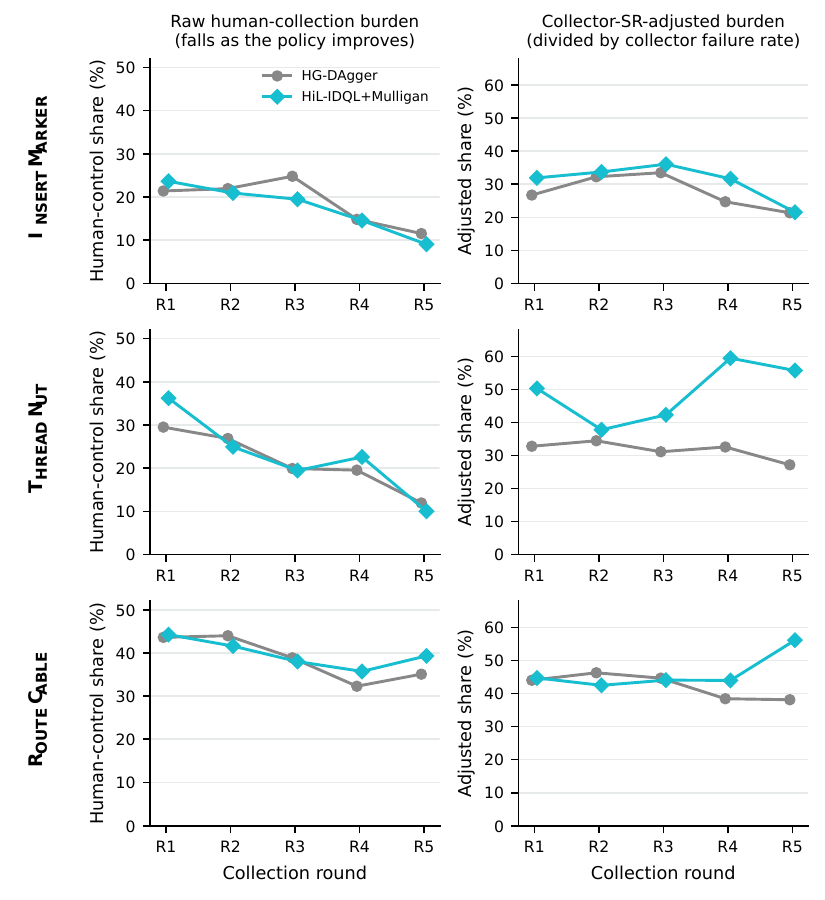}
  \caption{Human collection burden over DAgger rounds on \marker (top), \nut (middle), and \cable (bottom, $100$-episode increments, each normalized by the fresh-episode-weighted held-out success of the policies that collected its source rounds), for the uniform-sampling (grey, HG-DAgger) and \ours{} (teal, HiL-IDQL+\ours{}) collection campaigns.
    Left: raw burden (share of frames under human control in fresh collection episodes).
    Right: the same share normalized by the collection policy's failure rate $(1-\text{SR})$.
    On \cable, where complete success is still near the floor, the normalization is close to the identity.}
  \label{fig:83}
\end{figure}

\cref{fig:83} shows the collection burden (\cref{subsec:117}) for all three real-world tasks on shared per-column scales.
We count burden only on policy-initiated episodes and exclude counterfactual replays.
On \cable, the collector success rate that normalizes the right column is the held-out full-task success of the checkpoint that collected each source round, weighted by fresh episodes over the two source rounds of an increment.
On \nut, where the \ours{} campaign starts from harder states, the normalized burden is above HG-DAgger's from R3 on, as described in the main text.

\subsection{Training-data composition by round}
\label{app:21}
\backlink{\cref{subsec:114}}

\begin{figure}[!htbp]
  \centering
  \includegraphics[width=\linewidth]{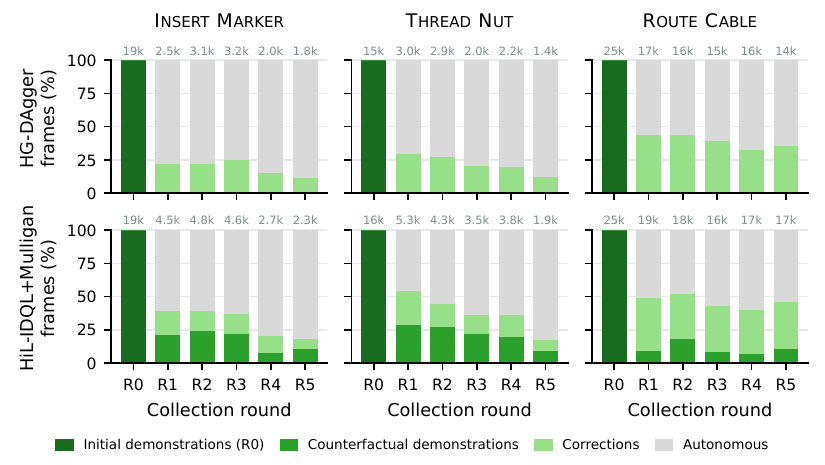}
  \caption{Composition of each round's training data for the uniform-sampling (top) and \ours{} (bottom) collection campaigns, as a share of valid frames.
    The number above each bar is the round's count of human-controlled frames.
    Both campaigns collect the same number of episodes per round.
    The uniform-sampling campaign's data train HG-DAgger, and the \ours{} campaign's data train both HG-DAgger+\ours{} and HiL-IDQL+\ours{}.}
  \label{fig:86}
\end{figure}

The uniform-sampling and \ours{} collection campaigns both start from $100$ initial demonstrations and collect the same number of supervised episodes per round: $50$ on \marker and \nut and $100$ on \cable.
\cref{fig:86} and \cref{tab:122} split each round's training frames by source.
Demonstrations are the human frames of the initial and counterfactual demonstrations, corrections are the human frames of episodes that the policy started, and autonomous frames are all policy-controlled frames, including those inside mixed-control counterfactual demonstrations.
We count the valid frames, after outcome trimming, of the episodes that each campaign trains on.
At the same episode budget, the \ours{} campaign's training data contain $19\%$, $28\%$, and $8\%$ more human frames than the uniform-sampling campaign's on \marker, \nut, and \cable.
The difference comes from the counterfactual demonstrations, which fill part of each round's episode quota with full demonstrations.

\begin{table}[!htbp]
  \centering
  \footnotesize
  \setlength{\tabcolsep}{4pt}
  \caption{Training frames per round and source for the uniform-sampling and \ours{} collection campaigns (\cref{fig:86}).
    A dash marks a source that the round does not contain.
    Demonstrations plus corrections give the human frames, and the \ours{} totals equal $\mathcal{D}_h$ and $\mathcal{D}_\text{all}$ in \cref{tab:123}.}
  \label{tab:122}
  \begin{tabular}{llrrrrrr}
  \hline
  & & \multicolumn{3}{c}{HG-DAgger} & \multicolumn{3}{c}{HiL-IDQL+\ours{}} \\
  Task & Round & Demos & Corrections & Autonomous & Demos & Corrections & Autonomous \\
  \hline
  \marker & R0 & $19{,}382$ & --- & --- & $19{,}096$ & --- & --- \\
   & R1 & --- & $2{,}496$ & $9{,}168$ & $2{,}380$ & $2{,}157$ & $6{,}966$ \\
   & R2 & --- & $3{,}057$ & $10{,}871$ & $2{,}920$ & $1{,}883$ & $7{,}507$ \\
   & R3 & --- & $3{,}208$ & $9{,}726$ & $2{,}759$ & $1{,}873$ & $7{,}898$ \\
   & R4 & --- & $2{,}017$ & $11{,}580$ & $1{,}010$ & $1{,}667$ & $10{,}741$ \\
   & R5 & --- & $1{,}809$ & $13{,}857$ & $1{,}298$ & $974$ & $10{,}397$ \\
  \cline{2-8}
   & Total & $19{,}382$ & $12{,}587$ & $55{,}202$ & $29{,}463$ & $8{,}554$ & $43{,}509$ \\
  \hline
  \nut & R0 & $15{,}429$ & --- & --- & $15{,}728$ & --- & --- \\
   & R1 & --- & $3{,}042$ & $7{,}269$ & $2{,}787$ & $2{,}495$ & $4{,}460$ \\
   & R2 & --- & $2{,}915$ & $7{,}935$ & $2{,}586$ & $1{,}705$ & $5{,}310$ \\
   & R3 & --- & $2{,}000$ & $8{,}042$ & $2{,}080$ & $1{,}407$ & $6{,}124$ \\
   & R4 & --- & $2{,}170$ & $8{,}935$ & $2{,}026$ & $1{,}786$ & $6{,}698$ \\
   & R5 & --- & $1{,}350$ & $9{,}946$ & $998$ & $938$ & $9{,}226$ \\
  \cline{2-8}
   & Total & $15{,}429$ & $11{,}477$ & $42{,}127$ & $26{,}205$ & $8{,}331$ & $31{,}818$ \\
  \hline
  \cable & R0 & $25{,}081$ & --- & --- & $25{,}494$ & --- & --- \\
   & R1 & --- & $16{,}951$ & $21{,}937$ & $3{,}371$ & $15{,}389$ & $19{,}501$ \\
   & R2 & --- & $16{,}453$ & $20{,}966$ & $6{,}045$ & $11{,}736$ & $16{,}530$ \\
   & R3 & --- & $15{,}461$ & $24{,}329$ & $3{,}055$ & $12{,}898$ & $21{,}049$ \\
   & R4 & --- & $16{,}252$ & $34{,}091$ & $2{,}930$ & $14{,}250$ & $25{,}654$ \\
   & R5 & --- & $14{,}231$ & $26{,}341$ & $3{,}880$ & $13{,}244$ & $20{,}449$ \\
  \cline{2-8}
   & Total & $25{,}081$ & $79{,}348$ & $127{,}664$ & $44{,}775$ & $67{,}517$ & $103{,}183$ \\
  \hline
\end{tabular}

\end{table}

\FloatBarrier

\clearpage\suppressfloats[t]

\section{Task and Environment Details}
\label{app:59}

This section describes the five tasks (\cref{app:60}), the initial states each task is reset to (\cref{app:58}), and how episodes are scored (\cref{app:57}).
Platform, control-loop, and episode-cap settings for both domains are in \cref{tab:133}.

\subsection{Platforms and tasks}
\label{app:60}
\label{app:61}

All real-world experiments run on a Franka Panda arm on the DROID platform~\cite{khazatsky2024droid}, with ZED stereo cameras and a SpaceMouse for teleoperation at $15$\,fps.
An operator watches every collection episode from the teleoperation station and may take over at any point to correct the policy, rescue the episode, or record a counterfactual demonstration.
All three real-world tasks end in an insertion or seating stage with tight clearance, where a small pose error is enough to fail.

In simulation we use Robosuite~\cite{zhu2020robosuite} with Robomimic~\cite{mandlekar2021robomimic} and MimicGen~\cite{mandlekar2023mimicgen} assets (MimicGen 1.0.0 on Robosuite 1.5.2), on a 7-DoF Franka arm at a $20$\,Hz environment control rate.
Both simulated tasks are square-nut insertion and differ only in the initial-state distribution.

\cref{tab:147} lists the tasks, and \cref{fig:102} shows one successful final-round episode of each real-world task, frame by stage.

\begin{table}[!htbp]
  \centering
  \footnotesize
  \setlength{\tabcolsep}{4pt}
  \caption{The five tasks.
    The last column is the median length of the round-0 teleoperated demonstrations over all arms ($200$ on \sqnarrow and $400$ on \sqbroad), and in the real world also the held-out round-0 block ($250$ per task).}
  \label{tab:147}
  \begin{tabular}{lp{8.8cm}r}
    \hline
    Task      & Objective                                                                                                                                                           & Median demo \\
    \hline
    \multicolumn{3}{l}{\emph{Real world}}                                                                                                                                                         \\
    \marker   & Grasp a pen from the table and insert it into a holder whose mouth leaves $\sim 1$\,mm of clearance.                                                                & $12.7$\,s   \\
    \nut      & Grasp a square nut and thread it onto a fixed-diameter peg.                                                                                                         & $10.1$\,s   \\
    \cable    & Seat a rope into two clips in succession.                                                                                                                           & $18.6$\,s   \\
    \multicolumn{3}{l}{\emph{Simulation}}                                                                                                                                                         \\
    \sqnarrow & Insert a square nut onto a fixed peg. This is the Robomimic Square task (MimicGen \texttt{Square\_D0}).                                                             & $8.3$\,s    \\
    \sqbroad  & The same insertion over a wider nut range with a randomized peg (MimicGen \texttt{Square\_D1}). Nut and peg placements that collide or are infeasible are rejected. & $8.6$\,s    \\
    \hline
  \end{tabular}
\end{table}

\subsection{Initial-state ranges and difficulty}
\label{app:58}

\cref{tab:145} gives the initial-state ranges.
Position ranges refer to object or target centers, and rotations are in-plane yaw.
In the real world, the operator places the objects from a per-episode reset-helper schematic (\cref{app:62}).
We compare these ranges with those of prior systems in \cref{app:45}.

\begin{table}[!htbp]
  \centering
  \footnotesize
  \setlength{\tabcolsep}{4pt}
  \caption{Initial-state ranges.
    Position ranges are the $x{\times}y$ extents, in centimeters, over which each object or target center is placed.
    Yaw is the in-plane rotation range of the movable object.
    Each \cable clip takes one of three discrete yaws, $0^\circ$ or $\pm45^\circ$ from its nominal orientation.}
  \label{tab:145}
  \begin{tabular}{lcp{7cm}l}
    \hline
    Task      & DoF & Position range ($x{\times}y$, cm)                              & Yaw                 \\
    \hline
    \marker   & $5$ & Pen $15.2{\times}30.5$, holder $5.1{\times}10.2$               & $360^\circ$         \\
    \nut      & $5$ & Nut $25.4{\times}45.7$, peg $5.1{\times}10.2$                  & $360^\circ$         \\
    \cable    & $7$ & Rope free end $40.6$ in $x$ only, each clip $40.6{\times}22.9$ & clips $\pm45^\circ$ \\
    \sqnarrow & $3$ & Nut $0.5{\times}11.5$, peg fixed                               & $360^\circ$         \\
    \sqbroad  & $5$ & Nut $23.0{\times}51.0$, peg $40.0{\times}40.0$                 & $360^\circ$         \\
    \hline
  \end{tabular}
\end{table}

\begin{figure}[!htbp]
  \centering
  \newcommand{\startpanel}[2]{%
    \begin{tikzpicture}
      \node[inner sep=0pt, anchor=north west] (img) at (0,0) {\includegraphics[width=0.495\linewidth]{#1}};
      \node[anchor=north west, inner sep=1.5pt, fill=black, fill opacity=0.6, text opacity=1, text=white, font=\scriptsize, align=left] at (img.north west) {#2};
    \end{tikzpicture}}
  \startpanel{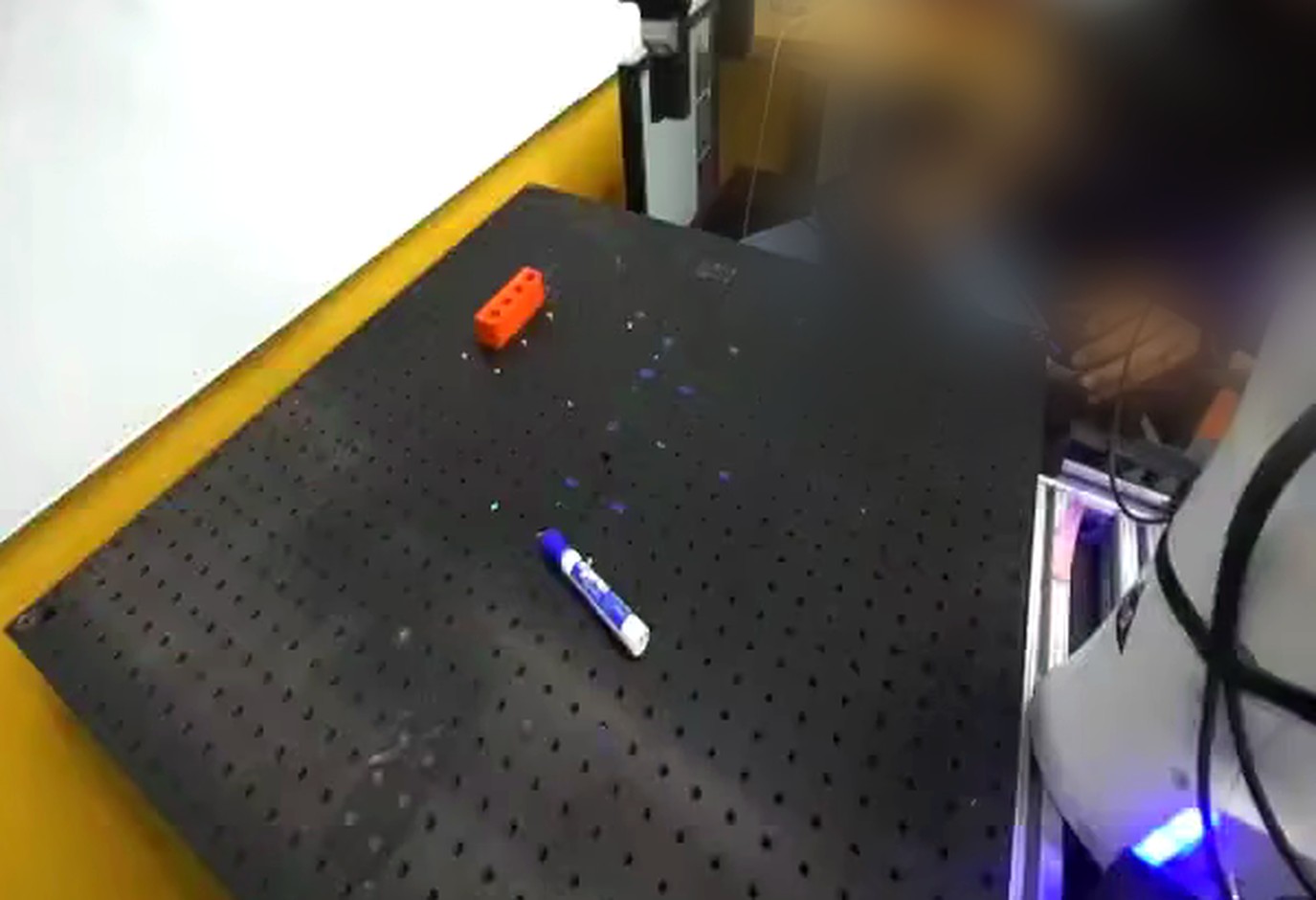}{(a) \marker, typical initial state:\\demonstration $12.6$\,s}\hfill
  \startpanel{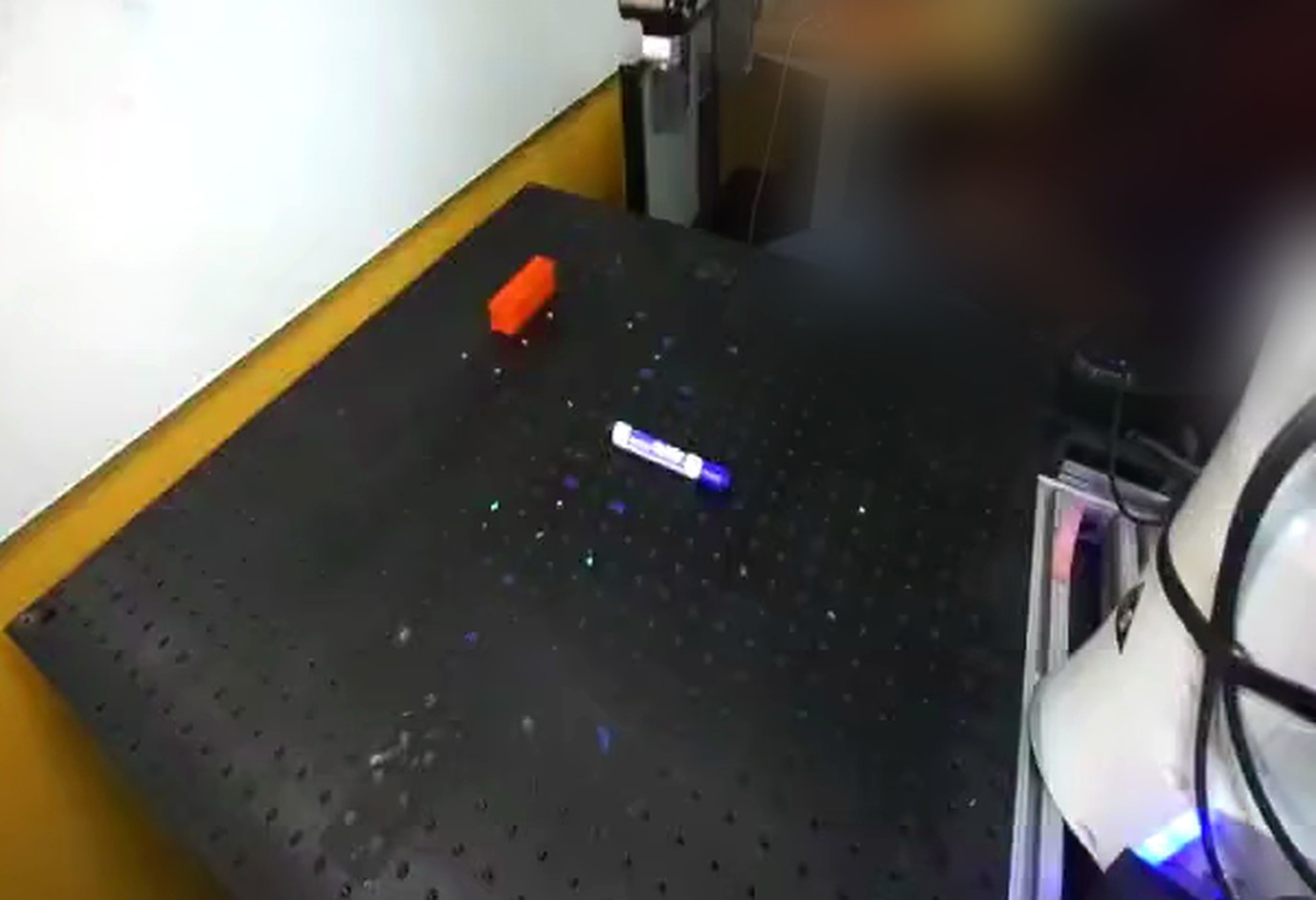}{(b) \marker, hardest initial state:\\demonstration $19.3$\,s}\\[2pt]
  \startpanel{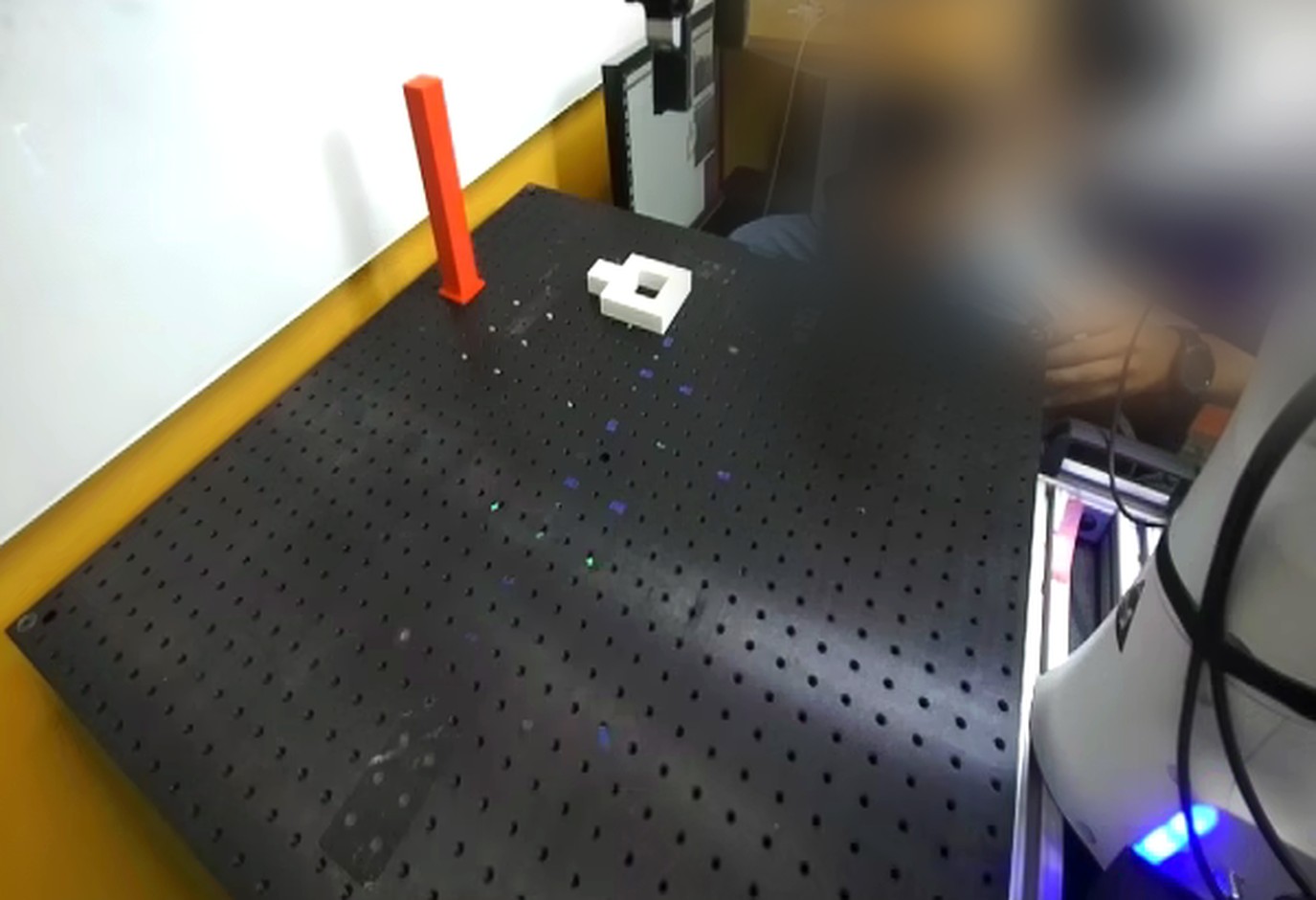}{(c) \nut, typical initial state:\\demonstration $10.1$\,s}\hfill
  \startpanel{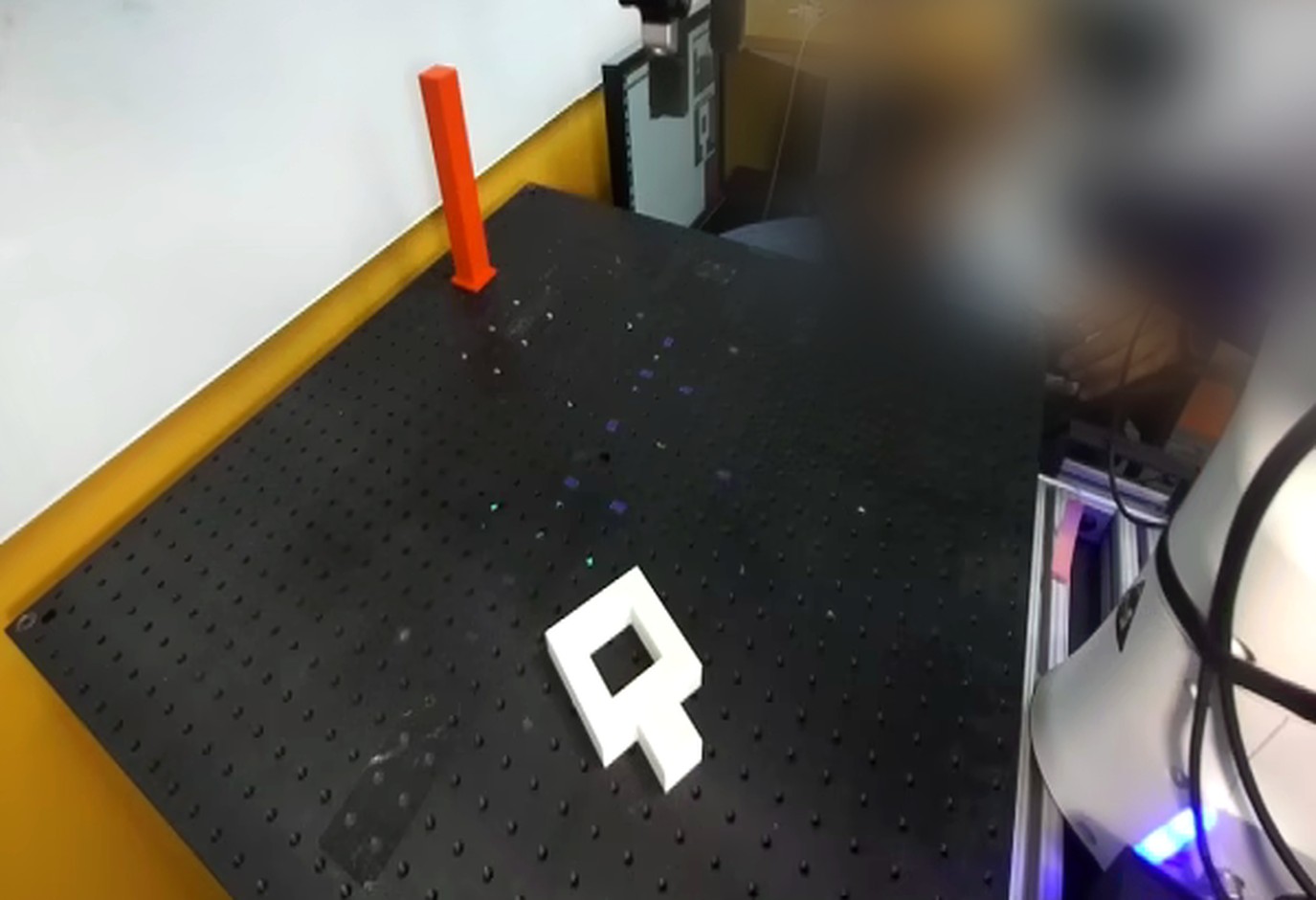}{(d) \nut, hardest initial state:\\demonstration $18.1$\,s}\\[2pt]
  \begin{tikzpicture}
    \node[inner sep=0pt, anchor=north west] (img) at (0,0) {\includegraphics[width=\linewidth]{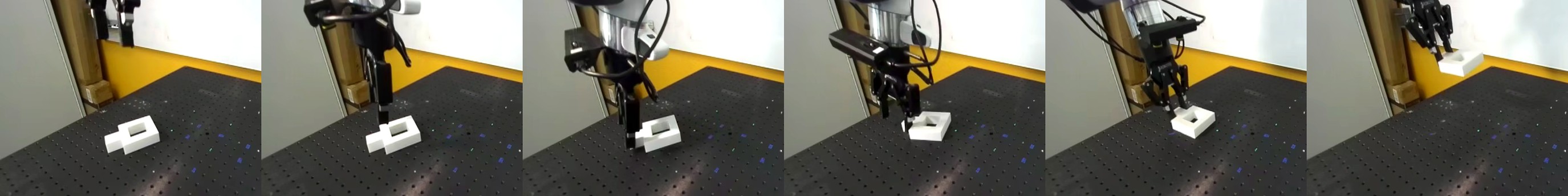}};
    \node[anchor=north west, inner sep=1.5pt, fill=black, fill opacity=0.6, text opacity=1, text=white, font=\scriptsize, align=left] at (img.north west) {(e) \nut,\\from (d)};
    \foreach \stage [count=\k from 0] in {start, finger by handle, rotate, reoriented, grasp, lift}
    \node[anchor=south, inner sep=1pt, fill=black, fill opacity=0.5, text opacity=1, text=white, font=\scriptsize] at ([xshift={(\k+0.5)*\linewidth/6}, yshift=1pt] img.south west) {\strut\stage};
  \end{tikzpicture}
  \caption{Typical and hardest initial states of \marker and \nut in round-0 teleoperation.
    (a)--(d) are the first frames from the side camera next to the robot base, which is on the right (operator blurred).
    Each hardest initial state is the first state of the task's longest round-0 demonstration, with the object near the minimum of its $x$ range and pointing toward the robot.
    Each typical initial state comes from a median-length demonstration whose object points away from the robot.
    The typical \marker initial state lies at nearly the same $x$ as the hardest one.
    (e) From the hardest \nut initial state, the robot places a finger next to the handle with the gripper yawed $180^\circ$ from the grasp orientation, rotates the nut, and then grasps it, seen from the opposite side camera.
    The three longest round-0 demonstrations of each task all start with the object within $21^\circ$ of pointing toward the robot.}
  \label{fig:76}
\end{figure}

Initial states differ widely in difficulty.
On \marker and \nut, the hardest states place the object at the minimum of its $x$ range, closest to the robot base, with the pen or the nut's handle pointing toward the robot.
On \nut these states also require a non-prehensile maneuver that rotates the nut before it can be grasped.
\cref{fig:76} shows a typical and a hardest initial state of each task and the maneuver.

\subsection{Success criteria and stage ladders}
\label{app:57}

\cref{tab:146} summarizes how each task is scored.
We pre-specify the real-world success criteria per task and score them from the evaluation video.
The binary outcome in the headline numbers is the operator's label, recorded at the end of every episode and reconciled in an outcome-editor pass before ingest.
An operator stop before the cap counts as a failure only when it is recorded as an unrecoverable state.
\cable success plots report the binary both-clips outcome, and \cref{tab:144} reports both \cable metrics.

We freeze each real-world evaluation cap per task at roughly three times the median successful policy episode length and never recompute it between rounds.
On \nut, round~0 ran with a $400$-step cap and rounds~1--3 were operator-terminated.
The frozen cap applies from round~4, so steps-to-success on \nut are comparable across rounds only from then on.
Collection episodes are operator-terminated and uncapped.

\begin{table}[!htbp]
  \centering
  \footnotesize
  \setlength{\tabcolsep}{4pt}
  \caption{Scoring and evaluation caps.
    Clip success rate is the mean \cable clip score ($0$--$2$) divided by two.
    Caps are in environment steps, at $15$\,Hz in the real world and $20$\,Hz in simulation.}
  \label{tab:146}
  \begin{tabular}{lllr}
    \hline
    Task      & Success                                  & Headline metric   & Cap             \\
    \hline
    \marker   & Marker released and fully seated (S7)    & success rate      & $800$ ($53$\,s) \\
    \nut      & Nut released and fully seated (S7)       & success rate      & $600$ ($40$\,s) \\
    \cable    & Both clips seated (S10)                  & clip success rate & $850$ ($57$\,s) \\
    \sqnarrow & Nut on the peg (Robosuite success check) & success rate      & $400$ ($20$\,s) \\
    \sqbroad  & Nut on the peg (Robosuite success check) & success rate      & $400$ ($20$\,s) \\
    \hline
  \end{tabular}
\end{table}

Episodes in the reviewed collection campaigns also carry an ordinal \emph{stage} label recording how far along the task's ladder the rollout got before it broke down.
These labels drive the earliest-breakdown-first hardness ordering of the real-world sampler (\cref{app:27}) and the per-stage decompositions of \cref{app:56}.
\cref{tab:141} lists the S0--S7 ladders of \marker and \nut, and \cref{tab:143} the S0--S10 ladder of \cable.

A vision-language model labels these stages from the episode video, blind to the policy and arm.
On \marker and \nut, we give Gemini 3.5 Flash the side and wrist-camera videos, the gripper-aperture trace, and crops of the grasp, release, and final frames.
We label each episode three times and keep the majority label.
The model cannot override the gripper caps of \cref{tab:141}, and the S7 rung always comes from the operator's outcome label.
We audit flagged and uncertain episodes by hand in a video review tool. Wherever a human label exists, it replaces the model's.
Scored before any correction, the labeler agrees with the human rung on $12$ of the $19$ \marker round-1 audit episodes and is within one rung on $18$.
For \cable, Gemini 3.7 Flash labels the $750$ held-out rollouts with no vote and no human adjudication. No reported result uses these labels.

\begin{table}[!htbp]
  \centering
  \footnotesize
  \setlength{\tabcolsep}{4pt}
  \caption{Stage ladders of the two binary real-world tasks.
    The labeler reports the furthest rung reached.
    S7 is the success predicate.
    An episode whose jaws never close is capped at S1, and one whose jaws never reopen at S3.
    On \marker, an episode that ends with the marker back on the table is capped at S2.}
  \label{tab:141}
  \begin{tabular}{lp{5.6cm}p{5.6cm}}
    \hline
    Rung & \marker                                                                                                   & \nut                                                                                         \\
    \hline
    S0   & No useful approach toward the marker.                                                                     & No useful approach toward the nut.                                                           \\
    S1   & Approach or grasp attempt without a usable grasp (missed close, or an end pinch that cannot be inserted). & Reach or grasp attempt without a grasp that can lift and carry the nut.                      \\
    S2   & Marker acquired with a usable grasp and lifted clear of the table.                                        & Nut grasped and lifted clear of the table, even if later lost.                               \\
    S3   & Marker brought to the holder area; pressing the holder body without the tip engaging the hole stays S3.   & Held nut reaches the peg region in an insertion-directed pose; hole not verified on the peg. \\
    S4   & Marker tip engages the hole opening but is not fully seated.                                              & Nut's hole visibly engages or aligns over the peg top; not seated at the base.               \\
    S5   & Gripper released a partial insertion that the holder retains.                                             & Gripper released a partial insertion; the nut hangs on the peg above the base.               \\
    S6   & Marker fully seated while still held at episode end.                                                      & Nut fully seated at the peg base but still held.                                             \\
    S7   & Marker released and remains fully seated (success).                                                       & Nut released and remains fully seated (success).                                             \\
    \hline
  \end{tabular}
\end{table}

\begin{table}[!htbp]
  \centering
  \footnotesize
  \setlength{\tabcolsep}{4pt}
  \caption{Stage ladder of \cable.
    S10 is the both-clips success predicate.}
  \label{tab:143}
  \begin{tabular}{lp{8cm}}
    \hline
    Rung   & \cable                                       \\
    \hline
    S0     & No useful approach.                          \\
    S1     & Approach or contact without a usable grasp.  \\
    S2     & A grasp sufficient to route.                 \\
    S3--S5 & Reach, align, and engage the first clip.     \\
    S6     & One clip seated.                             \\
    S7--S9 & Reach, align, and engage the remaining clip. \\
    S10    & Both clips seated (success).                 \\
    \hline
  \end{tabular}
\end{table}

\clearpage\suppressfloats[t]

\section{Evaluation Protocol Details}
\label{app:25}
\backlink{\cref{subsec:114}}

Task-specific success criteria and stage labeling are in \cref{app:57}.

We follow recommended practice for physical-robot policy evaluation~\cite{kressgazit2024evaluation}, with matched initial conditions across compared policies, pre-specified success criteria, metrics that complement success rate, and intervals in addition to point estimates.

\paragraph{Separating evaluation from collection.}
Interventions during DAgger collection could double as failure labels, but we keep evaluation and collection in separate phases, for four reasons:
\begin{itemize}
  \item Every policy in a round is scored on the same initial states, so comparisons are paired.
  \item Evaluating on Sobol initial states gives a lower-variance estimate of the average success rate.
        The initial states that performance-guided collection picks skew toward hard ones, and uniform random draws have high variance at these sample sizes.
  \item Evaluation rollouts feed the next round's choice of initial states and enter value learning directly.
  \item Operators correct before an irrecoverable mistake, so some interventions fall on episodes that would have succeeded, which inflates the failure rate measured during collection.
\end{itemize}

In simulation, we score each policy on a fixed held-out evaluation set, reused across rounds, using the final checkpoint at a pre-committed training-step count (no post-hoc checkpoint selection).
In the real world, each round scores all of its policies together on a shared held-out block of $50$ scrambled-Sobol initial states, disjoint from every initial state used for collection.
On \marker and \nut, each round uses a fresh block, and we retire used evaluation seeds into the sampler's design pool instead of reusing them.
Because we score each round on a new block, and the environment can drift from day to day, success rates are comparable across methods within a round but not directly across rounds.
On \cable, we instead scored the $15$ policies of all rounds together on one block, drawn after the last collection round and disjoint from every collection and earlier evaluation state, so round-to-round comparisons on that task are also paired on the same initial states and less exposed to drift between sessions.

\paragraph{Rounds scored over several sessions.}
In two rounds, we scored the reranked policy in later sessions on new blocks, after the round's actor block was complete.
Each plotted success rate pools all blocks on which that checkpoint was scored in its round, since every block is an independent Sobol draw from the same held-out distribution.
Paired tests use only initial states on which both compared methods ran within one session.
HG-DAgger+\ours{}, the one method scored in every session, succeeds on $23$ and $24$ of $50$ initial states in the two \marker R2 sessions and on $29$, $23$, and $25$ of $50$ in the three \nut R3 sessions.
When we restrict \nut R3 to its last session, the only one that scores all three methods, HG-DAgger succeeds on $20/50$ and HiL-IDQL+\ours{} on $31/50$, compared with $49/100$ and $62/100$ pooled.

\begin{figure}[t]
  \centering
  \includegraphics[width=\linewidth]{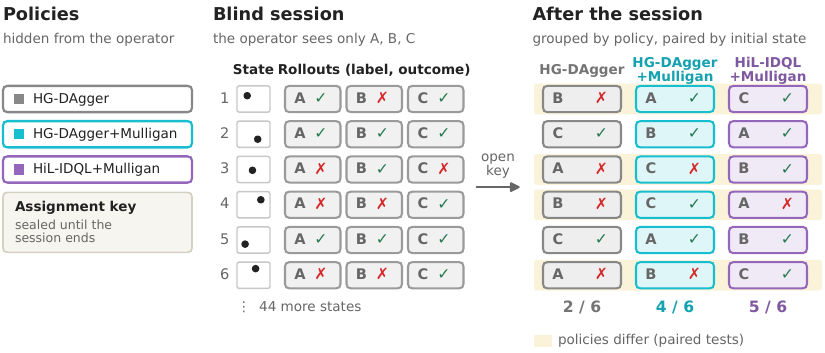}
  \caption{\textbf{Blinded, paired evaluation} (schematic, shuffles and outcomes are illustrative).
    The operator places the objects at each initial state of the held-out Sobol block, and every compared policy runs from that placement under an anonymous label, with the labels reshuffled for every initial state.
    Outcomes are recorded under the labels only.
    After the session, the sealed assignment is opened, and the outcomes are regrouped by policy, one row per shared initial state.
    Success rates come from the column totals, and the paired tests compare two policies on the initial states where their outcomes differ (shaded rows).}
  \label{fig:74}
\end{figure}

\paragraph{Blinded protocols.}
All teleop collection, DAgger corrections, and evaluations are blinded.
We combine the initial states of all arms into one scrambled list and run the compared policies under anonymous labels on this shared set.
This hides the arm from the operator and spreads environmental drift, such as lighting and motor temperature, across arms.
\cref{fig:74} illustrates one evaluation session.
\citet{kressgazit2024evaluation} report a policy dropping from $13/20$ to $0/8$ on \emph{identical} initial conditions after an unrelated change to the lab, so we interleave compared policies within one session.
Sobol evaluation states cover the space more evenly than uniform samples (\cref{fig:77}), and on \marker and \nut a fresh block every round keeps iterative development from overfitting to a fixed set of initial states.
To place objects quickly and blindly, the operator follows a per-episode reset-helper schematic (\cref{fig:93}).
We release video of every evaluation episode behind the compared methods' reported results, alongside per-episode outcomes and the initial-state list each session was scored on, so readers can check the pairing across arms, the blinding, and the initial states realized at reset.

\paragraph{Uncertainty quantification.}
For simulation policy-performance outcomes, we report two-sided Student-t 95\% confidence intervals across five training seeds, each estimated from a large number of trials.
In the real world, each plotted policy is one training run, so per-arm whiskers describe evaluation-sampling variability rather than training-run variability.
The Wald binomial standard error~\cite{wald1943,brown2001interval} degenerates near 0\% and 100\%, so for binary success rates we use the Wilson score interval~\cite{wilson1927} with $z{=}1$ as a one-standard-error interval.
The \cable clip success rate is not binary.
Each episode scores $0$, $1/2$, or $1$, so its whiskers are the standard error of the mean score over episodes.
Real-world comparisons use paired per-state statistics, with exact McNemar tests~\cite{mcnemar1947} on discordant pairs and 95\% bootstrap confidence intervals~\cite{efron1979bootstrap} on the paired success-rate delta (\cref{app:55}).

\paragraph{Released evaluations.}
The Hugging Face release (\url{https://huggingface.co/mulligan}) contains 2,550 evaluation episodes of the headline methods: 850 on \marker, 950 on \nut, and 750 on \cable.
With the no-CF ablation, it contains 2,850.
Two further datasets hold the superseded \marker and \nut R5 screening evaluations ($400$ episodes).
All three \nut R5 arms are scored on a later fresh Sobol block, because the critic trained on rollouts from the earlier R5 initial states.
The Policy Arena at \url{https://arena.mulligan.page} lets readers browse these episodes and the per-session comparisons.

\clearpage\suppressfloats[t]

\section{Data and Collection Protocol}
\label{app:20}

\subsection{The four data streams and how frames are routed}
\label{app:23}

Supervised deployment produces four streams: base demonstrations, human corrections during a rollout, autonomous successes, and autonomous failures.
Every frame carries a \emph{source} tag $\in \{\text{policy}, \text{human}\}$ and a binary \emph{intervention} flag, and every completed episode carries an \emph{outcome} $\in \{\text{success}, \text{fail}\}$.
These three tags decide which frames reach the actor and which reach the critic. We never select frames by hand.

\begin{figure}[t]
  \centering
  \includegraphics[width=\linewidth]{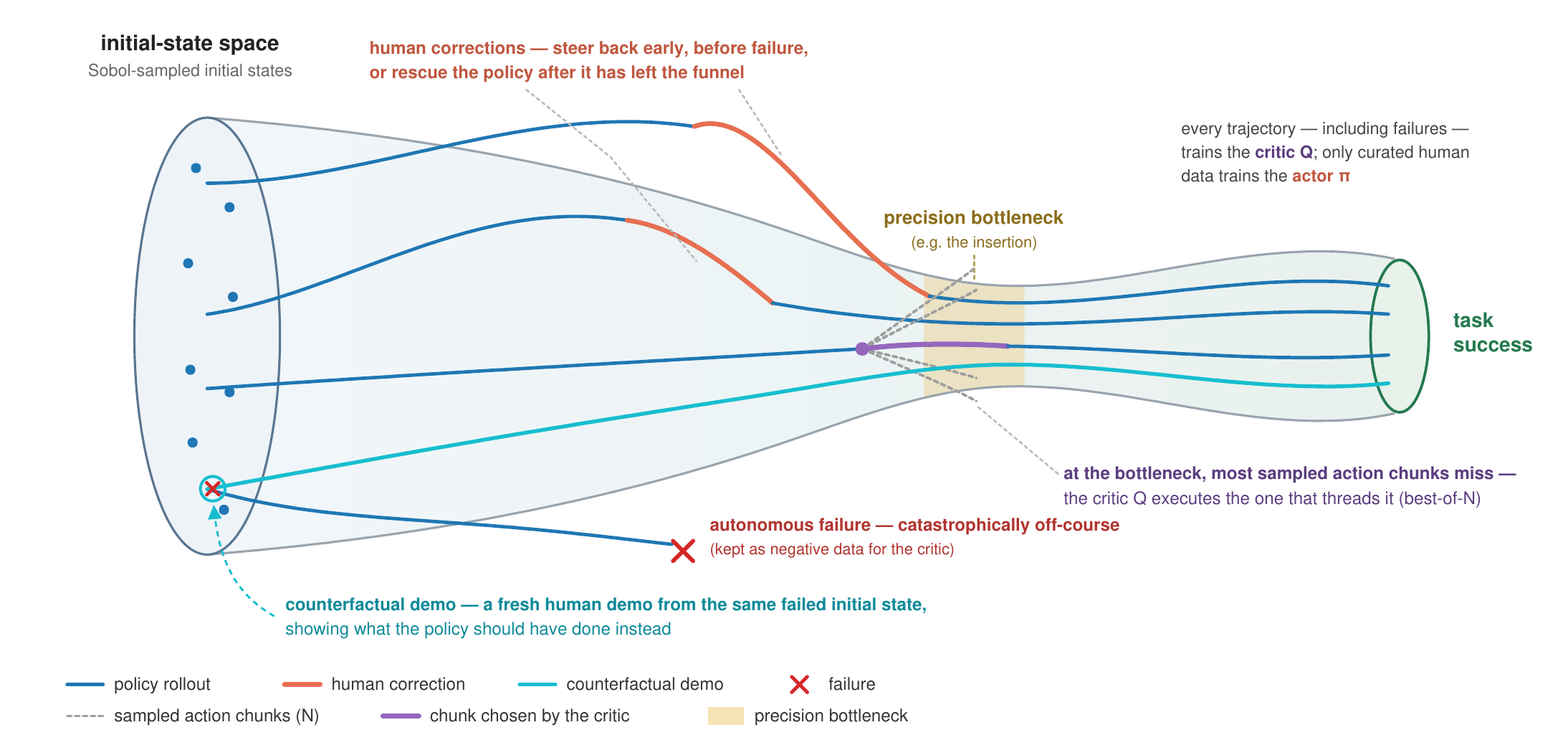}
  \caption{\textbf{One supervised deployment round as a trajectory funnel.} Each rollout starts from a Sobol-sampled initial state and succeeds only if it passes a precision bottleneck such as the insertion.
    The operator can correct \emph{early}, steering the policy back before it fails, or \emph{rescue} it after it has left the funnel.
    When a rollout fails without intervention, we keep it as negative data for the critic, and the operator may record a counterfactual demonstration from its initial state.
    Most action chunks sampled at the bottleneck miss. Best-of-$N$ reranking with the IQL critic executes one that passes.}
  \label{fig:75}
\end{figure}

\cref{fig:75} shows one supervised deployment round.
The actor trains only on $\mathcal{D}_h$, the human frames from \emph{successful} trajectories (base demonstrations, corrections that rescued an episode, and counterfactual demonstrations).
The critic trains on $\mathcal{D}_\text{all}$, every frame of every evaluation rollout and collection episode, whether human or policy and whether the episode succeeded or not.
Autonomous successes and failures, which interactive imitation alone discards, reach only the critic.

\subsection{Round budgets and dataset ledger}
\label{app:22}

Round~0 records $100$ teleoperated demonstrations per arm on every task except \sqbroad, which records $200$.
Each later real-world round collects $M = 50$ HG-DAgger episodes per arm on \marker and \nut and $M = 100$ on \cable, and is evaluated on a held-out block of $50$ initial states shared across all arms in that round.
On \cable, each of R1--R4 pools two $50$-episode collection sessions and R5 is a single $100$-episode session. We sum frame counts over a round's sessions.
Each later simulation round collects $M = 100$ episodes per arm.
\cref{tab:123} gives the per-task budgets and final-round training-set sizes.
$\mathcal{D}_h$, the actor's set, contains every episode with human-controlled frames (demonstrations, corrected episodes, and counterfactual replays). Its frame count includes only the human-controlled frames, since those are what the actor trains on.
$\mathcal{D}_\text{all}$ adds the autonomous frames of every collected episode.
The critic also trains on autonomous rollouts of earlier policies (held-out evaluations in the real world, separate diagnostic rollouts in simulation), so we list its final-round set separately.
Frame counts of $\mathcal{D}_h$ and $\mathcal{D}_\text{all}$ are valid frames after the outcome-edited prefix. Critic-set frames are the recorded lengths of the training datasets. None of these counts measure operator wall-clock time.

\begin{table}[tbp]
  \centering
  \footnotesize
  \setlength{\tabcolsep}{3pt}
  \caption{Collection budgets and final-round training-set sizes for the HG-DAgger+\ours{} arm.
    $M$ counts episodes per round per arm after round~0, which records $100$ demonstrations per arm ($200$ on \sqbroad).
    ``$\mathcal{D}_h$'' is the actor's training set (episodes with human frames, with the human-controlled frame count), ``$\mathcal{D}_\text{all}$ frames'' counts all valid frames of the collected episodes, ``critic set'' is the final-round critic's training set including rollouts of earlier policies, and ``collected'' counts every recorded round-0 and collection episode across both the uniform-sampling and \ours{} arms.}
  \label{tab:123}
  \begin{tabular}{llcccccc}
  \hline
  Task & Rounds & $M$ & \shortstack{$|\mathcal{D}_h|$\\episodes} & \shortstack{$\mathcal{D}_h$\\frames} & \shortstack{$\mathcal{D}_\text{all}$\\frames} & \shortstack{Critic set\\(episodes / frames)} & \shortstack{Collected\\episodes} \\
  \hline
  \marker & R0--R5 & $50$ & $294$ & $38{,}017$ & $81{,}526$ & $825$ / $262{,}164$ & $824$ \\
  \nut & R0--R5 & $50$ & $301$ & $34{,}536$ & $66{,}354$ & $900$ / $209{,}953$ & $824$ \\
  \cable & R0--R5 & $100$ & $585$ & $112{,}292$ & $215{,}475$ & $1{,}200$ / $548{,}588$ & $1{,}334$ \\
  \sqnarrow & R0--R3 & $100$ & $229$ & $25{,}216$ & $70{,}563$ & $700$ / $130{,}983$ & $800$ \\
  \sqbroad & R0--R3 & $100$ & $443$ & $57{,}983$ & $100{,}781$ & $900$ / $208{,}219$ & $1{,}000$ \\
  \hline
\end{tabular}

\end{table}

$|\mathcal{D}_h|$ leaves out the collection episodes that the policy completed without a correction: $56$ on \marker, $49$ on \nut, $15$ on \cable, $171$ on \sqnarrow, and $57$ on \sqbroad.

\subsection{Counterfactual-demonstration protocol}
\label{app:14}

A counterfactual demonstration is a full human demonstration from the \emph{same} initial state at which the policy just failed, with no policy frames (\cref{sec:108}).
An HG-DAgger correction picks up from whatever state the policy reached on its own, which limits it to teaching recovery.
For a counterfactual, the operator resets the scene to the recorded initial state and teleoperates the whole task.

A counterfactual costs a full demonstration of operator time, so the operator records one by choice, not after every failure.
The collection software offers the option after every episode the operator drove to completion. The operator takes it when the correction did not itself show the intended behavior (\cref{sec:108}).
Each counterfactual counts toward the with-CF arm only and displaces the last coverage-fill initial state from that arm's $M$-episode budget, so we can compare the arms with and without CFs at the same total number of episodes.
\cref{tab:119} lists the counterfactuals per round.
The operator recorded most of them at promoted failure states, and about a quarter at coverage-fill states where the policy also failed.

\begin{table}[tbp]
  \centering
  \footnotesize
  \caption{Counterfactual demonstrations per collection round in the HG-DAgger+\ours{} arm, out of $50$ episodes on \marker and \nut and $100$ on \cable.}
  \label{tab:119}
  \begin{tabular}{lccccc}
    \hline
    Task    & R1   & R2   & R3   & R4   & R5   \\
    \hline
    \marker & $12$ & $15$ & $14$ & $9$  & $9$  \\
    \nut    & $17$ & $16$ & $15$ & $15$ & $9$  \\
    \cable  & $13$ & $24$ & $11$ & $10$ & $12$ \\
    \hline
  \end{tabular}
\end{table}

\clearpage
\subsection{Reset-helper schematics for real-world collection}
\label{app:62}
\backlink{\cref{sec:111}}

\begin{figure}[!h]
  \centering
  \begin{minipage}[t]{0.32\linewidth}
    \centering
    \includegraphics[width=\linewidth]{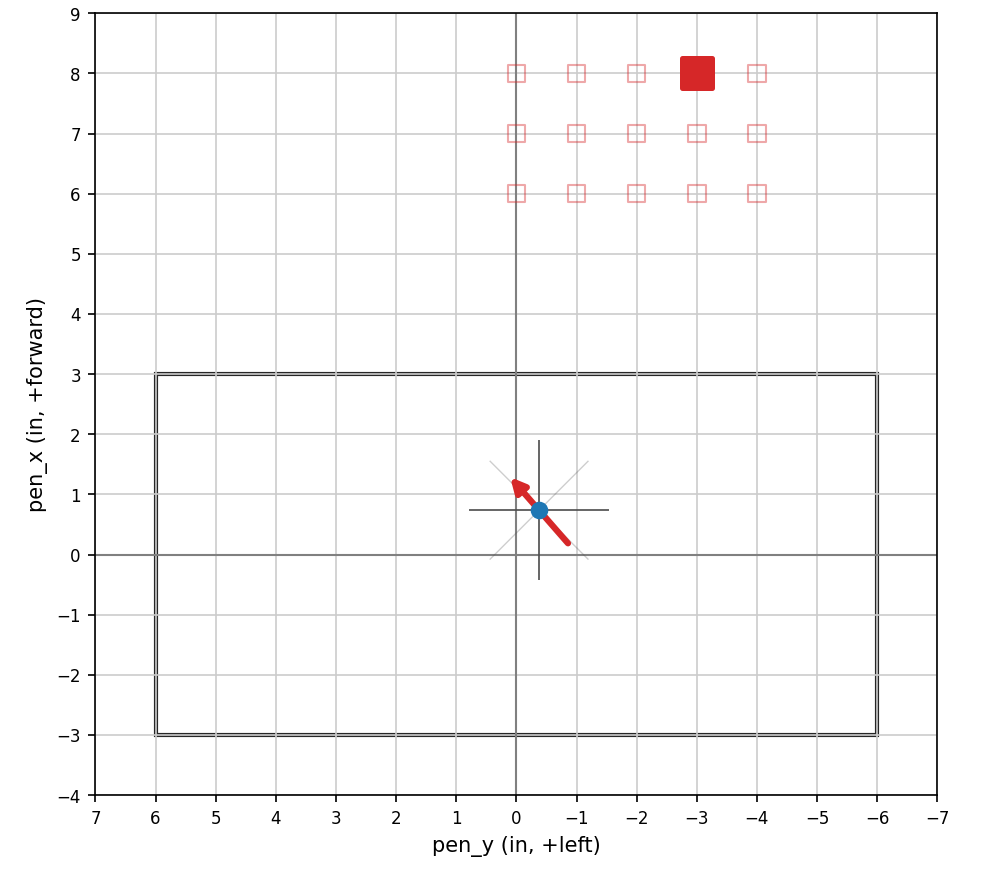}
  \end{minipage}\hfill
  \begin{minipage}[t]{0.32\linewidth}
    \centering
    \includegraphics[width=\linewidth]{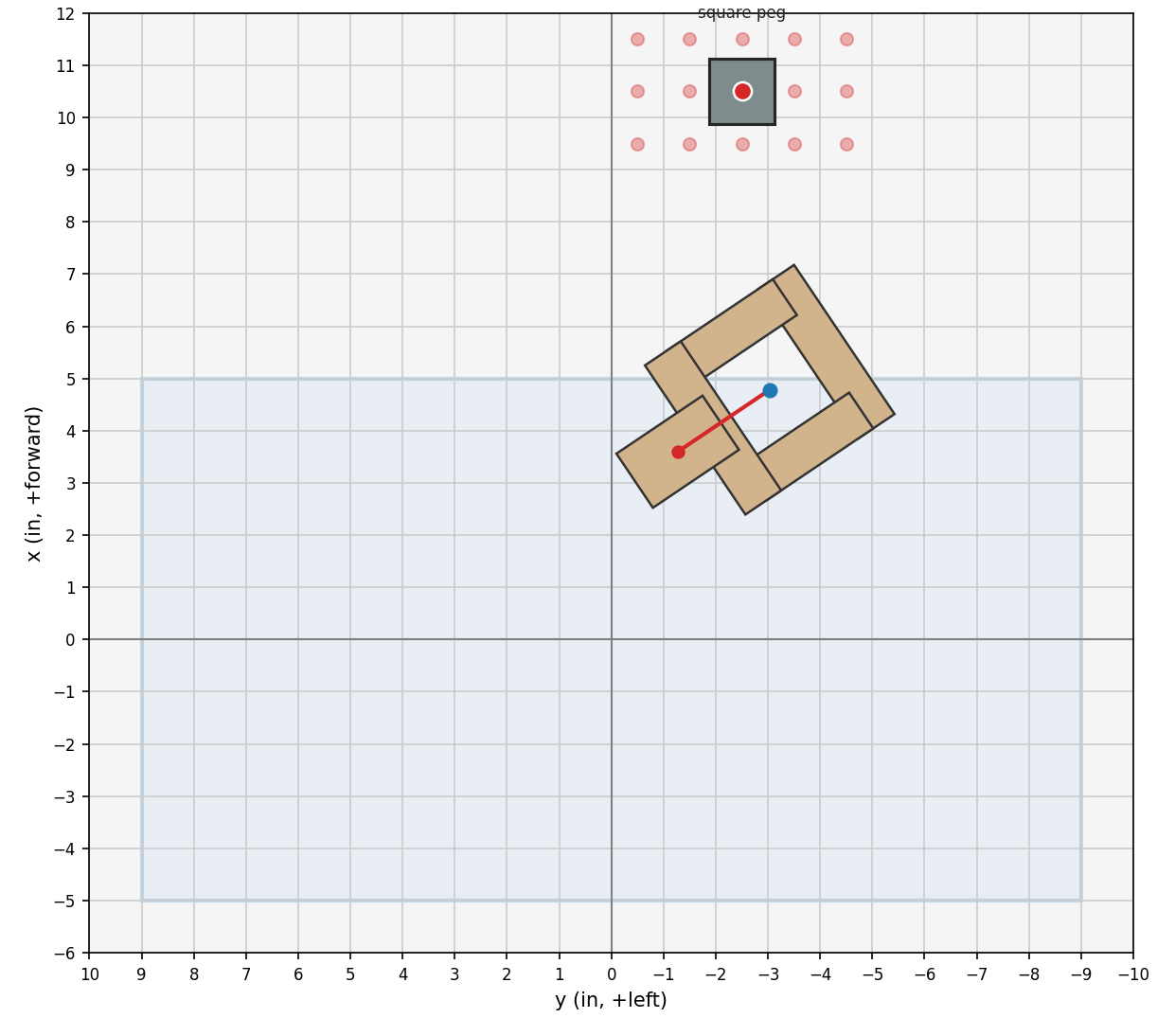}
  \end{minipage}\hfill
  \begin{minipage}[t]{0.32\linewidth}
    \centering
    \includegraphics[width=\linewidth]{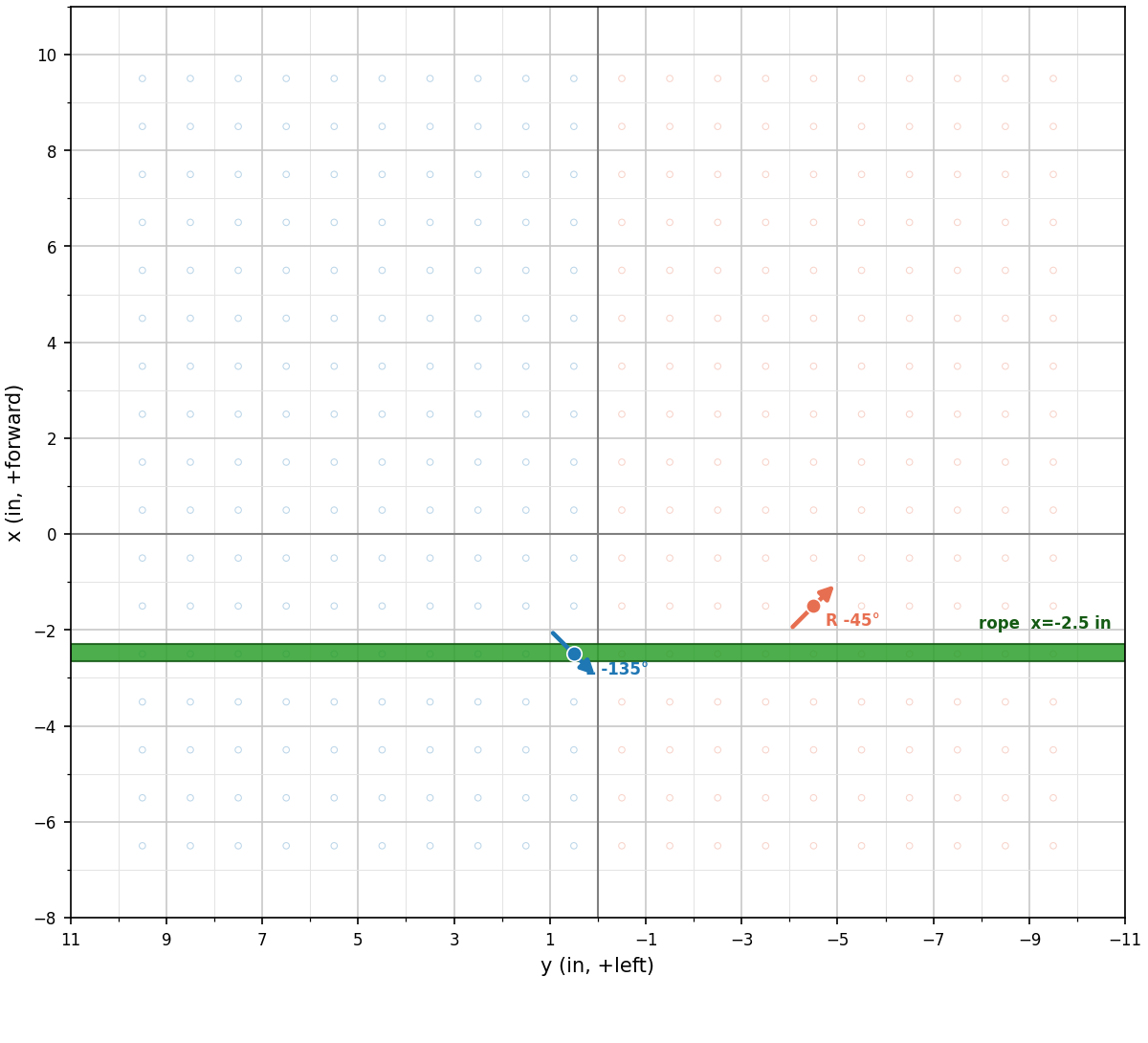}
  \end{minipage}
  \caption{Per-episode reset-helper schematics for placing objects at each sampled initial state.
    Each card shows the allowed object-center grid, the chosen locations, and the sampled object poses (position and yaw) in table coordinates.
    Left: \marker (pen + holder).
    Middle: \nut (nut + peg).
    Right: \cable (cable + two clips).}
  \label{fig:93}
\end{figure}

\cref{fig:93} shows the reset-helper schematic of each real-world task.

\clearpage\suppressfloats[t]

\section{Initial-State Sampling Details}
\label{app:50}
\backlink{\cref{sec:111}}

\subsection{Difficulty varies sharply across the initial-state space}
\label{app:10}

\begin{figure}[t]
  \centering
  \includegraphics[width=\linewidth]{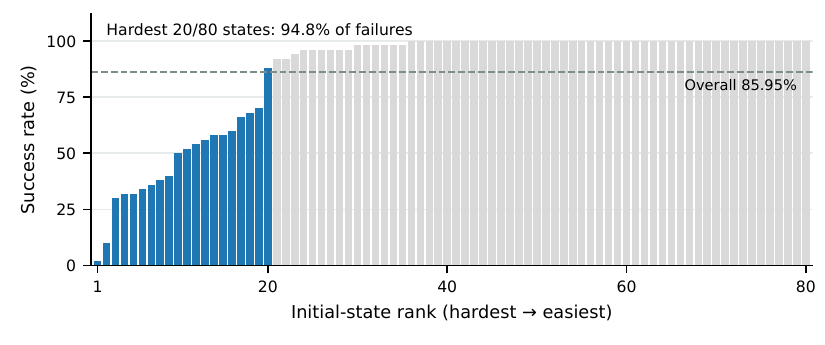}
  \caption{Success rate by initial state on \sqnarrow ($80$ lattice initial states over the full reset range, $50$ rollouts each, sorted) for a round-2 policy trained with coverage-only (Sobol) sampling, seed 1 of three.
    Overall success is $85.95\%$ (dashed line), and the hardest $25\%$ of the states account for $94.8\%$ of failures.
    We recovered the per-state success counts exactly from the original evaluation plot, whose bars fall on the $2$-point grid that $50$ rollouts allow.}
  \label{fig:69}
\end{figure}

As \cref{fig:80} and \cref{fig:69} show, not all of $\mathcal{S}_0$ is equally difficult, which is the premise of \cref{sec:111}.
We evaluate a round-2 \sqnarrow policy from $80$ fixed initial states on a lattice spanning the full reset range ($2$ $x$-values $\times$ $5$ $y$-values $\times$ $8$ yaws), with $50$ rollouts from each.
Overall the policy succeeds $86\%$ of the time. Most initial states sit at or near $100\%$, and a small tail of position/yaw combinations fails most of the time.

This policy comes from the coverage-only arm of that campaign, so where its data were collected cannot explain the structure.
It was trained on $100$ base demonstrations and two rounds of $100$ supervised episodes each, and all $300$ initial states were drawn from the same scrambled Sobol sequence over the full reset range, with no failure promotion and no hardness weighting.
In both collection rounds the operator followed the standard protocol and chose to record $20$ and $22$ counterfactual demonstrations.
Even with the same collection density everywhere in $\mathcal{S}_0$, the failures concentrate.
Uniform sampling spends effort in proportion to volume and puts most of it on states the policy has already mastered.
\textsc{SelectInitialStates} (\cref{alg:2}) moves that budget to the failing tail.

Only the two endpoints of the $5$\,mm $x$ range enter the lattice.
On the same lattice, the other two training seeds reach $77.9\%$ and $81.9\%$.

\subsection{Sobol vs.\ uniform coverage}
\label{app:54}
\backlink{\cref{sec:111,subsec:116}}

\paragraph{Low-discrepancy sequences.} A low-discrepancy sequence spreads every prefix evenly over $[0,1]^d$.
Its \emph{star discrepancy}, the largest gap between the fraction of points in an anchored box and the volume of that box, falls as $\mathcal{O}((\log n)^d / n)$.
I.i.d.\ uniform samples fluctuate at $\mathcal{O}(n^{-1/2})$ and, at our budget sizes ($n = 50$--$100$), leave clumps and holes.
The Sobol sequence~\citep{sobol1967,joe2008sobol} balances each block of $2^m$ consecutive points exactly across recursive halvings of the domain. We materialize draws in base-2 blocks to keep that balance and slice them by index range (\cref{app:53}).
Owen scrambling~\citep{owen1998scrambling} randomly permutes the binary digits of each coordinate without breaking this balance. Each scrambled point is marginally uniform, which makes a scrambled draw an unbiased sample of $\mathcal{S}_0$, and independent scramblings give independent replicates.

\begin{figure}[t]
  \centering
  \includegraphics[width=\linewidth]{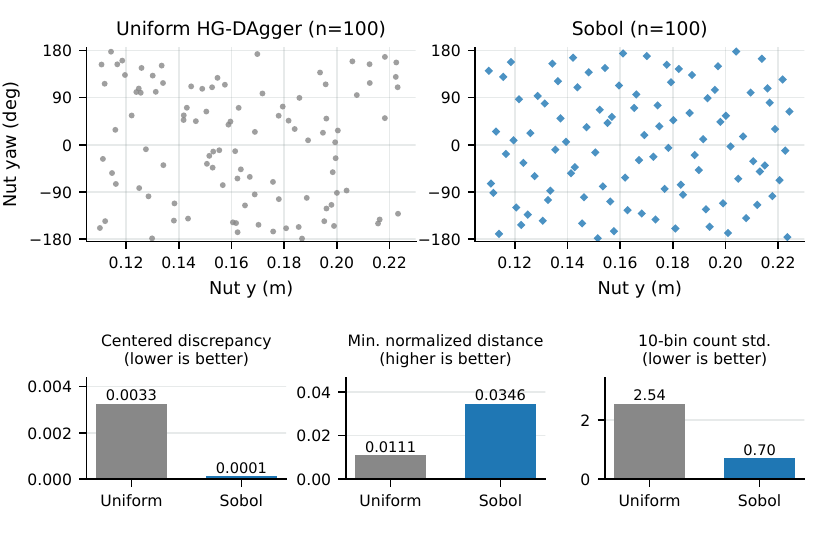}
  \caption{\sqnarrow R0 initial-state coverage of uniform HG-DAgger (left) and Sobol (right), $n=100$ each, over nut $y$ and yaw.
    Bottom: centered discrepancy (distinct from the star discrepancy defined above), minimum Euclidean nearest-neighbor distance on the normalized unit square, and the mean of the two marginal count standard deviations over ten equal-width bins.}
  \label{fig:77}
\end{figure}

\cref{fig:77} plots the \sqnarrow R0 initial states of the uniform and Sobol arms compared in \cref{subsec:116}.
The Sobol initial states cover nut $y$ and yaw more evenly. All three statistics at the bottom of the figure agree, with lower centered discrepancy, a larger minimum nearest-neighbor distance, and flatter marginal bin counts for Sobol.

\subsection{Sobol sequence configuration}
\label{app:53}

All coverage draws use a scrambled Sobol sequence~\citep{sobol1967,owen1998scrambling,joe2008sobol} over the per-task initial-state bounds, materialized in base-2 blocks and sliced by index range.
The dimensions each sequence spans are:
\begin{itemize}
  \item \textbf{\sqnarrow (sim):} the initial-state space is 3-D (nut $x$, $y$, yaw), but the deployed sequence is 2-D over nut $y$ and yaw, from a fixed $1024$-point pool at scramble seed $42$.
        Nut $x$ is drawn independently uniform from its own seed, and each round consumes the next contiguous block of stream indices.
  \item \textbf{\sqbroad (sim):} 5-D, adding peg $x$ and $y$, with rejection of nut/peg placements that collide or are infeasible.
  \item \textbf{Real (\nut, \marker, \cable):} a Sobol draw over each task's placement dimensions (e.g.\ nut $x$, $y$, and yaw for \nut, and the rope's free $x$-position and each clip's $x$, $y$, and yaw for \cable), snapped to the physical dot grid where a task uses discrete placements.
\end{itemize}
A seed ledger keeps the initial states used for collection and for evaluation disjoint.
In simulation, the headline evaluation sets are fixed and reused across rounds, and the diagnostic rollouts whose failures are promoted into collection (\cref{sec:111}) draw a fresh disjoint Sobol range each round.
On \marker and \nut, when a failed evaluation state is promoted into collection, we retire its seed, and the next round's evaluation reserves a new disjoint range (\cable reuses one block, \cref{app:25}).

\subsection{The shared sampler}
\label{app:49}

Both domains run \textsc{SelectInitialStates} (\cref{alg:2}), with the coverage fill of the main text extended by an optional hardness tilt.
The fill maximizes $d(c,\, \mathcal{P} \cup \mathcal{G} \cup \mathcal{X})\,(1+\beta\,h(c))$, where $d(c, \cdot)$ is the nearest-neighbor distance from candidate $c$ to the set of previously collected initial states $\mathcal{P}$, promoted failures $\mathcal{G}$, and fills selected so far $\mathcal{X}$, and $h(c) \in [0,1]$ is a task-specific hardness estimate (\cref{app:28,app:27}) with weight $\beta \ge 0$.
With $\beta = 0$, the fill is farthest-point coverage~\cite{gonzalez1985clustering,eldar1997fps}. For small $\beta$ the distance factor dominates, so coverage is preserved while supervision tilts toward hard regions.

\begin{figure}[t]
  \centering
  \includegraphics[width=\linewidth]{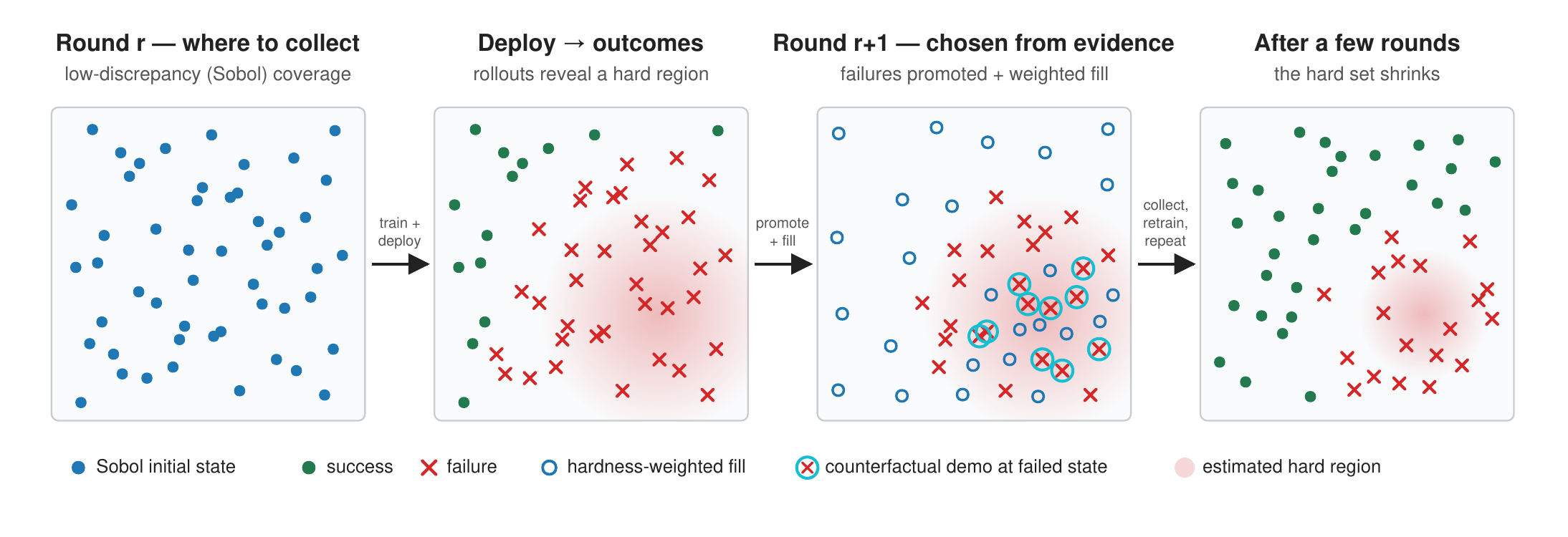}
  \caption{\textbf{Performance-guided initial-state design across rounds} (schematic, points are scrambled-Sobol draws).
    A collection round covers the joint initial-state space with a Sobol set.
    Deploying the trained policy reveals a hard region (red).
    The next round's collection set promotes the observed failures verbatim and spends the remaining budget on the coverage fill, optionally hardness-weighted.
    During collection, initial states where the policy fails again can receive a counterfactual demonstration.
    Over rounds, the hard set shrinks.}
  \label{fig:101}
\end{figure}

The farthest-point and nearest-neighbor computations are identical across tasks.
Only the hardness signal $h$, its weight $\beta$, and the promotion cap $m_f$ vary by domain. The subsections below give each.
\cref{fig:101} shows how the set of initial states evolves over rounds.

\subsection{Hardness signal: simulation}
\label{app:28}

The simulation hardness signal comes from the DAgger episodes and diagnostic rollouts of the current policy, two streams that real-world deployment also provides.
After each round, we inspect where diagnostic failures and DAgger interventions concentrate in the initial-state space and summarize this as a geometric heuristic $h(c) \in [0,1]$ over the initial-state parameters (on \sqnarrow, a weighted sum of nut-$y$ and yaw terms in $[0,1]$ with weights summing to one, and on \sqbroad, a ridge regression of diagnostic episode length on seven features of the nut and peg placement, min-max normalized over the candidate pool).
We pick the weight $\beta$ during the same inspection, typically in $[0.5, 1.0]$. It has to tilt the fill visibly toward the failure-dense regions and still pass the coverage guardrails of \cref{app:11}.

In three rounds (\sqnarrow R3 and \sqbroad R2--R3), the \ours{} fill also used a term derived from evaluation. This term is not part of the method in \cref{sec:111} and has no real-world counterpart.
It gave each candidate initial state a hardness bonus based on the previous round's \ours{} success in that state's cell of the fixed headline evaluation set ($80$ cells on \sqnarrow, $720$ on \sqbroad).
On \sqnarrow R3 the bonus carried weight $1.0$ next to the geometric term at $0.5$. On \sqbroad ($\beta = 1.0$) it replaced the fitted term at R2 and was averaged with it at R3.
The initial states used for collection stay disjoint from those used for evaluation. The bonus still steers collection toward the evaluation cells where the previous policy failed, information the uniform baseline never sees.
We checked whether this inflated the gap between each learner with and without \ours{} (\cref{tab:140}). The gaps at these rounds are not systematically larger.
On \sqnarrow, R3 has the smallest gap of the four rounds, $+1.8$ and $+1.5$ points against $+3.0$ to $+8.5$ at R0--R2. The \sqbroad R3 gap ($+7.6$, $+7.8$) sits below R1's $+8.5$ for both learners.
Only \sqbroad R2 ($+11.1$, $+11.8$) exceeds the rounds without the term, and its intervals overlap those of R1.

Diagnostic failures also feed the next round directly through verbatim failure promotion (\cref{alg:2}).
Once promoted, an initial state is retired, and the next round's diagnostic rollouts draw from a fresh, disjoint Sobol range (\cref{app:53}).
Because the simulation budget covers every observed failure, we do not order promotions by severity (cf.\ \cref{app:27}).
We report simulation success rates on the fixed Sobol evaluation sets of \cref{sec:105}.
A separate fixed set of 400 initial states only monitors training and never enters the sampler.

\subsection{Hardness signal: real world}
\label{app:27}

The real world gives $n \approx 50$ evaluation episodes per round, too few rollouts per state to estimate per-cell success rates. We define hardness by \emph{failure stage} instead.
We label each evaluation failure by the stage at which the rollout broke down (e.g.\ grasp $\to$ transport $\to$ insert $\to$ seat $\to$ release) with the vision-language stage labeler of \cref{app:57}.
A state whose rollout broke down earlier is harder, because success requires passing every stage in order. Verbatim promotion (up to the cap $m_f$) takes failures \emph{earliest-breakdown-first}.
\cable collection before the second half of round~4 used a different order, by clip score and then time to failure.
We keep $\beta = 0$ in the coverage fill unless a geometric hardness term predicts failure on that task's earlier rollouts.
At round~5 this gave $\beta = 0.5$ on \nut and $\beta = 0$ on \marker, where rounds~3 and~4 had run at $0.05$ before we dropped the term. \cable used $\beta = 0$ through round~4 and $\beta = 1.0$ at round~5.
In the farthest-point distance, we down-weight dimensions that show no hardness signal. The fill then spreads mainly over the axes that do, such as the object's pose.

On \cable, we tested eight pre-registered hardness terms on the rollouts before the round-4 design. None predicted failure consistently across rounds and arms, and the fills through round~4 were pure coverage under a stratified quota.
At round~5, a descriptive analysis of every earlier \cable rollout with a recorded initial state found that held-out success, first-clip seating, operator burden, and teleoperation time all agreed on one axis, the bend of the rope path at the first clip.
Success drops from $17.6\%$ in the straightest bend bucket to $2.5\%$ in the sharpest.
The round-5 coverage fill used that term at $\beta = 1.0$, with promotion unchanged.

\subsection{Budget split and coverage guardrails}
\label{app:11}

Each round's budget $M$ goes first to verbatim failure promotion, one episode per observed failure up to the cap $m_f$. The hardness-weighted coverage fill takes the rest.
The real-world tasks use $M = 50$ and $m_f = 25$ on \marker and \nut, and $M = 100$ and $m_f = 50$ on \cable.
Simulation uses $M = 100$ and promotes every diagnostic failure of the round.

At round~5, \nut split its budget into 21 promotions and 29 fills, and \cable into 50 and 50.
On \cable, 21 promotions are initial states from the previous round's held-out evaluation on which both the actor and the reranked actor failed, and 29 are earlier evaluation failures that had not been promoted. We report \cable results on a separate block, drawn after the last collection round and disjoint from every collection and earlier evaluation state.
\marker at round~5 is the exception, with 30 promotions and 20 fills.
Twenty of those promotions are failures of the actor-only arm, and the cap $m_f = 25$ applies to them. The other 10 are states on which the reranked arm regressed, a separate tier for critic supervision.
A floor of 15 coverage fills bounds the total, so promotion can reach at most 35.
Two guardrails keep the hardness weighting from collapsing coverage:
\begin{itemize}
  \item \textbf{Coverage ceiling.} Hardness weighting may shift where coverage is spent but may not leave large regions unsampled. We enforce this by keeping the cumulative fill's $p_{95}$ nearest-neighbor coverage radius at or below that of a fresh pure-Sobol block of the same size.
  \item \textbf{No reuse.} We deduplicate candidate fills against all previously collected initial states and all evaluation states.
        The fitted hardness term learns only from collection episodes and diagnostic rollouts.
        The exception is the evaluation-grid term of \sqnarrow R3 and \sqbroad R2--R3 (\cref{app:28}).
\end{itemize}

\clearpage\suppressfloats[t]

\section{Actor and Critic Implementation}
\label{app:36}
\backlink{\cref{sec:109,subsec:114}}

The code is at \url{https://github.com/ankile/mulligan}.

\subsection{Actor}
\label{app:3}

In both settings, the actor is an action-chunked diffusion policy~\cite{chi2023diffusionpolicy} executing $6$-step chunks (prediction horizon $8$ in simulation, $12$ in the real world), trained with a behavior-cloning loss on the curated human pool $\mathcal{D}_h$ (\cref{sec:109}).
No critic gradients reach the actor in any arm.
\cref{tab:129} lists the actor hyperparameters.

\subsection{Critic}
\label{app:17}

The critic uses discount $\gamma \in [0.99, 0.997]$ depending on task horizon, Polyak-averaged target networks (rate $\rho = 0.005$), and base expectile/quantile level $\tau_\text{base} = 0.7$.
Rewards are the sparse task outcome.
In simulation, we add the intervention penalty of \citet{luo2024rlif}, $-1$ on intervention frames (\cref{tab:132}).
The deployed real-world critics set that penalty to zero.
We take the minimum over an ensemble of two $Q$-networks and clip TD targets.
In simulation, the clip range is the empirical return range, $\text{clip}(\hat{y}, G_\text{min}, G_\text{max})$, and in the real world it is a fixed per-task range, $[0, 1]$ on \marker and \nut and $[0, 2]$ on \cable.
A larger, REDQ-style ensemble~\cite{chen2021redq} gave no benefit (\cref{app:42}).
\cref{tab:132,tab:131} list the critic hyperparameters.

\paragraph{$\lambda$-return target.} The real-world critics regress onto the $\lambda$-return of \cref{sec:109}, the exponentially weighted average of chunked $n$-step targets $y^{(n)}$ with weights $(1-\lambda)\lambda^{n-1}$, truncated at the episode horizon so that the largest $n$ spans the full episode ($798$ steps on \marker, $480$ on \nut, $846$ on \cable, each a multiple of the $6$-step chunk).
$\lambda$ starts at $0.995$, is held for the first $10$k updates ($12.6$k on \cable), and follows a cosine schedule to $0$ by $150$k updates ($188$k on \cable), after which the target is the plain chunked target of \cref{eq:68} for the rest of training.
We anneal $\lambda$ so that early training sees low-bias long-horizon returns while the value estimate is poor, and the deployed checkpoints spend their final updates on the low-variance one-chunk target.

\paragraph{Distributional value head.} The human-in-the-loop simulation results and the real-world critics use DIVL~\citep{wang2026divl}, a categorical value head over $101$ fixed atoms with an HL-Gauss target.
The quantile level is $\tau = \mathrm{clip}(\tau_\text{base} - \alpha\,\mathcal{H}, \tau_\text{min}, \tau_\text{max})$, with $\alpha = 0.4$, $\tau_\text{min} = 0.5$, and $\tau_\text{max} = 0.95$, where $\mathcal{H}$ is the entropy normalized by the logarithm of the number of atoms.
The simulation comparison in \cref{app:24} supports using the same formulation in both domains but does not show an improvement over scalar IQL.

\subsection{Action-magnitude modulation in the critic}
\label{app:26}

The round-5 critics for \marker and \nut and all \cable critics use an observation-conditioned affine transformation of the action input to $Q$ (\cref{tab:131}).
With it, the critic compares action differences at a scale set by the state.
That scale matters when a task alternates between large transport motions and small insertion corrections, and \citet{paresmorlans2026vlacontactrich} report a sharp drop in action magnitude during contact for contact-rich VLAs.

\paragraph{Transformation.}
Let $x(s)$ concatenate the frozen visual features and normalized proprioception, and let $\bar a_{j,d}$ be the critic's globally z-scored action at chunk step $j$, dimension $d$.
A two-layer MLP with a $256$-unit ReLU hidden layer maps $x(s)$ to six offsets $\mu_d(s)$ and six log-scales $\ell_d(s)$.
Before either $Q$ head scores the chunk, we apply
\begin{equation}
  \widetilde a_{j,d} =
  \begin{cases}
    \bigl(\bar a_{j,d}-\mu_d(s)\bigr)
    \exp\!\bigl(-\operatorname{clip}(\ell_d(s),-2,2)\bigr), & d=1,\ldots,6, \\
    \bar a_{j,d},                                           & d=7.
  \end{cases}
  \label{eq:66}
\end{equation}
We transform only the six arm dimensions and pass the gripper through unchanged, with one set of state-dependent parameters shared by every chunk step and candidate action.
We implement it as FiLM-style feature-wise affine modulation~\citep{perez2018film} of the critic's action input.

\paragraph{Training and deployment.}
The transformation starts as the identity because we initialize the MLP's final weights and biases to zero.
The $Q$ loss trains its parameters, and the target $Q$ heads that construct the $V$ labels use a Polyak-averaged copy.
At deployment, the transform acts only when the critic scores proposals.
FiLM can change which proposal the robot executes but not the proposals themselves.

\paragraph{Evidence and scope.}
Offline, the evidence for FiLM was mixed.
On \nut, FiLM candidates improved the point estimate of the candidate-ranking diagnostic in both tested training seeds, but its confidence intervals included no improvement, and FQE ranked the FiLM candidate below the other tested critics.
On \marker, frozen-state probes showed changes in selected action magnitude and progress along a human-annotated direction, with results depending on checkpoint and training seed.
The later \marker robot screen compared critics that all used FiLM, while the \nut screen compared FiLM with changes to batch size or the value quantile.
The reported round-5 critics also combine FiLM with other training changes.
These comparisons do not isolate FiLM's contribution to the reported robot gains.

\subsection{Best-of-\texorpdfstring{$N$}{N} reranking and inference cost}
\label{app:9}

At deployment, the frozen actor draws $N$ action chunks in a single batched forward pass, the critic scores all $N$, and the argmax chunk is executed (\cref{eq:65}).
The $N$ samples share one denoising pass along a batch dimension, and the critic is a small MLP head on already-computed features.
Reranking widens the batch instead of adding $N$ sequential policy evaluations.
\cref{tab:130} lists the deployed $N$ and the selection settings.

We timed this on the robot for \nut R3 at $N = 16$ over $797$ action chunks of $20$ held-out episodes.
From the start of candidate generation to the argmax selection, a chunk takes $39.1$\,ms at the median, $45.1$\,ms at the $99$th percentile, and $48.8$\,ms at worst, with stage medians of $4.0$\,ms for image encoding, $24.0$\,ms for batched diffusion sampling with $8$ DDIM steps, and $4.2$\,ms for critic scoring.
A chunk of $6$ steps at $15$\,Hz is a $400$\,ms budget, so this uses at most $12.2\%$ of it.

\subsection{Simulation}
\label{app:38}

The actor is a state-based diffusion U-Net~\cite{chi2023diffusionpolicy}, paired with an MLP critic with batch size $1024$.
We train for a fixed number of steps ($150$k for \sqnarrow, $250$k for \sqbroad) across five replicate runs per arm, with $\gamma = 0.99$.
The critic uses the DIVL value head (\cref{app:24}), with base quantile $\tau_\text{base} = 0.7$ adjusted by the predicted entropy with sensitivity $\alpha = 0.4$ and clipped to $[0.5, 0.95]$.
We deploy reranked simulation arms with $N = 32$.
The autonomous baselines and REDQ ablations use a scalar IQL critic.
The deployed \sqnarrow Sobol sequence draws only two of the three initial-state axes (\cref{app:61}), and \cref{app:53} gives the exact Sobol configuration.
The simulation columns of \cref{tab:129,tab:132,tab:131,tab:133} give the full configuration.

\subsection{Real world}
\label{app:37}

The actor is an image-based diffusion policy~\cite{chi2023diffusionpolicy} with a ResNet18~\cite{he2016resnet} encoder per camera (ImageNet~\cite{deng2009imagenet} initialization and normalization, $224 \times 224$ inputs, batch size $64$), paired with a Vision-IQL critic that takes its encoder from the deployed actor and keeps it frozen.
CNN encoders such as this one show stronger invariance to task-irrelevant visual variation than ViTs~\citep{yardi2025sim2real}.
The critic fits only the $Q$ and $V$ heads on features precomputed over $10$ augmented views per frame.
We evaluate value-based reranking with $N=16$ on \marker at R2--R4 and \nut at R3--R4, and with $N=32$ on both tasks at R5 and on \cable at every reranked round (R3--R5).
At \marker R2 the reranked policy did not improve on its actor ($10$ rescues, $11$ regressions), so the headline plots the actor alone at that round.
The real-world columns of \cref{tab:129,tab:132,tab:131,tab:133} give the full configuration.

\subsection{Compute and training budgets}
\label{app:16}

\cref{tab:148} lists the logged runtime of the $386$ training runs behind the reported results.
These are the actor with its scalar critic and the DIVL critic for each of the $34$ human-in-the-loop simulation cells and five seeds, and the $36$ actors and $10$ critics of the released real-world curves.
Every run used a single GPU, over $90\%$ of the hours on H100, L40S, RTX 4500 Ada, A40, and RTX A6000 cards.
The runs total $801.5$ GPU-hours, including setup and in-run evaluation.
The total excludes separate evaluation jobs, feature caching, failed runs, autonomous baselines, and development sweeps.
The per-run step budgets are in \cref{tab:129,tab:131}.

\begin{table}[htbp]
  \centering
  \footnotesize
  \setlength{\tabcolsep}{4pt}
  \caption{Logged runtime of the training runs behind the reported results, in hours on one GPU.
    Simulation actors and their scalar critics were trained jointly.}
  \label{tab:148}
  \begin{tabular}{llrrrr}
\hline
Task & Fit & Runs & Median h & Range h & Sum GPU-h \\
\hline
\sqnarrow & Actor + scalar critic & 80 & 1.90 & 1.33--5.17 & 166.9 \\
\sqnarrow & DIVL critic & 80 & 0.65 & 0.53--1.17 & 56.9 \\
\sqbroad & Actor + scalar critic & 90 & 3.01 & 1.74--6.53 & 298.8 \\
\sqbroad & DIVL critic & 90 & 0.97 & 0.83--1.90 & 91.3 \\
\marker & Actor & 12 & 3.16 & 1.82--7.85 & 46.3 \\
\marker & Critic & 4 & 5.29 & 1.61--10.30 & 22.5 \\
\nut & Actor & 12 & 4.34 & 2.73--7.80 & 56.1 \\
\nut & Critic & 3 & 6.55 & 1.22--10.21 & 18.0 \\
\cable & Actor & 12 & 3.12 & 1.72--3.27 & 33.6 \\
\cable & Critic & 3 & 3.69 & 3.17--4.42 & 11.3 \\
\hline
Total & & 386 & & & 801.5 \\
\hline
\end{tabular}

\end{table}
\FloatBarrier

\clearpage\suppressfloats[t]

\section{Hyperparameters}
\label{app:35}
\backlink{\cref{sec:109}}

We read the values below off the configurations that produced the reported results.
When simulation and the real world share a value, it spans both columns.
The reasoning behind these choices is in \cref{app:36}.
\cref{app:50} derives the sampler, and the ablations in \cref{app:63} settled the critic objective.

\newlength{\apphpkeywidth}
\newlength{\apphpvaluewidth}
\newlength{\apphpgap}
\setlength{\apphpkeywidth}{4.5cm}
\setlength{\apphpvaluewidth}{4.0cm}
\setlength{\apphpgap}{0.2cm}
\newcolumntype{K}{>{\raggedright\arraybackslash}p{\apphpkeywidth}}
\newcolumntype{V}{>{\centering\arraybackslash}p{\apphpvaluewidth}}
\newcolumntype{S}{!{\hspace{\apphpgap}\vrule\hspace{\apphpgap}}}
\newcolumntype{B}{>{\centering\arraybackslash}p{\dimexpr 2\apphpvaluewidth + 2\tabcolsep + 2\apphpgap + \arrayrulewidth\relax}}

\paragraph{\cable entries.}
On \cable, a round is one $100$-episode increment of \cref{app:20}, and the table names \cable wherever its value differs from \marker and \nut.
Its three deployed critics (rounds 3--5) all use FiLM DIVL heads, $\gamma = 0.997$, the annealed TD($\lambda$) target through horizon $846$, four independent target views per transition, a warmup-cosine learning rate from $10^{-4}$ to a $10^{-5}$ floor, and $N = 32$ at deployment.

\subsection{Actor}
\label{app:30}

In both domains, the actor is an action-chunked diffusion policy~\cite{chi2023diffusionpolicy} with a conditional U-Net over the action chunk, one conditioning observation step, and DDIM sampling with $8$ denoising steps at inference.
Simulation encodes a state vector and the real world uses per-camera ResNet18 features.
As \cref{tab:129} shows, the prediction horizon and optimizer preset also differ.

\begin{table}[tbp]
  \centering
  \footnotesize
  \setlength{\tabcolsep}{4pt}
  \renewcommand{\arraystretch}{1.15}
  \caption{Diffusion-policy actor.
    Simulation uses five replicate runs per arm, and the real world one run per arm per round.
    We fixed the training step count before seeing any result and deployed the final checkpoint in both domains.}
  \label{tab:129}
  \begin{tabular}{KVSV}
    \hline
    Hyperparameter                  & Simulation                                                                                     & Real world                                                                                                                                    \\
    \hline
    \multicolumn{3}{l}{\emph{Architecture}}                                                                                                                                                                                                                                          \\
    Backbone                        & \multicolumn{2}{B}{1-D conditional U-Net over the action chunk}                                                                                                                                                                                \\
    U-Net channel widths            & \multicolumn{2}{B}{$[512, 1024]$}                                                                                                                                                                                                              \\
    Kernel size / group-norm groups & \multicolumn{2}{B}{$5$ / $8$}                                                                                                                                                                                                                  \\
    Diffusion-step conditioning     & \multicolumn{2}{B}{$128$-D embedding, FiLM scale modulation}                                                                                                                                                                                   \\
    Observation encoder             & none (state vector)                                                                            & ResNet18, one per camera                                                                                                                      \\
    Encoder initialization          & ---                                                                                            & ImageNet, fully fine-tuned                                                                                                                    \\
    Encoder normalization layers    & ---                                                                                            & batch norm, retained                                                                                                                          \\
    Feature pooling                 & ---                                                                                            & spatial softmax, $32$ keypoints                                                                                                               \\
    \hline
    \multicolumn{3}{l}{\emph{Horizons}}                                                                                                                                                                                                                                              \\
    Observation steps               & \multicolumn{2}{B}{$1$}                                                                                                                                                                                                                        \\
    Prediction horizon              & $8$                                                                                            & $12$                                                                                                                                          \\
    Executed steps per query $h$    & \multicolumn{2}{B}{$6$}                                                                                                                                                                                                                        \\
    \hline
    \multicolumn{3}{l}{\emph{Diffusion}}                                                                                                                                                                                                                                             \\
    Training schedule               & \multicolumn{2}{B}{squared-cosine, $100$ timesteps, $\beta \in [10^{-4},\, 2 \times 10^{-2}]$}                                                                                                                                                 \\
    Parameterization                & \multicolumn{2}{B}{$\epsilon$-prediction, samples clipped to $[-1, 1]$}                                                                                                                                                                        \\
    Inference sampler               & \multicolumn{2}{B}{DDIM, $8$ steps}                                                                                                                                                                                                            \\
    \hline
    \multicolumn{3}{l}{\emph{Observations and actions}}                                                                                                                                                                                                                              \\
    Proprioception                  & eef.\ position, quaternion, gripper ($9$-D)                                                    & eef.\ pose, gripper ($7$-D)                                                                                                                   \\
    Other observations              & simulator object state                                                                         & ---                                                                                                                                           \\
    Cameras consumed                & ---                                                                                            & $2$ (\marker, \nut), $3$ (\cable)                                                                                                             \\
    Image resolution                & ---                                                                                            & $224 \times 224$ after a fixed per-camera crop                                                                                                \\
    Image augmentation              & ---                                                                                            & brightness, contrast, saturation, hue, sharpness, small affine; at most $3$ per sample                                                        \\
    Image normalization             & ---                                                                                            & ImageNet mean and standard deviation                                                                                                          \\
    Action space                    & operational-space delta pose, gripper ($7$-D)                                                  & Cartesian velocity, gripper ($7$-D) on \marker and \nut; relative pose anchored at the measured end-effector pose, gripper ($10$-D) on \cable \\
    State/action normalization      & \multicolumn{2}{B}{min--max}                                                                                                                                                                                                                   \\
    \hline
    \multicolumn{3}{l}{\emph{Optimization}}                                                                                                                                                                                                                                          \\
    Optimizer                       & AdamW                                                                                          & Adam                                                                                                                                          \\
    Learning rate                   & $3 \times 10^{-4}$                                                                             & $10^{-4}$                                                                                                                                     \\
    Betas                           & $(0.9, 0.999)$                                                                                 & $(0.95, 0.999)$                                                                                                                               \\
    Weight decay                    & $10^{-4}$                                                                                      & $10^{-6}$                                                                                                                                     \\
    Schedule                        & constant                                                                                       & cosine, $500$ warmup steps                                                                                                                    \\
    Gradient clipping (global norm) & $1.0$                                                                                          & $10.0$                                                                                                                                        \\
    Weight EMA                      & \multicolumn{2}{B}{not used}                                                                                                                                                                                                                   \\
    Batch size                      & $1024$                                                                                         & $64$                                                                                                                                          \\
    Training steps                  & $150$k (\sqnarrow), $250$k (\sqbroad)                                                          & $100$k                                                                                                                                        \\
    Replicate runs per arm          & $5$                                                                                            & $1$                                                                                                                                           \\
    Precision                       & bfloat16 autocast, TF32 matmuls                                                                & fp32                                                                                                                                          \\
    Actor training data             & \multicolumn{2}{B}{$\mathcal{D}_h$: human frames of successful episodes}                                                                                                                                                                       \\
    \hline
  \end{tabular}
\end{table}

\subsection{Critic}
\label{app:32}

We use the categorical value objective for the real-world critics and the human-in-the-loop simulation results (\cref{tab:132,tab:131}), and scalar IQL for the simulation autonomous baselines and REDQ ablations.
In simulation, the $Q$ and $V$ heads train while the actor and normalizer stay frozen.
On the robot, we freeze the deployed actor's vision encoder and fit the heads on cached features.

\begin{table}[tbp]
  \centering
  \footnotesize
  \setlength{\tabcolsep}{4pt}
  \renewcommand{\arraystretch}{1.15}
  \caption{Critic objective and reward.
    Both domains fit a categorical value distribution, with a quantile level that adapts to the predicted entropy.
    Simulation entries are for the reported human-in-the-loop cells and real-world entries for the deployed round-5 critics, with the task named wherever the three tasks differ.}
  \label{tab:132}
  \begin{tabular}{KVSV}
    \hline
    Hyperparameter             & Simulation                                                                            & Real world                                                                                                                                                                                                                                                                        \\
    \hline
    \multicolumn{3}{l}{\emph{Objective}}                                                                                                                                                                                                                                                                                                                                                                   \\
    Value head                 & categorical (DIVL)                                                                    & categorical (DIVL)                                                                                                                                                                                                                                                                \\
    Expectile / base quantile  & \multicolumn{2}{B}{$\tau_\text{base} = 0.7$}                                                                                                                                                                                                                                                                                                                              \\
    Discount $\gamma$          & $0.99$                                                                                & $0.99$ (\nut), $0.997$ (\marker, \cable)                                                                                                                                                                                                                                          \\
    Target-network rate $\rho$ & \multicolumn{2}{B}{$0.005$ (Polyak)}                                                                                                                                                                                                                                                                                                                                      \\
    $Q$-ensemble               & \multicolumn{2}{B}{$2$ networks, min aggregation for both the TD label and reranking}                                                                                                                                                                                                                                                                                     \\
    TD target                  & \multicolumn{2}{B}{$h$-step chunked with a $V$-bootstrap, $h = 6$}                                                                                                                                                                                                                                                                                                        \\
    TD($\lambda$) curriculum   & none                                                                                  & $\lambda$-return over chunked $n$-step targets up to the episode horizon ($798$ steps on \marker, $480$ on \nut, $846$ on \cable), $\lambda$ held at $0.995$ then cosine-annealed to $0$ (hold $10$k, zero at $150$k on \marker and \nut; hold $12.6$k, zero at $188$k on \cable) \\
    Target clipping            & empirical per-episode return range                                                    & $[0, 1]$ (\marker, \nut), $[0, 2]$ (\cable)                                                                                                                                                                                                                                       \\
    \hline
    \multicolumn{3}{l}{\emph{Reward}}                                                                                                                                                                                                                                                                                                                                                                      \\
    Task reward                & sparse: $1$ on success frames                                                         & sparse: $1$ held over the terminal tail of a successful episode ($2$ for \cable, one per seated clip)                                                                                                                                                                             \\
    Intervention penalty       & $-1$ at intervention frames                                                           & none                                                                                                                                                                                                                                                                              \\
    Reward shift               & \multicolumn{2}{B}{$0$}                                                                                                                                                                                                                                                                                                                                                   \\
    Terminal handling          & terminal chunks kept in the loss; padded actions drawn uniformly                      & terminal tails supervised without bootstrapping; timed-out episodes get no terminal anchor                                                                                                                                                                                        \\
    \hline
  \end{tabular}
\end{table}

\begin{table}[tbp]
  \centering
  \footnotesize
  \setlength{\tabcolsep}{4pt}
  \renewcommand{\arraystretch}{1.15}
  \caption{Critic network, optimization, and distributional value head.
    Real-world values are those of the deployed round-5 critics.
    Where the three tasks differ, the task is named.}
  \label{tab:131}
  \begin{tabular}{KVSV}
    \hline
    Hyperparameter               & Simulation                                                                                      & Real world                                                                                                                                              \\
    \hline
    \multicolumn{3}{l}{\emph{Network and optimization}}                                                                                                                                                                                                                                      \\
    $Q$ and $V$ trunk            & \multicolumn{2}{B}{MLP $[1024, 1024, 1024]$ with layer norm}                                                                                                                                                                                              \\
    $Q$ input                    & \multicolumn{2}{B}{observation features concatenated with the flattened $6$-step chunk}                                                                                                                                                                   \\
    Visual features              & ---                                                                                             & the deployed actor's encoder, frozen; features precomputed over $10$ augmented views per frame                                                          \\
    Proprioception dropout       & $0$                                                                                             & $0.5$                                                                                                                                                   \\
    Extra critic head            & none                                                                                            & state-conditioned FiLM rescaling of the critic's action input (\cref{app:26}), used in the round-5 \marker and \nut critics and in all \cable critics \\
    Learning rate ($Q$ and $V$)  & $3 \times 10^{-4}$                                                                              & $10^{-4}$                                                                                                                                               \\
    Learning-rate schedule       & constant                                                                                        & cosine with warmup to a $0.1$ floor; $10$k warmup steps ($12.6$k on \cable)                                                                             \\
    Weight decay                 & \multicolumn{2}{B}{$10^{-4}$}                                                                                                                                                                                                                             \\
    Gradient clipping            & \multicolumn{2}{B}{$1.0$}                                                                                                                                                                                                                                 \\
    Batch size                   & $1024$                                                                                          & $256$                                                                                                                                                   \\
    Training steps               & $150$k (\sqnarrow), $250$k (\sqbroad)                                                           & $250$k (\marker, \nut), $314$k (\cable; $270$k at R3--R4)                                                                                               \\
    Deployed checkpoint          & final step                                                                                      & $200$k (\marker, \nut), final (\cable)                                                                                                                  \\
    Batch composition            & $50\%$ human success, $50\%$ DAgger                                                             & uniform over all cached frames                                                                                                                          \\
    Critic training data         & \multicolumn{2}{B}{$\mathcal{D}_\text{all}$: every transition, regardless of source or outcome}                                                                                                                                                           \\
    Relation to the actor        & actor and normalizer frozen                                                                     & frozen actor encoder                                                                                                                                    \\
    \hline
    \multicolumn{3}{l}{\emph{Distributional value head}}                                                                                                                                                                                                                                     \\
    Atoms                        & $101$                                                                                           & $101$                                                                                                                                                   \\
    Support                      & empirical return range widened by $5\%$                                                         & $[-0.05, 1.05]$ (\marker, \nut); empirical return range widened by $5\%$ (\cable)                                                                       \\
    HL-Gauss $\sigma$            & $0.75 \times$ atom spacing                                                                      & $0.75 \times$ atom spacing                                                                                                                              \\
    Entropy sensitivity $\alpha$ & $0.4$                                                                                           & $0.4$                                                                                                                                                   \\
    Quantile clip                & $[0.5, 0.95]$                                                                                   & $[\tau_\text{min}, \tau_\text{max}] = [0.5, 0.95]$                                                                                                      \\
    \hline
  \end{tabular}
\end{table}

\subsection{Best-of-\texorpdfstring{$N$}{N} reranking}
\label{app:31}

At each policy query, the frozen actor draws $N$ action chunks in one batched denoising pass and the critic scores each with $Q(s, a_{t:t+h})$ over the executed $h = 6$ steps.
The robot executes the highest-scoring chunk.
We take a plain argmax, with no temperature, top-$k$ softening, or stage gate.
\cref{app:9} reports the inference cost.

\begin{table}[tbp]
  \centering
  \footnotesize
  \setlength{\tabcolsep}{4pt}
  \renewcommand{\arraystretch}{1.15}
  \caption{Reranking at deployment.
    Real-world rows list the reranked rounds, and actor-only evaluations use $N=1$.
    \cref{app:44} discusses the choice of $N$.}
  \label{tab:130}
  \begin{tabular}{>{\raggedright\arraybackslash}p{4.8cm}>{\raggedright\arraybackslash}p{8.0cm}}
    \hline
    Setting                     & Value                                                           \\
    \hline
    Candidates $N$, simulation  & $32$                                                            \\
    Candidates $N$, \marker     & $16$ at R2--R4, $32$ at R5                                      \\
    Candidates $N$, \nut        & $16$ at R3--R4, $32$ at R5                                      \\
    Candidates $N$, \cable      & $32$ at R3--R5                                                  \\
    Scored quantity             & $Q(s, a_{t:t+6})$, min over the $2$-network ensemble            \\
    Selection rule              & argmax                                                          \\
    Candidate draw              & one batched denoising pass over $N$ copies of the observation   \\
    Rescoring cadence           & once per executed chunk, that is every $6$ steps                \\
    Reranking during collection & simulation: the DAgger collector is itself deployed at $N = 32$ \\
    \hline
  \end{tabular}
\end{table}

\subsection{Initial-state sampler}
\label{app:34}

\cref{tab:134} instantiates the procedure of \cref{app:49}.
Both domains maximize $d(c, \cdot)\,(1 + \beta\,h(c))$ with the same greedy farthest-point loop, although \sqnarrow ranks by $d^2$.
The domains draw fill candidates from different pools and differ in how they gate failures before promotion and how much weight the hardness term carries.
The deployed $\beta$ also changes with task and round.
At $\beta = 0$, as on \marker at round~5 and on \cable through round~4, the fill reduces to pure coverage.

\begin{table}[tbp]
  \centering
  \footnotesize
  \setlength{\tabcolsep}{4pt}
  \setlength{\apphpkeywidth}{3.0cm}
  \setlength{\apphpvaluewidth}{4.8cm}
  \caption{Initial-state sampler.
    All budgets are per arm per round.
    In simulation, ``uncapped'' promotion promotes every failure among the round's $100$ diagnostic rollouts (the first $100$ of $200$ at \sqbroad R1), which came to $1$--$21$ promotions on \sqnarrow and $14$--$51$ on \sqbroad.}
  \label{tab:134}
  \begin{tabular}{KVSV}
    \hline
    Setting              & Simulation                                                                                                                                                                                                         & Real world                                                                                                                        \\
    \hline
    \multicolumn{3}{l}{\emph{Sobol draw}}                                                                                                                                                                                                                                                                                                                                         \\
    Sampled dimensions   & $2$ (\sqnarrow: nut $y$, yaw), $5$ (\sqbroad)                                                                                                                                                                      & $5$ (\marker, \nut), $7$ (\cable)                                                                                                 \\
    Remaining dimensions & \sqnarrow nut $x$: drawn independently uniform                                                                                                                                                                     & ---                                                                                                                               \\
    Scrambling           & \multicolumn{2}{B}{linear matrix scrambling with a digital shift (SciPy), always on}                                                                                                                                                                                                                                                                   \\
    Pool size            & $1024$ (\sqnarrow), $2048$ before rejection (\sqbroad)                                                                                                                                                             & $32768$ (\marker, \nut), $4096$ (\cable)                                                                                          \\
    Rejection            & \sqbroad: nut--peg center distance above $13.3$\,cm                                                                                                                                                                & \cable: clip separation at least $5.1$\,cm                                                                                        \\
    Placement snapping   & none                                                                                                                                                                                                               & $1$-inch grid on fixture placements                                                                                               \\
    Index bookkeeping    & \multicolumn{2}{B}{contiguous disjoint blocks per round; a consumed block is retired}                                                                                                                                                                                                                                                                  \\
    \hline
    \multicolumn{3}{l}{\emph{Budget}}                                                                                                                                                                                                                                                                                                                                             \\
    R0 demonstrations    & $100$ ($200$ on \sqbroad)                                                                                                                                                                                          & $100$                                                                                                                             \\
    Budget $M$ (R1 on)   & $100$                                                                                                                                                                                                              & $50$ (\marker, \nut), $100$ (\cable)                                                                                              \\
    Promotion cap $m_f$  & uncapped                                                                                                                                                                                                           & $25$ on the failure pool ($50$ on \cable); a floor of $15$ fills bounds the total                                                 \\
    Realized promotions  & $1$--$21$ (\sqnarrow), $14$--$51$ (\sqbroad)                                                                                                                                                                       & $21$ (\nut R5), $50$ (\cable R5), $30$ (\marker R5)                                                                               \\
    Fill count           & $M$ minus promotions and perturbations                                                                                                                                                                             & $29$ (\nut R5), $50$ (\cable R5), $20$ (\marker R5)                                                                               \\
    \hline
    \multicolumn{3}{l}{\emph{Failure promotion}}                                                                                                                                                                                                                                                                                                                                  \\
    Failure source       & the round's diagnostic rollouts                                                                                                                                                                                    & earlier rounds' held-out evaluation blocks, never the reported \cable block \\
    Eligibility          & every failed initial state                                                                                                                                                                                                 & failures of the \ours{} arm, of both actor and reranked actor once both ran \\
    Ordering             & none                                                                                                                                                                                                               & earliest failure stage, then episode length                                                                                       \\
    Replay               & \multicolumn{2}{B}{verbatim: the recorded initial state is reproduced exactly}                                                                                                                                                                                                                                                                         \\
    Deduplication        & none                                                                                                                                                                                                               & exact match against every previously collected initial state                                                                              \\
    Local perturbations  & \sqnarrow: one state around each of the $20$ longest diagnostic episodes, in $\mathcal{G}$ (nut $x$ redrawn uniformly, Gaussian noise of std.\ $0.025$\,m on nut $y$ and $0.05$\,rad on yaw)                                                                                                                         & none                                                                                                                              \\
    \hline
    \multicolumn{3}{l}{\emph{Hardness-weighted farthest-point fill}}                                                                                                                                                                                                                                                                                                              \\
    Objective            & \multicolumn{2}{B}{$\arg\max_c\; d(c,\, \mathcal{P} \cup \mathcal{G} \cup \mathcal{X})\,(1 + \beta\,h(c))$, greedy and deterministic; \sqnarrow uses $d^2$, and at \sqbroad R3 $\mathcal{P}$ omits the R2 initial states}                                                                                                                                      \\
    Distance             & \multicolumn{2}{B}{normalized to the unit cube, yaw treated as periodic (\sqbroad: not in the initial distance to $\mathcal{P} \cup \mathcal{G}$)}                                                                                                                                                                                                     \\
    Axis weights         & none                                                                                                                                                                                                               & $0.70$ on the holder (\marker), $0.30$ on the peg (\nut), none (\cable)                                                           \\
    Fill candidate pool  & fresh uniform draws: $50$k (\sqnarrow); $9$k at R1, $7$k at R2--R3 (\sqbroad)                                                                                                                                      & the scrambled Sobol pool above                                                                                                    \\
    Hardness signal $h$  & an episode-length fit over the initial-state parameters, combined at R3 with an evaluation-grid cell term that replaces it on \sqbroad R2 (\cref{app:28})                                                        & per-task geometric term, screened each round                                                                                      \\
    Deployed $\beta$     & \sqnarrow: $0.5$ on the fitted term, plus $1.0$ on the cell term at R3; \sqbroad: $1.0$ (fitted term at R1, cell term at R2, their mean at R3)                                                                     & $0.5$ (\nut R5), $0$ (\marker R5), $0$ (\cable through R4) and $1.0$ on a rope-bend prior (\cable R5); \marker R3--R4 used $0.05$ \\
    Coverage guardrail   & \multicolumn{2}{B}{the cumulative $p_{95}$ nearest-neighbor radius is held at or below that of a fresh Sobol block of the same size (\sqbroad R2--R3: within $5\%$ of the worse of that block and a uniform draw)}                                                                                                                                     \\
    \hline
  \end{tabular}
\end{table}

\subsection{Environment, control, and the improvement loop}
\label{app:33}

The real-world episode caps in \cref{tab:133} follow the rule of \cref{app:57}.
\cref{tab:145} lists the per-task initial-state ranges, and \cref{app:25} gives the evaluation protocol.

\begin{table}[tbp]
  \centering
  \footnotesize
  \setlength{\tabcolsep}{4pt}
  \renewcommand{\arraystretch}{1.15}
  \caption{Environment, control loop, and improvement loop.}
  \label{tab:133}
  \begin{tabular}{KVSV}
    \hline
    Setting                    & Simulation                                                                        & Real world                                                     \\
    \hline
    Platform                   & Robosuite with MimicGen assets                                                    & Franka Panda on DROID                                          \\
    Control rate               & $20$\,Hz                                                                          & $15$\,Hz                                                       \\
    Controller                 & operational-space pose, delta, base frame                                         & Cartesian velocity                                             \\
    Per-step action limits     & $5$\,cm, $0.5$\,rad                                                               & ---                                                            \\
    Cameras                    & none; state observations                                                          & ZED stereo, HD720 captured at $60$\,fps                        \\
    Teleoperation              & \multicolumn{2}{B}{SpaceMouse, with operator-triggered interventions}                                                                              \\
    Evaluation episode cap     & $400$ steps                                                                       & $600$ (\nut), $800$ (\marker), $850$ (\cable)                  \\
    Collection episode cap     & \multicolumn{2}{B}{none; the operator ends the episode}                                                                                            \\
    Rounds run                 & R0--R3                                                                            & R0--R5                                                         \\
    Episodes per round per arm & $100$ ($200$ at R0 on \sqbroad)                                                   & $50$ ($100$ on \cable), $100$ at R0                            \\
    Held-out evaluation set    & $8000$ (\sqnarrow) or $30000$ (\sqbroad) fixed Sobol initial states, reused across rounds & $50$ fresh Sobol initial states per round (one shared block on \cable) \\
    Training-time evaluation   & $400$ fixed Sobol initial states, at the final step only                                  & held-out action-reconstruction loss every $5$k steps           \\
    Replication unit           & $5$ training runs                                                                 & evaluation episodes                                            \\
    \hline
  \end{tabular}
\end{table}

\clearpage\suppressfloats[t]

\section{Baseline Details}
\label{app:5}

Each comparison below changes one design choice. The actor architecture, the critic architecture where there is a critic, and the per-round collection budget stay fixed across all arms.

\subsection{HG-DAgger}
\label{app:7}

Our HG-DAgger baseline~\cite{kelly2019hgdagger} shares the budget, round structure, blinding, evaluation protocol, and actor training recipe of \ours{}.
The two differ in where episodes start and what the operator collects there.
Its initial states are i.i.d.\ uniform draws over $\mathcal{S}_0$, and the operator records corrections but no counterfactual demonstrations.
Because we collect both arms in the same blinded, interleaved sessions, drift in operator skill or in the environment affects both equally.

\subsection{Autonomous self-improvement arms}
\label{app:6}

The autonomous arms start from the supervised arms' round-0 checkpoint and then improve with no human in the loop, attempting as many episodes per round as the supervised arms collect.
They differ in how they use their own rollouts.
\emph{Autonomous IL}~\cite{pomerleau1988alvinn,chi2023diffusionpolicy} retrains on all of them, \emph{filtered BC}~\cite{mirchandani2024scaling} retrains on the successful ones only, and \emph{batch-online IDQL}~\cite{hansenestruch2023idql,dong2025batchonline} trains a critic on all of them and deploys with best-of-$N$ reranking.
Its \emph{$+$ success BC} variant also retrains the actor on its successful rollouts.
All of them share the round-0 fixed-grid evaluation. From round 1 on, IDQL deploys with $N = 32$ and the two imitation arms with $N = 1$.

\subsection{RECAP learner comparison}
\label{app:41}

Our conclusions could depend on the HiL-IDQL learner (a BC actor, an IQL critic, and best-of-$N$ reranking).
To test this, we train RECAP on the same collected datasets and score it on the same fixed-grid evaluation sets.
Its prediction horizon ($8$), execution horizon ($6$), and advantage horizon ($6$) match the actor configuration used throughout.

As \cref{tab:136} shows, RECAP is below the matched HiL-IDQL control in every cohort, and the gap is widest on \sqbroad.
Both learners act from a single actor sample ($N{=}1$), and we take each difference seed by seed over the same five seeds.

\begin{table}[tbp]
  \centering
  \footnotesize
  \setlength{\tabcolsep}{4pt}
  \caption{RECAP vs.\ the matched HiL-IDQL learner on identical data.
    Mean $\pm$ standard error over five seeds.
    Differences are computed per seed, with two-sided Student-t $95\%$ confidence intervals on four degrees of freedom.
    Each seed is scored at its final checkpoint on the fixed held-out grid ($8{,}000$ episodes on \sqnarrow, $30{,}000$ on \sqbroad).}
  \label{tab:136}
  \begin{tabular}{lccr@{\hskip 6pt}c}
  \hline
  Cohort & DP$+$IQL & RECAP & \multicolumn{1}{c}{$\Delta$ (pp)} & Paired $95\%$ CI \\
  \hline
  \sqnarrow R2 & $95.39 \pm 0.30$ & $87.02 \pm 0.45$ & $-8.37$ & $[-9.43, -7.31]$ \\
  \sqnarrow R3 & $95.68 \pm 0.31$ & $90.03 \pm 0.42$ & $-5.65$ & $[-6.64, -4.66]$ \\
  \sqbroad R3 & $86.03 \pm 0.81$ & $65.36 \pm 0.67$ & $-20.67$ & $[-23.75, -17.59]$ \\
  \hline
\end{tabular}

\end{table}

\subsection{HiL-SERL in simulation}
\label{app:8}

\begin{figure}[tbp]
  \centering
  \includegraphics[width=\linewidth]{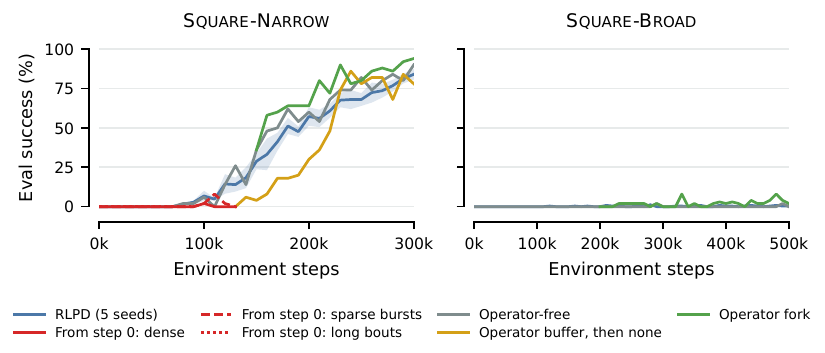}
  \caption{HiL-SERL eval success per session (one seed each) against RLPD (five-seed mean, shaded $\pm$ standard error) on \sqnarrow{} (left) and \sqbroad{} (right).
    Operator forks branch from the operator-free run at $150$k and $200$k.
    The operator-buffer arm continues the sparse-burst session from $130$k with no operator.}
  \label{fig:70}
\end{figure}

We ported HiL-SERL~\cite{luo2024hilserl} to \sqnarrow{} and \sqbroad{} as a second human-in-the-loop baseline.
For the learner we use the SAC form of RLPD~\cite{ball2023rlpd} from the EXPO recipe ($10$ critics, UTD $20$, 50/50 demo/online batches), starting from the round-0 demonstrations that every other arm uses ($100$ on \sqnarrow{}, $200$ on \sqbroad{}).
When the operator takes over with a SpaceMouse, the intervened frames of successful episodes go into the demonstration half of the buffer.
With no operator, the split learner/actor system matches the reference RLPD implementation ($0.90$ vs.\ $0.84 \pm 0.03$ eval success at $300$k environment steps, one seed vs.\ five).
The reference RLPD runs used the EXPO stack on Robosuite 1.4.1 and MuJoCo 3.4.0, while every other simulation result in this paper, including HiL-SERL, uses Robosuite 1.5.2 and MuJoCo 3.3.7.
Eval success is the success rate of $50$ policy-only episodes from fixed initial states, scored every $10$k environment steps, and $\pm$ is the standard error over seeds.
We run one seed per HiL-SERL arm.
\Cref{fig:70} shows the eval curve of every session and \cref{tab:128} its operator cost.
The HiL-SERL curves in \cref{fig:79} are the two operator forks: the operator-free run up to the fork, then the forked session with the operator.

\begin{table}[tbp]
  \centering
  \footnotesize
  \setlength{\tabcolsep}{3pt}
  \caption{HiL-SERL operator cost per session, one seed each.
    A fork counts only its episodes after the fork.
    Steps: the environment-step range run under the protocol.
    Operator hours: summed wall time of the episodes.
    Human frames: share of environment steps under operator control.
    Training success: share of training episodes that succeeded, with operator help where given.
    Policy-only: successes among training episodes without a takeover.
    Final and peak eval: at the last checkpoint and the maximum after the fork.
    Total rows sum the sessions with an operator.
    Their human frames and takeover episodes are pooled shares.}
  \label{tab:128}
  \begin{tabular}{lrrrrrrrrr}
  \hline
  \begin{tabular}[b]{@{}c@{}}Operator \\ protocol\end{tabular} & \begin{tabular}[b]{@{}c@{}}Env \\ steps\end{tabular} & \begin{tabular}[b]{@{}c@{}}Epi- \\ sodes\end{tabular} & \begin{tabular}[b]{@{}c@{}}Operator \\ hours\end{tabular} & \begin{tabular}[b]{@{}c@{}}Human \\ frames\end{tabular} & \begin{tabular}[b]{@{}c@{}}Takeover \\ episodes\end{tabular} & \begin{tabular}[b]{@{}c@{}}Training \\ success\end{tabular} & \begin{tabular}[b]{@{}c@{}}Policy-only \\ successes\end{tabular} & \begin{tabular}[b]{@{}c@{}}Final \\ eval\end{tabular} & \begin{tabular}[b]{@{}c@{}}Peak \\ eval\end{tabular} \\
  \hline
  \multicolumn{10}{l}{\sqnarrow} \\
  Step 0, dense & $0$--$130$k & $558$ & $6.0$ & $47.5\%$ & $80\%$ & $0.37$ & $0/112$ & $0.00$ & $0.02$ \\
  Step 0, sparse bursts & $0$--$130$k & $781$ & $3.3$ & $22.9\%$ & $54\%$ & $0.13$ & $0/363$ & $0.00$ & $0.08$ \\
  Step 0, long bouts & $0$--$90$k & $471$ & $2.7$ & $51.3\%$ & $66\%$ & $0.24$ & $0/162$ & $0.00$ & $0.00$ \\
  Buffer, then none & $130$--$300$k & $674$ & -- & $0.0\%$ & $0\%$ & $0.55$ & $374/674$ & $0.78$ & $0.86$ \\
  After takeoff & $150$--$300$k & $1{,}310$ & $3.1$ & $15.6\%$ & $35\%$ & $0.93$ & $819/856$ & $0.94$ & $0.94$ \\
  \textit{Total} & $503$k & $3{,}120$ & $15.1$ & $32.3\%$ & $52\%$ & & & & \\
  \hline
  \multicolumn{10}{l}{\sqbroad} \\
  Fork at 200k & $200$--$500$k & $1{,}660$ & $7.9$ & $44.9\%$ & $74\%$ & $0.56$ & $1/436$ & $0.02$ & $0.08$ \\
  \hline
  \textbf{Total, both tasks} & $803$k & $4{,}780$ & $23.0$ & $37.0\%$ & $60\%$ & & & & \\
  \hline
\end{tabular}

\end{table}

\paragraph{\sqnarrow{} sessions.}
\begin{itemize}
  \item \textbf{Operator from step 0}, three protocols: dense takeovers; $25$ autonomous failed episodes ($9{,}756$ environment steps), then sparse takeover bursts with operator fail-termination; long takeover bouts with fail-termination.
        None exceeded $0.08$ eval success by $91$--$131$k environment steps, and $0$ of $637$ policy-only training episodes succeeded, over $12.0$ operator hours.
        The operator-free run was at $0.26$ by $130$k.
  \item \textbf{Operator buffer, then no operator.} When the sparse-burst session continued from $130$k without takeovers, the policy took off $50$k steps later than the operator-free run (eval first reached $0.3$ at $200$k, not $150$k) and reached $0.78$ at $300$k ($0.90$ without an operator).
  \item \textbf{Operator after takeoff.} When we forked the operator-free run at $150$k (eval $0.36$) and added the operator, the policy reached $0.94$ at $300$k ($0.90$ without) and first reached $0.8$ at $210$k ($250$k without), for $3.1$ operator hours and $15.6\%$ human frames.
        The gap at $300$k is about one standard error.
  \item \textbf{RLPD ablations} (five seeds each) to test whether the $\sim$$80$k-step zero-success phase is set by failure data or by critic updates: early termination of doomed training episodes, and UTD $40$ (actor updates unchanged).
          Neither changed learning speed (whole-curve mean eval within $\pm0.12$ of the reference, $95\%$ CI).
\end{itemize}
On \sqnarrow{}, human takeovers from step 0 did not substitute for the RL phase and slowed it.
Once the policy succeeded on its own, they sped up the rest of training, so we run HiL-SERL as a fork of an operator-free run after takeoff.
This protocol can show that an operator speeds up training after first success, but not that it shortens the time to first success.

\paragraph{\sqbroad{}.} The operator-free run, like RLPD, does not take off.
Its best eval success over $10^6$ environment steps is $1/50$, so there is no takeoff point to fork from.
We forked the operator-free run at $200$k instead, where its policy succeeded in $0$ of $50$ eval episodes and lifted the nut in $13$, and continued with the operator to $500$k.
The operator helped but did not solve the task.
Over the $30$ checkpoints from $210$k to $500$k, the forked policy's eval was above the operator-free run's at $19$ and below it at none, averaged $0.020$ against $0.001$ ($0.004$ for the five-seed RLPD mean), and peaked at $0.08$ against $0.02$.
Eval ended at $0.02$ at $500$k, and $1$ of $436$ policy-only training episodes succeeded, after $7.9$ operator hours with $44.9\%$ of frames under operator control.
From $440$k, the operator took over in $85$--$96\%$ of episodes and assisted training success rose to $0.95$, but the policy's own eval did not follow.

\paragraph{Takeover gripper defect.} Until we fixed it partway through the HiL-SERL runs, the SpaceMouse gripper toggle carried over between takeover bouts and was copied into the operator's first action.
In the sparse-burst session, $128$ of $228$ bout starts ($56\%$) flipped the gripper.
The defect affected the dense-takeover session and the first $\sim$$62$k environment steps of the sparse-burst session, where flipped bouts made up $0.7\%$ of the demonstration half at the fix.
  The long-bout session, both forks, and the operator-buffer arm ran on the fixed code.
  The defect does not explain the operator-from-step-0 result, since the long-bout session ran entirely on the fixed code and also ended at $0.00$ eval.

  \paragraph{Summary.} HiL-SERL is less sample-efficient than HiL-IDQL+\ours{}.
  On \sqnarrow{} it needs over ten times the environment steps to reach $0.9$ (\cref{sec:105}), and on \sqbroad{} it stays below $0.1$ over $500$k steps, where HiL-IDQL+\ours{} reaches $0.86$ with $68$k collected frames (\cref{fig:79}).

\clearpage\suppressfloats[t]

\section{Value-Learning Diagnostics and Ablations}
\label{app:63}
\backlink{\cref{subsec:116}}

\subsection{Real-world value learning}
\label{app:64}

We train the real-world critic separately from the actor.
It uses the actor's frozen visual encoder together with proprioception to predict state values and action-chunk values.
Two $Q$ heads score the actor's proposals, and a distributional $V$ head supplies the value targets (\cref{app:17,app:37}).
Its training pool grows across rounds to include demonstrations, corrections, counterfactual demonstrations, and autonomous rollouts, including failures.
Completed evaluation rollouts from the \ours{} campaign can enter later critics' training pools.
A block the critic trained on measures only how well it fit that block, so offline validation uses the blocks excluded from its training.

\paragraph{Offline critic selection.}
We asked whether diagnostics on recorded trajectories can pick the critic and training checkpoint under which best-of-$N$ reranking performs best on the robot.
We checked outcome discrimination, return calibration, value propagation, action sensitivity, and fitted-Q evaluation.
A critic can recognize successful trajectories and still fail to choose better actions.
For a fixed state, $V(s)$ is constant across proposals, so ranking by $Q(s,a)-V(s)$ is identical to ranking by $Q(s,a)$.
Accurate state values alone do not show that the critic can tell the actor's candidate actions apart.

\begin{figure}[t]
  \centering
  \includegraphics[width=\linewidth]{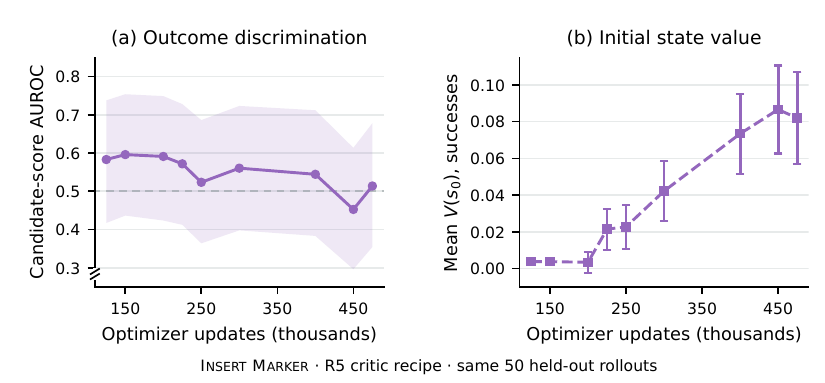}
  \caption{Offline diagnostics across training budgets for the round-5 \marker critic recipe, using the same $50$ excluded rollouts ($23$ successes, $27$ failures).
    (a) How well the maximum $Q$ over $32$ sampled actions predicts the recorded episode outcome, as AUROC on an axis zoomed to $[0.25,0.85]$.
    Shading gives pointwise $95\%$ stratified episode-bootstrap intervals ($5{,}000$ resamples), and the dashed line marks chance.
    (b) Mean initial value on the $23$ successful episodes $\pm$ one standard error across episodes.
    We took the $125$k and $150$k checkpoints from the original run, and the $200$k--$475$k checkpoints from a longer run with the same recipe and seed.}
  \label{fig:82}
\end{figure}

\paragraph{Outcome discrimination and calibration.}
We measured AUROC against recorded episode outcomes using $V(s_0)$, frame-level values, and the mean value over the last quarter of each episode.
We also looked at value traces, at correlation with realized return and task progress, at return-prediction error, and at how far apart successful and failed episodes scored.
When we compared training and excluded blocks, we found degradation that training fit hid.
For example, late-value AUROC for an \marker round-4 critic fell from $0.943$ to $0.897$ and from $0.910$ to $0.862$ on two excluded blocks between $125$k and $150$k updates, while its training-view AUROC rose to $1.000$.
We kept the earlier checkpoint even though the block-level confidence intervals overlapped.
Different metrics also peaked at different checkpoints.
As \cref{fig:82} shows, extending the original round-5 \marker critic recipe from $150$k to $475$k updates increased mean initial value on the same $23$ successful held-out episodes from $0.004$ to $0.082$, while candidate-maximum-$Q$ outcome AUROC on the containing $50$-episode block changed from $0.596$ to $0.514$.
The paired bootstrap $95\%$ interval for that change, $[-0.254,\,0.085]$, is too wide for this block to resolve a decline.

\paragraph{Diagnostics on sampled actions.}
To look at the scores reranking actually uses, we had each critic score the same stored banks of actor proposals.
We computed outcome AUROC of the maximum and mean candidate $Q$, the recorded action's score, score spread and top-score margins, and sensitivity to action perturbations.
They showed weak or unstable action dependence.
Because the robot never executed the alternative proposals, every label was still the outcome of a recorded trajectory.
A critic could raise its candidate-score AUROC just by telling easy states from hard ones, without choosing better actions within a state.
In the \nut round-5 screen, the apparent improvements over the first deployed round-5 critic had bootstrap intervals containing zero, and differences between training seeds were comparable to the differences between recipes.
We examined the rollout blocks separately, since pooling blocks with different outcome frequencies can inflate discrimination.
We used these screens to choose candidates for robot tests.
They did not identify an offline winner.

\begin{figure}[t]
  \centering
  \includegraphics[width=\linewidth]{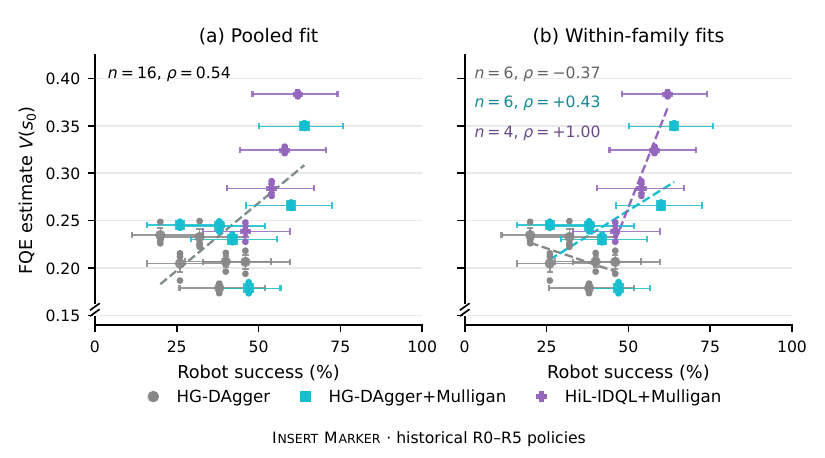}
  \caption{FQE backtest on $16$ historical \marker policies, shown with (a) a pooled fit and (b) separate fits by policy family.
    Dashed lines are descriptive least-squares fits, and $\rho$ denotes Spearman correlation.
    The value axes are zoomed to $[0.14,0.425]$.
    Markers average three evaluator seeds, small dots show individual seeds, and vertical whiskers give one standard error across seeds.
    Horizontal whiskers are Wilson $95\%$ intervals from recorded robot counts ($50$ attempts per policy, except $100$ for round-2 DP), without adjustment for repeated initial states.
    Evaluators use $\gamma=0.995$, timeout bootstrapping, and their final $40$k checkpoint.
    FQE scores every policy on the same $150$ round-5 reset images ($50$ initial states), but each robot success rate comes from that policy's own historical round.
    The round-5 BoN point uses the original $31/50$ critic, not the critic selected later.
    Because the policies are related and the two axes use different initial-state distributions, the pooled fit is hard to interpret.}
  \label{fig:81}
\end{figure}

\paragraph{Fitted-Q evaluation.}
We also tested fitted-Q evaluation (FQE), which fits separate policy-evaluation critics to estimate the return of each fixed actor or reranked policy from logged transitions.
We backtested FQE on \marker with $\gamma=0.995$, bootstrapping through time-limit truncations and keeping the final converged checkpoint rather than the one with the smallest validation Bellman error.
Pooled over $16$ historical policies, the FQE estimates correlated with measured robot success at Spearman $\rho=0.54$ (\cref{fig:81}).
The association depended on policy family, with $\rho=-0.37$ for HG-DAgger ($6$ policies), $0.43$ for HG-DAgger+\ours{} ($6$ policies), and $1.00$ for HiL-IDQL+\ours{} (only $4$ policies).
The nine-policy subset, excluding HG-DAgger and the round-0 DP actor, had $\rho=0.78$.
This retrospective result did not transfer to comparing a DP actor with its reranked version.
FQE favored reranking in several comparisons where the robot measurements were near-ties, and adding baseline data to the fit did not change this.
In the \nut round-5 candidate screen, FQE estimates across critics ranged only from $0.487$ to $0.503$, comparable to evaluator-seed variation.
The FiLM candidate ranked highly by candidate-score AUROC but lowest among those critics by FQE.
All three candidates we then tested on the robot scored $36/50$.
Offline, we could not settle the close policy comparisons that deployment needed.

\paragraph{Practical training-budget choice.}
With no reliable offline metric, we fell back on critics that had worked on the robot before.
We used them as references and kept the update-to-data ratio (UTD) comparable as the training pool grew.
Here UTD counts critic optimizer updates per loaded training transition and ignores batch size.
At a fixed batch size, it is proportional to how often each transition is sampled.
On \nut, keeping $150$k updates while the pool grew from $80{,}300$ to $133{,}953$ and then $173{,}874$ transitions let UTD fall from $1.87$ to $1.12$ and then $0.86$.
Later $200$k--$250$k candidate checkpoints brought the step-based ratio back to $1.15$--$1.44$, within the range of the earlier successful runs.
The critic deployed for the headline \nut R5 evaluation trained on $201{,}788$ transitions that add the rollouts of the superseded R5 evaluations and used its $200$k-update checkpoint (UTD $0.99$).
The ratio is only a heuristic, and it misses batch-size changes, which alter data exposure without moving a step-based ratio.
In the end, we used the offline diagnostics to flag degradation and robot evaluations to measure performance.

\subsection{Critic training-data composition}
\label{app:18}

We test whether reranking depends on the critic's non-human data.
On the final R3 with-CF datasets of \sqnarrow and \sqbroad, we train new DIVL $Q/V$ heads behind the seed-matched frozen actor and normalizer, with the critic sampling only chunks from $\mathcal{D}_h$, those whose every executed action is human-controlled and whose episode succeeded ($24{,}591$ of $126{,}783$ valid chunks on \sqnarrow and $56{,}583$ of $202{,}819$ on \sqbroad).
The control for each seed is the existing all-data DIVL critic of \cref{app:24}, trained with the deployed sampler ($50\%$ human-success anchors, $50\%$ DAgger anchors).
The two arms use the same datasets, normalizer, return bounds, architecture, optimizer, number of updates ($150$k on \sqnarrow, $250$k on \sqbroad), and seed.
We evaluate each final checkpoint at $N{=}1$ and $N{=}32$ on $400$ fixed Sobol initial states.
$N{=}1$ uses only the actor, so we can check that both arms deploy the same policy.

Restricting the critic to successful human chunks does not improve reranking (\cref{tab:120}).
On \sqnarrow the human-only critic is lower in all five seeds, and the $95\%$ interval of the difference excludes zero.
On \sqbroad, the interval covers zero.
At $N{=}1$ both arms deploy the identical actor, so their differences are pure evaluation noise.
That noise has a seed-level standard deviation of $1.6$ and $1.2$ points, as large as the $N{=}32$ effects.
On these probes, even the all-data critic's gain over the actor is indistinguishable from zero.
We cannot attribute a reranking benefit to the non-human data at this round.
The ablation shows only that removing that data does not help, and that it costs about one point on \sqnarrow, where success is near ceiling.
All of this rests on final probes over $400$ initial states, not the $8{,}000$- and $30{,}000$-state grids of \cref{app:24}.
Both R3 datasets come from rounds whose sampler used the evaluation-grid cell term (\cref{app:28}), which affects both arms equally.

\begin{table}[!htbp]
  \centering
  \footnotesize
  \caption{Human-only versus all-data critic behind the same frozen R3 actor, five paired seeds per task, final-step success on $400$ fixed Sobol initial states.
    $\Delta$ subtracts all-data from human-only success in percentage points, and the rerank gain subtracts $N{=}1$ from $N{=}32$ success within an arm.
    We report means over seeds with two-sided Student-$t$ $95\%$ intervals over the five seed differences.
    The critic plays no role at $N{=}1$.}
  \label{tab:120}
  \setlength{\tabcolsep}{3.5pt}
  \begin{tabular}{lrrrrrr}
  \hline
  Task & \begin{tabular}[b]{@{}c@{}}All-data \\ $N{=}32$\end{tabular} & \begin{tabular}[b]{@{}c@{}}Human-only \\ $N{=}32$\end{tabular} & \begin{tabular}[b]{@{}c@{}}$\Delta$ at \\ $N{=}32$\end{tabular} & \begin{tabular}[b]{@{}c@{}}$\Delta$ at $N{=}1$ \\ (same actor)\end{tabular} & \begin{tabular}[b]{@{}c@{}}Rerank gain \\ all-data\end{tabular} & \begin{tabular}[b]{@{}c@{}}Rerank gain \\ human-only\end{tabular} \\
  \hline
  \sqnarrow & $96.25\%$ & $95.15\%$ & \begin{tabular}[t]{@{}r@{}}$-1.10$ \\ {\scriptsize $[-1.91, -0.29]$}\end{tabular} & \begin{tabular}[t]{@{}r@{}}$+0.05$ \\ {\scriptsize $[-1.88, +1.98]$}\end{tabular} & \begin{tabular}[t]{@{}r@{}}$+0.40$ \\ {\scriptsize $[-0.89, +1.69]$}\end{tabular} & \begin{tabular}[t]{@{}r@{}}$-0.75$ \\ {\scriptsize $[-1.83, +0.33]$}\end{tabular} \\
  \sqbroad & $85.80\%$ & $85.00\%$ & \begin{tabular}[t]{@{}r@{}}$-0.80$ \\ {\scriptsize $[-2.96, +1.36]$}\end{tabular} & \begin{tabular}[t]{@{}r@{}}$+0.05$ \\ {\scriptsize $[-1.38, +1.48]$}\end{tabular} & \begin{tabular}[t]{@{}r@{}}$-0.40$ \\ {\scriptsize $[-1.93, +1.13]$}\end{tabular} & \begin{tabular}[t]{@{}r@{}}$-1.25$ \\ {\scriptsize $[-2.80, +0.30]$}\end{tabular} \\
  \hline
\end{tabular}

\end{table}

\subsection{Critic-objective variations}
\label{app:19}

\begin{figure}[t]
  \centering
  \includegraphics[width=\linewidth]{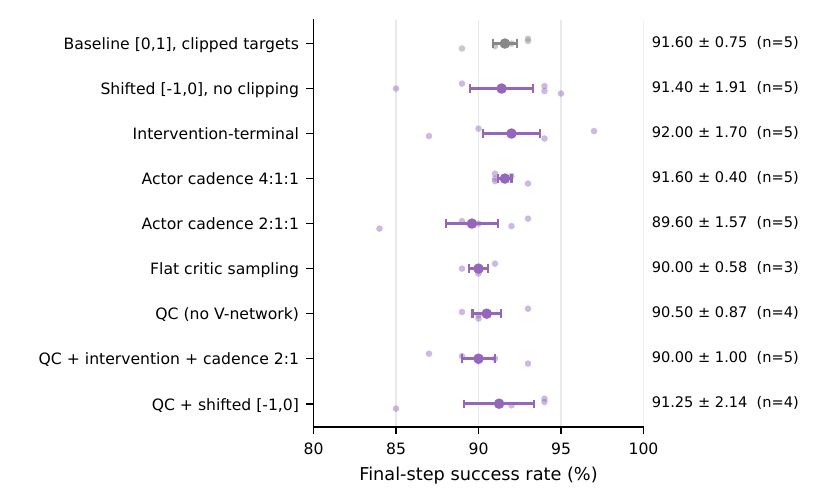}
  \caption{Final-step ($300$k) success rates for all nine critic-objective conditions in \cref{tab:121}.
    Large dots and error bars show the mean and one standard error, and small dots show the completed seeds.
    The two QC conditions with four seeds exclude failed runs, and flat sampling has three seeds.}
  \label{fig:103}
\end{figure}

We ablate several design choices for the critic objective.
Their final-step means are similar (\cref{tab:121,fig:103}).
These runs come from a superseded ten-round DAgger campaign on the same MimicGen task as \sqnarrow.
It shares no datasets with the R0--R3 campaign of \cref{tab:123}, so absolute rates here are not comparable to any main-text \sqnarrow number.
Runs use an MLP critic (hidden dims $[1024,1024,1024]$ with layer norm), chunk size $5$, $128$ action samples, batch size $1024$, $300$k training steps, and $100$ evaluation episodes every $50$k steps.
We report final-step success averaged over seeds.
The deployed simulation critic uses chunk size $6$ (\cref{app:35}).

\begin{table}[tbp]
  \centering
  \footnotesize
  \caption{Critic-objective variations on the \sqnarrow task, final-step ($300$k) success rate, mean $\pm$ standard error over training seeds, on a superseded ten-round DAgger campaign rather than the R0--R3 campaign of \cref{tab:123}.
    The reported critics use the unshifted reward and clipped targets of the baseline row (\cref{tab:132}).
    ``QC'' denotes the no-$V$-network variant that takes a best-of-$N$ max over $Q$ as the TD target.}
  \label{tab:121}
  \begin{tabular}{lcc}
\hline
Condition & Seeds & Final SR (\%) \\
\hline
Baseline, reward $[0,1]$, clipped targets & 5 & $91.60 \pm 0.75$ \\
Shifted reward $[-1,0]$, no target clipping & 5 & $91.40 \pm 1.91$ \\
Intervention frames treated as terminal & 5 & $92.00 \pm 1.70$ \\
Actor update cadence $4{:}1{:}1$ & 5 & $91.60 \pm 0.40$ \\
Actor update cadence $2{:}1{:}1$ & 5 & $89.60 \pm 1.57$ \\
Flat (uniform) critic sampling & 3 & $90.00 \pm 0.58$ \\
QC, no $V$-network & 4 & $90.50 \pm 0.87$ \\
QC $+$ intervention-terminal $+$ cadence $2{:}1$ & 5 & $90.00 \pm 1.00$ \\
QC $+$ shifted reward $[-1,0]$ & 4 & $91.25 \pm 2.14$ \\
\hline
\end{tabular}

\end{table}

\FloatBarrier
\subsection{Distributional value head (DIVL) vs.\ scalar IQL critic}
\label{app:24}

\paragraph{Adaptive DIVL in simulation.}
We compare adaptive DIVL against a scalar-IQL critic in the $34$ human-in-the-loop simulation cells across rounds R0--R3, with five matched seeds per cell.
Both critics train on the same data and rerank $N{=}32$ proposals from the same actor, and each seed is scored on the same grid of $8{,}000$ initial states for \sqnarrow and $30{,}000$ for \sqbroad.
Because adaptive DIVL changes two things at once, the distributional value head and the entropy-adaptive quantile ($\alpha=0.4$), this comparison does not isolate the distributional form.

\cref{tab:125} summarizes the paired differences, computed as DIVL minus scalar-IQL success within each cell and seed, averaged over the five seeds of a cell, and weighted equally across cells in the task and combined rows.
They are small, and their sign varies from cell to cell.
We report the human-in-the-loop simulation $N{=}32$ results with these DIVL heads because they match the value formulation of the real-world critics.

\begin{table}[!htbp]
  \centering
  \footnotesize
  \caption{Adaptive DIVL versus scalar IQL on the $34$ human-in-the-loop simulation cells, five paired seeds per cell.
    $\Delta$ subtracts scalar from DIVL success in percentage points and is averaged over seeds within a cell.
    The mean weights every cell equally, and the range and sign counts are over cell means.}
  \label{tab:125}
  \begin{tabular}{lrrrr}
\hline
Task & Cells & Mean $\Delta$ (pp) & Cell range (pp) & Cells favoring DIVL / scalar \\
\hline
\sqnarrow & $16$ & $+0.09$ & $[-0.47, +0.55]$ & $9$ / $7$ \\
\sqbroad & $18$ & $+0.00$ & $[-0.32, +0.18]$ & $11$ / $7$ \\
Both tasks & $34$ & $+0.04$ & $[-0.47, +0.55]$ & $20$ / $14$ \\
\hline
\end{tabular}

\end{table}

\paragraph{Earlier fixed-versus-adaptive screens.}
\cref{tab:124} separates the two components in three regimes.
\texttt{divl-fixed} uses the distributional value network with a fixed quantile ($\alpha=0$), and \texttt{divl-adaptive} also enables the entropy-adaptive quantile ($\alpha=0.4$).
Within a row the conditions are matched.
Across rows the campaign, seed count, reranking budget, and evaluation set all change, which rules out a controlled comparison across success regimes.
The table is also our only fixed-versus-adaptive comparison, since the comparison above evaluates adaptive DIVL alone.

\begin{table}[!htbp]
  \centering
  \footnotesize
  \caption{DIVL vs.\ scalar IQL, $\Delta$ in percentage points against the scalar control within each regime, matched within a row but not across rows.
    The low-SR and $\sim$$94\%$ rows are round-$1$ \sqbroad and \sqnarrow cells, $5$ seeds, final-step $N{=}32$ SR on $400$ fixed-Sobol initial states.
      The $\sim$$90\%$ row is round $15$ of a superseded \sqbroad campaign, $3$ seeds, $N{=}128$ SR on $200$ episodes.
    Scalar controls show mean $\pm$ SE.
    The distributional form alone (\texttt{divl-fixed}) gives no consistent gain, while the adaptive variant has a positive mean difference at low success and negative differences in both high-success regimes.}
  \label{tab:124}
  \begin{tabular}{lccc}
\hline
Regime (scalar-IQL base SR) & scalar IQL & \texttt{divl-fixed} & \texttt{divl-adaptive} \\
\hline
Low SR ($\sim70\%$) & $69.95 \pm 0.74$ & $+1.45$ & $+4.05$ \\
High SR ($\sim90\%$) & $90.17 \pm 0.73$ & $-1.17$ & $-0.83$ \\
High SR ($\sim94\%$) & $93.80 \pm 0.65$ & $+0.35$ & $-0.35$ \\
\hline
\end{tabular}

\end{table}

We also tested whether the default success-biased critic sampler accounts for the small differences in the high-success regime.
We re-ran the $\sim$$94\%$ \sqnarrow round-$1$ cell under three critic samplers.
  The scalar-control means stay similar, and neither DIVL variant improves consistently across the three (\cref{tab:126}).

  \begin{table}[!htbp]
  \centering
  \footnotesize
  \caption{DIVL under three critic samplers in the $\sim$$94\%$ regime ($5$ seeds each, final-step $N{=}32$ SR on $400$ fixed-Sobol initial states, $\Delta$ against the scalar control \emph{within} each sampler).
\texttt{straddled} is success-biased ($50\%$ human success / $50\%$ DAgger), \texttt{balanced\_outcome} is failure-biased ($50\%$ success / $50\%$ fail-or-intervened), and \texttt{flat} is uniform over the natural distribution.}
\label{tab:126}
\begin{tabular}{lccc}
\hline
Critic sampler & scalar IQL (control) & \texttt{divl-fixed} & \texttt{divl-adaptive} \\
\hline
\texttt{straddled} & $93.80 \pm 0.65$ & $94.15$ ($+0.35$) & $93.45$ ($-0.35$) \\
\texttt{balanced\_outcome} & $93.95 \pm 0.70$ & $92.60$ ($-1.35$) & $93.25$ ($-0.70$) \\
\texttt{flat} & $93.80 \pm 0.50$ & $93.55$ ($-0.25$) & $92.70$ ($-1.10$) \\
\hline
\end{tabular}

\end{table}

\subsection{Critic ensemble size (REDQ)}
\label{app:42}

We use a $2$-$Q$ min throughout.
\cref{tab:137} compares it against a larger, REDQ-style~\cite{chen2021redq} ensemble of $10$ critics under two aggregations, holding the actor and critic architectures fixed.
Neither larger ensemble improves the mean success rate, and REDQ-2 has a lower mean than the control.
The late round is round $15$ of the superseded \sqbroad campaign behind the $\sim$$90\%$ row of \cref{tab:124}.
  The same ablation on the main campaign's \sqbroad round-$1$ cell is likewise flat.

  \begin{table}[tbp]
  \centering
  \footnotesize
  \caption{Critic ensemble size on \sqbroad at round $15$ of a superseded DAgger campaign.
$5$ seeds each, $300$k steps, final-step $N{=}32$ success rate (mean $\pm$ standard error).
  ``REDQ-2'' takes a min over a random pair of critics resampled at every $Q$ evaluation.
  ``Single'' draws one random critic per evaluation.}
  \label{tab:137}
  \begin{tabular}{lccc}
\hline
Critic ensemble & $|$ensemble$|$ & Final $N{=}32$ SR (\%) & $\Delta$ \\
\hline
$2$-$Q$ min & $2$ & $95.70 \pm 0.86$ & --- \\
REDQ-2 & $10$ & $93.50 \pm 0.57$ & $-2.20$ \\
Single random critic & $10$ & $95.60 \pm 0.62$ & $-0.10$ \\
\hline
\end{tabular}

  \end{table}

  \subsection{Reranking budget \texorpdfstring{$N$}{N}}
  \label{app:44}

  \cref{tab:130} lists the deployed budgets, and the actor-only ($N=1$) and reranked ($N=32$) results in \cref{app:29} measure the effect of reranking with the deployed simulation critics.
  On the superseded campaigns with scalar critics, larger budgets give no further gain.
  \sqbroad at round $15$ reaches $89.5 \pm 0.8\%$, $92.0 \pm 0.5\%$, and $91.6 \pm 0.6\%$ at $N = 1$, $32$, and $128$ ($11$ seeds, $200$ episodes each), and \sqnarrow at round $10$ reaches $96.3$, $97.6$, and $97.6\%$ ($5$ seeds).

  On the robot, four \nut R5 critics rescoring fixed banks of $32$ proposals changed the outcome AUROC of the top-scored candidate by less than $0.003$ between $N=16$ and $N=32$.

\clearpage\suppressfloats[t]

\section{Extended Related Work}
\label{app:43}
\backlink{\cref{sec:112}}

Here we extend \cref{sec:112} and compare the reset ranges and data budgets that prior real-world systems report with ours.

\subsection{Real-world RL for robot manipulation}
\label{app:48}

Reinforcement learning on physical robots predates deep policies.
Policy search over motor primitives learned motor skills on real arms~\citep{peters2008motorskills,kober2009motorprimitives,kober2013survey}, and model-based policy search learned manipulator control from a handful of trials~\citep{deisenroth2011pilco,deisenroth2011manipulator}.
Guided policy search then trained visuomotor policies from pixels end to end on a real robot~\citep{levine2016visuomotor}, followed by asynchronous off-policy training across several arms~\citep{gu2017asynchronous}, large-scale self-supervised grasping~\citep{levine2018grasping,kalashnikov2021mtopt}, and model-free and model-based training on low-cost hands and arms~\citep{haarnoja2018sacapps,zhu2019dexterous,nagabandi2020pddm,zhu2020ingredients,wu2023daydreamer}.
Later systems ran without human resets~\citep{gupta2021resetfree,hu2023reboot,sharma2023medal,mendonca2024mobile} or ran iterated batch RL across a fleet~\citep{lampe2024stacking,bousmalis2023robocat}.
CASHER~\citep{torne2024casher} moves much of this learning into simulation, running RL in crowdsourced reconstructions of real scenes so that performance grows faster than the human effort spent.

Sample-efficient real-world RL trains policies directly on the physical robot, without corrections from a human in the loop.
QT-Opt~\citep{kalashnikov2018qtopt} demonstrated large-scale grasping from pixels but required months of robot time across a fleet.
SERL~\citep{luo2024serl} and CQN~\citep{seo2024cqn} bring online training down to minutes on a single robot, under narrow initial-state distributions (\cref{tab:138}).
IBRL~\citep{hu2023ibrl} bootstraps RL exploration with an imitation policy for faster convergence, and RL-100~\citep{lei2025rl100} combines diffusion policies with online RL and action chunking.
Both learn from autonomous exploration alone.
VLAC~\citep{zhai2025vlac} finetunes a VLA on the real robot with dense progress rewards from a vision-language critic.
We instead keep the operator in the loop so that supervision reaches states the policy cannot yet solve, and update the policy offline between rounds.
We compile the supervision each system spends per task in \cref{app:39}.

\subsection{RL finetuning of expressive policies}
\label{app:47}

Applying RL to action-chunked diffusion, flow, or VLA policies is difficult because optimizing such a policy against a Q-function requires backpropagation through the multi-step denoising chain, which is unstable~\citep{li2026qam,dong2025expo}.

\emph{Direct policy optimization} updates the expressive policy's parameters with an RL objective.
DPPO~\citep{ren2024dppo} and \citet{mcallister2025flow} apply policy gradients through the full sampling chain.
ConRFT~\citep{chen2025conrft} and flow Q-learning~\citep{park2025flowqlearning} route Q-function gradients through a single-step distilled policy, and QAM~\citep{li2026qam} uses adjoint matching to construct stable step-wise objectives for flow policies that use the critic's action gradient.

\emph{Auxiliary residual or edit networks} keep the base policy fixed and train a small network to refine its outputs: residual action corrections~\citep{ankile2024resip,ankile2025resfit,yuan2024policydecorator}, RL in the latent noise space of the sampler~\citep{wagenmaker2025dsrl}, residual RL combined with distillation~\citep{xiao2025pld}, and EXPO~\citep{dong2025expo}, which trains a small Gaussian edit policy alongside on-the-fly value-based action selection.
These leave the base objective untouched at the cost of an additional network.

\emph{Sampling-based policy extraction} draws several candidates from the base policy and executes the highest-valued one.
IDQL~\citep{hansenestruch2023idql} and V-GPS~\citep{nakamoto2024vgps} rerank samples with values learned by IQL~\citep{kostrikov2022iql} and Cal-QL~\citep{nakamoto2023cal}, respectively, without action chunking.
PA-RL~\citep{mark2024parl} uses Q-optimized actions as supervised targets to finetune OpenVLA~\citep{kim_openvla_2024}.
\citet{li2025qchunking} extend temporal-difference learning to action chunks, which our chunked value targets build on (\cref{sec:109}).
Our best-of-$N$ reranking comes from this family and treats the policy as a black box that proposes actions, which lets us apply it to any actor class without changing the base model.

\FloatBarrier

\subsection{Initial-state randomization: ours and prior real-world RL}
\label{app:45}
\backlink{\cref{sec:113,subsec:114}}

\cref{tab:138} records the initial-state randomization ranges that prior real-world RL and HiL-RL systems report, alongside ours.
We take all numbers verbatim from the cited papers' setup descriptions or supplementary material.
For each system we list the task with the widest published range together with its most precision-critical task, since randomization width and required precision trade off against each other.

\begin{table}[tbp]
  \centering
  \footnotesize
  \setlength{\tabcolsep}{2.75pt}
  \newcolumntype{L}{>{\raggedright\arraybackslash}p{1.85cm}}
  \newcolumntype{P}{>{\raggedright\arraybackslash}p{3.15cm}}
  \caption{Initial-state randomization as \emph{reported} by prior real-world RL systems, and ours.
    ``Varies'' distinguishes randomization of the object pose from randomization of the arm's start pose with the object fixed, and marks the tasks in which the \emph{target} the object must reach is also randomized.
    DoF counts the independent dimensions of the initial state that vary (three for an object's $x$, $y$, and yaw).
    Where a paper reports a range for one element only, we take the remaining elements to be fixed.
    We quote position ranges as published.
    Orientation is in-plane (yaw).
    Table numbers under ``Source'' refer to each paper's own numbering (appendix tables for HiL-SERL and ConRFT).
    $^{\ddagger}$RL-100 states that the blank's and the half-orange's rotations are randomized but gives no range.}
  \label{tab:138}
  \begin{tabular}{llLcPll}
    \hline
    Method                          & Task            & Varies          & DoF & Position range                                           & Yaw                   & Source   \\
    \hline
    HiL-SERL~\citep{luo2024hilserl} & IKEA side panel & arm             & 3   & $8$\,cm in $x$, $y$                                      & $1^\circ$             & Tab.\ 6  \\
    HiL-SERL~\citep{luo2024hilserl} & Cable clipping  & object          & 3   & $4$\,cm in $x$, $y$                                      & $10^\circ$            & Tab.\ 5  \\
    ConRFT~\citep{chen2025conrft}   & Pick banana     & object          & 2   & $3$\,cm in $x$, $y$                                      & none                  & Tab.\ V  \\
    ConRFT~\citep{chen2025conrft}   & Insert wheel    & object          & 2   & $2$\,cm in $x$, $y$                                      & none                  & Tab.\ XI \\
    RL-100~\citep{lei2025rl100}     & Box folding     & object          & 3   & $75{\times}55$\,cm                                       & unstated$^{\ddagger}$ & Suppl.   \\
    RL-100~\citep{lei2025rl100}     & Orange juicing  & object          & 3   & $29{\times}20$\,cm tray                                  & unstated$^{\ddagger}$ & Suppl.   \\
    RL-100~\citep{lei2025rl100}     & Pouring         & object + target & 4   & cup $39{\times}16$, plate $28{\times}20$\,cm             & none                  & Suppl.   \\
    RL-100~\citep{lei2025rl100}     & Unscrewing      & object          & 2   & $12.5$\,cm-diam.\ circle                                 & none                  & Suppl.   \\
    SERL~\citep{luo2024serl}        & Cable routing   & arm             & 2   & $20{\times}20$\,cm                                       & none                  & Tab.\ 2  \\
    SERL~\citep{luo2024serl}        & PCB insertion   & arm             & 2   & $10{\times}10$\,cm                                       & none                  & Tab.\ 2  \\
    \hline
    Ours                            & \marker         & object + target & 5   & pen $15.2{\times}30.5$, holder $5.1{\times}10.2$\,cm     & $360^\circ$           & ---      \\
    Ours                            & \nut            & object + target & 5   & nut $25.4{\times}45.7$, peg $5.1{\times}10.2$\,cm        & $360^\circ$           & ---      \\
    Ours                            & \cable          & object + target & 7   & rope $40.6$\,cm in $x$, each clip $40.6{\times}22.9$\,cm & clips $\pm45^\circ$   & ---      \\
    Ours (sim)                      & \sqnarrow       & object          & 3   & nut $0.5{\times}11.5$\,cm, peg fixed                     & $360^\circ$           & ---      \\
    Ours (sim)                      & \sqbroad        & object + target & 5   & nut $23.0{\times}51.0$, peg $40.0{\times}40.0$\,cm       & $360^\circ$           & ---      \\
    \hline
  \end{tabular}
\end{table}

We caution against reading this table as a ranking by area, for four reasons that cut in both directions.

\textbf{What is randomized differs.} SERL and the scripted-reset tasks of HiL-SERL vary the \emph{arm's} initial pose while the object stays fixed.
SERL expresses proprioception in the frame of the end-effector's initial pose, which its authors note is ``equivalent to physically moving the target when viewed relatively from the frame attached to the end-effector''~\citep{luo2024serl}.
The observation absorbs the reset by design. A reset of this kind leaves the policy's relative problem unchanged, and its extent cannot be compared with variation in where the object is.

\textbf{A stated region upper-bounds the variation exercised.} RL-100's box-folding blank is described as randomized in translation and rotation over a $75 \times 55$\,cm region, but the same paper reports that training demonstrations are concentrated near-aligned, at $\theta \in [-30^\circ, 30^\circ]$.
The paper also states that it does ``not maximize initial randomization'' and instead emphasizes task-specific difficulty~\citep{lei2025rl100}.
An entry in this column gives the region a paper says it samples from, not the spread its rollouts reach.
RL-100 differs from the other rows in budget and sensing as well.
It spends an order of magnitude more per task, $1{,}890$ episodes over $31.5$ robot-hours for orange juicing and $1{,}950$ episodes over $24.5$ hours for towel folding, where the human-in-the-loop systems typically use tens of demonstrations and one to two hours of online training.
Its policies see calibrated point clouds. The system maps depth into the robot's root frame through Charuco intrinsics and an AprilTag extrinsic and then crops the cloud to the region that holds the task geometry.
Ours act on raw RGB with no calibration or segmentation step, and an equal range does not mean an equal problem.

\textbf{Center ranges versus object regions.} We report the range of the object's \emph{center}, and the envelope the whole object sweeps is larger by roughly its footprint.
When a paper instead quotes a region the object must lie \emph{within}, its center range is smaller by about the same amount, which for a towel or a paperboard blank is a large correction.

\textbf{Evaluation protocols.} With few exceptions, each paper reports success rates as totals from its own sessions, with the arm order, blinding, and per-episode reset procedure left to the reader.
This matters because a policy can drop from $13/20$ to $0/8$ on identical initial conditions after an unrelated change to the lab~\citep{kressgazit2024evaluation}.
We tabulate what each paper states.
Our own evaluations pair the arms on the same held-out Sobol states, run blind and interleaved within a session, and place each reset from a per-episode schematic (\cref{app:25}).
Readers can check the pairing and the realized initial states in the released video of every evaluation episode behind the compared methods' reported results, which comes with its per-episode outcome (\url{https://huggingface.co/mulligan}).

In the prior grasp-and-insert tasks at millimeter clearance that randomize the object pose, the region spans a few centimeters and yaw at most $10^\circ$.
The tasks with large stated regions end in folding, pouring, or seating a compliant object.
On \marker and \nut, we place the object anywhere in a region tens of centimeters across, at any yaw, and the task still ends in a millimeter-clearance insertion.

\FloatBarrier

\subsection{Data and interaction budgets: ours and prior real-world RL}
\label{app:39}
\backlink{\cref{sec:113}}

\cref{tab:138} records how wide each system's initial-state distribution is.
\cref{tab:135} records how much supervision and on-robot experience each system spent covering it.
Neither table compares cleanly across systems, since the robots, tasks, required precision, and units all differ.
HiL-SERL and SERL count environment transitions in a replay buffer while ConRFT and RL-100 count episodes, and each paper folds scripted resets, intended stops and onboard computation into its ``training time'' in its own way.
We keep each paper's unit and mark a cell \emph{unstated} when the paper does not give the number.

\begin{table}[tbp]
  \centering
  \footnotesize
  \setlength{\tabcolsep}{2pt}
  \newcolumntype{D}{>{\raggedright\arraybackslash}p{1.35cm}}
  \newcolumntype{I}{>{\raggedright\arraybackslash}p{1.95cm}}
  \newcolumntype{A}{>{\raggedright\arraybackslash}p{2.45cm}}
  \newcolumntype{W}{>{\raggedright\arraybackslash}p{1.25cm}}
  \caption{Data and interaction budgets as \emph{reported} by prior real-world RL systems, and ours.
    Rows match \cref{tab:138} task for task wherever the source paper reports per-task numbers.
    ``Demos'' are the offline human demonstrations that initialize learning, ``Human corrections'' are in-the-loop interventions, and ``Autonomous experience'' is the data the policy collected on the robot, in the paper's own unit.
    ``Robot time'' is the wall-clock number each paper headlines.
    Its definition varies, with scripted motion, intended stops and onboard computation included in HiL-SERL and ConRFT, and computation, resets and intended stops in SERL.
    RL-100 reports data-collection time per stage, and the entry here sums its three stages (demonstration, iterative offline RL, online RL).
    Table numbers under ``Source'' refer to each paper's own numbering.
    $^{\dagger}$Both human-in-the-loop systems report the intervention \emph{rate}, the ratio of intervened to total timesteps, only as a training curve, for every task in HiL-SERL (Figs.~5 and~26) and for representative tasks in ConRFT (Figs.~3--4).
    Neither reports a count of interventions or a total intervened time for any task.
    $^{\ddagger}$HiL-SERL's ``total number of RL transitions'' is the replay-buffer size at convergence and includes human-intervention timesteps, so it is not purely autonomous experience.
    The p1/p2 entries are the two side panels, which the paper trains and times separately.
    $^{*}$ConRFT states only that the policy collects ``approximately 80 to 120 successful and failed trajectories'' during the online stage, for all tasks together.
    $^{\S}$These systems have no in-the-loop corrections and train from demonstrations plus autonomous rollouts.
    $^{\P}$Sum over the two Orange Juicing subtasks (Placing and Removal), which Tab.~S3 reports separately.
    $^{\|}$Our rows count the HG-DAgger+\ours{} arm's own training data at the final round (\cref{tab:123}), split into human-controlled frames (demonstrations, corrections, and counterfactual replays) and autonomous frames of the same episodes, at $15$\,Hz.
    For our rows, ``robot time'' adds up the recorded length of that arm's collection episodes and of the held-out evaluations of the three headline arms.
    It counts full episodes, including any tail that a later outcome review trimmed, and leaves out resets and operator setup.
    On \cable, we measured session wall-clock time from save timestamps at about three times the recorded trajectory time.}
  \label{tab:135}
  \begin{tabular}{llDIAWl}
    \hline
    Method                          & Task            & Demos                 & Human corrections                                                  & Autonomous experience                                    & Robot time                            & Source      \\
    \hline
    HiL-SERL~\citep{luo2024hilserl} & IKEA side panel & $20$                  & rate curve only$^{\dagger}$                                        & $31{,}000$ / $36{,}000$ transitions$^{\ddagger}$ (p1/p2) & $2$ / $1.75$\,h                       & Tab.\ 1, 6  \\
    HiL-SERL~\citep{luo2024hilserl} & Cable clipping  & $20$                  & rate curve only$^{\dagger}$                                        & $28{,}000$ transitions$^{\ddagger}$                      & $1.25$\,h                             & Tab.\ 1, 5  \\
    ConRFT~\citep{chen2025conrft}   & Pick banana     & $20$                  & rate curve only$^{\dagger}$                                        & $80$--$120$ episodes$^{*}$                               & $45$\,min                             & Tab.\ I, V  \\
    ConRFT~\citep{chen2025conrft}   & Insert wheel    & $30$                  & rate curve only$^{\dagger}$                                        & $80$--$120$ episodes$^{*}$                               & $60$\,min                             & Tab.\ I, XI \\
    RL-100~\citep{lei2025rl100}     & Box folding     & $200$ ($3$\,h)        & none$^{\S}$                                                        & $608$ offline + $186$ online epi.                        & $15.3$\,h                             & Tab.\ S3    \\
    RL-100~\citep{lei2025rl100}     & Orange juicing  & $109$ ($2$\,h)$^{\P}$ & none$^{\S}$                                                        & $791$ offline + $990$ online epi.$^{\P}$                 & $31.5$\,h$^{\P}$                      & Tab.\ S3    \\
    RL-100~\citep{lei2025rl100}     & Pouring         & $64$ ($1$\,h)         & none$^{\S}$                                                        & $741$ offline + $129$ online epi.                        & $9.3$\,h                              & Tab.\ S3    \\
    RL-100~\citep{lei2025rl100}     & Unscrewing      & $31$ ($0.5$\,h)       & none$^{\S}$                                                        & $467$ offline + $288$ online epi.                        & $8$\,h                                & Tab.\ S3    \\
    SERL~\citep{luo2024serl}        & Cable routing   & $20$                  & none$^{\S}$                                                        & unstated                                                 & $31$\,min                             & Tab.\ 2     \\
    SERL~\citep{luo2024serl}        & PCB insertion   & $20$                  & none$^{\S}$                                                        & unstated                                                 & $20$\,min                             & Tab.\ 2     \\
    \hline
    Ours                            & \marker         & $100$ / arm (R0)      & R0--R5, $50$ epi.\ / round / arm; $38{,}017$ human frames$^{\|}$   & $43{,}509$ frames$^{\|}$                                 & $91$\,min $+$ $367$\,min eval$^{\|}$  & ---         \\
    Ours                            & \nut            & $100$ / arm (R0)      & R0--R5, $50$ epi.\ / round / arm; $34{,}536$ human frames$^{\|}$   & $31{,}818$ frames$^{\|}$                                 & $74$\,min $+$ $277$\,min eval$^{\|}$  & ---         \\
    Ours                            & \cable          & $100$ / arm (R0)      & R0--R5, $100$ epi.\ / round / arm; $112{,}292$ human frames$^{\|}$ & $103{,}183$ frames$^{\|}$                                & $244$\,min $+$ $322$\,min eval$^{\|}$ & ---         \\
    \hline
  \end{tabular}
\end{table}

Our rows come from \cref{tab:123} and \cref{app:14} and leave out the simulation tasks, whose episodes cost no robot time.
We did not log session time, so our robot times are recorded trajectory time at $15$\,Hz (collection plus held-out evaluation) rather than the wall-clock hours these papers headline, and they exclude resets, object placement, and operator pauses.
The one session we timed from save timestamps, a $200$-episode \cable collection, took about $5.2$\,h of active session time for $104$\,min of recorded trajectory.
At that ratio, a wall-clock figure comparable to the other rows would be roughly three times ours.
Every frame carries a source tag, which makes our human-versus-autonomous split exact.
The split covers only the arm's own training data, not the evaluation rollouts that the critic also trains on.

\end{document}